\newif
\iflong
\longtrue

\iflong
\documentclass[letterpaper,11pt]{article}
\else
\documentclass{sig-alternate-2013}
\fi

\usepackage{makeidx}  

\iflong
\usepackage[margin=1in,dvips]{geometry}
\else
\renewcommand{\paragraph}[1]{{\vskip 12pt\noindent \textit{\textbf{#1}}}

}
\fi

\usepackage{amsthm}

\usepackage{graphicx}
\usepackage{verbatim}
\usepackage{amssymb}
\usepackage{amsmath}
\usepackage{float}
\usepackage{enumerate}
\usepackage{subfig}
\usepackage{hyperref}
\hypersetup{
    colorlinks=false,
    pdfborder={0 0 0},
}

\usepackage[boxed,section]{algorithm}
\usepackage{algpseudocode}

\newtheorem{theorem}{Theorem}[section]

\newtheorem{lemma}[theorem]{Lemma}
\newtheorem{corollary}[theorem]{Corollary}
\newtheorem{definition}[theorem]{Definition}
\newtheorem{claim}[theorem]{Claim}

\newcommand{\abs}[1]{|#1|}
\newcommand{\tabs}[1]{\left|#1\right|}
\newcommand{\inner}[1]{\langle#1\rangle}

\newcommand{\norm}[1]{\lVert#1\rVert}

\newcommand{\wh}{\widehat}

\usepackage{etoolbox}

\makeatletter
\newcommand{\define}[4][ignore]{%
  \ifstrequal{#1}{ignore}{}{
  \@namedef{thmtitle@#2}{#1}}%
  \@namedef{thm@#2}{#4}%
  \@namedef{thmtypen@#2}{lemma}%
  \newtheorem{thmtype@#2}[theorem]{#3}%
  \newtheorem*{thmtypealt@#2}{#3~\ref{#2}}%
}
\newcommand{\state}[1]{%
  \@ifundefined{thmtitle@#1}{
  \begin{thmtype@#1}
    }{
  \begin{thmtype@#1}[\@nameuse{thmtitle@#1}]
  }
    \label{#1}
    \@nameuse{thm@#1}
  \end{thmtype@#1}
}
\newcommand{\restate}[1]{%
  \@ifundefined{thmtitle@#1}{
    \begin{thmtypealt@#1}
    }{
  \begin{thmtypealt@#1}[\@nameuse{thmtitle@#1}]
  }
    \@nameuse{thm@#1}
  \end{thmtypealt@#1}
}
\makeatother

{\makeatletter
 \gdef\xxxmark{%
   \expandafter\ifx\csname @mpargs\endcsname\relax 
     \expandafter\ifx\csname @captype\endcsname\relax 
       \marginpar{xxx}
     \else
       xxx 
     \fi
   \else
     xxx 
   \fi}
 \gdef\xxx{\@ifnextchar[\xxx@lab\xxx@nolab}
 \long\gdef\xxx@lab[#1]#2{{\bf [\xxxmark #2 ---{\sc #1}]}}
 \long\gdef\xxx@nolab#1{{\bf [\xxxmark #1]}}
  \long\gdef\xxx@lab[#1]#2{}\long\gdef\xxx@nolab#1{}%
}

\usepackage{xcolor}
\definecolor{DarkGreen}{rgb}{0.1,0.5,0.1}
\definecolor{DarkRed}{rgb}{0.5,0.1,0.1}
\definecolor{DarkBlue}{rgb}{0.1,0.1,0.5}
\def\ShowAuthNotes{1}
\ifnum\ShowAuthNotes=1
\newcommand{\authnote}[2]{{ \footnotesize \bf{\color{DarkRed}[#1's Note:
{\color{DarkBlue}#2}]}}}
\else
\newcommand{\authnote}[2]{}
\fi

\DeclareMathOperator*{\argmin}{arg\,min}

\DeclareMathOperator*{\median}{median}
\DeclareMathOperator*{\E}{\mathbb{E}}
\DeclareMathOperator*{\Var}{\mathbb{V}}

\DeclareMathOperator*{\poly}{poly}

\def\R{\mathbb{R}}
\def\C{\mathbb{C}}
\def\Z{\mathbb{Z}}

\def\X{\mathcal{X}}

\def\eps{\epsilon}

\makeatletter 
\newenvironment{proof*}[1][\proofname]{\par 
  \normalfont \partopsep=\z@skip \topsep=\z@skip 
  \trivlist 
  \item[\hskip\labelsep 
        \itshape 
    #1\@addpunct{.}]\ignorespaces 
}{%
  \endtrivlist\@endpefalse 
} 
\makeatother

\newcommand{\Norm}[1]{\left\lVert#1\right\rVert}
\newcommand{\mper}{\,.}

\newcommand{\Set}[1]{\left\{#1\right\}}

\newcommand{\Psymb}{\mathbb{P}}

\newcommand{\Normal}{\mathrm{N}}
\newcommand{\defeq}{\,\stackrel{{\mathrm{def}}}{=}\,}
\DeclareMathOperator*{\Varr}{Var}
\DeclareMathOperator{\TV}{TV}
\DeclareMathOperator*{\ProbOp}{\Psymb}
\renewcommand{\Pr}{\ProbOp}
\newcommand{\trans}{\top}
\newcommand{\cA}{{\cal A}}
\newcommand{\cB}{{\cal B}}
\newcommand{\cC}{{\cal C}}

\iflong\else
\newfont{\mycrnotice}{ptmr8t at 7pt}
\newfont{\myconfname}{ptmri8t at 7pt}

\permission{Permission to make digital or hard copies of all or part of this work for personal or classroom use is granted without fee provided that copies are not made or distributed for profit or commercial advantage and that copies bear this notice and the full citation on the first page. Copyrights for components of this work owned by others than ACM must be honored. Abstracting with credit is permitted. To copy otherwise, or republish, to post on servers or to redistribute to lists, requires prior specific permission and/or a fee. Request permissions from permissions@acm.org.}
\conferenceinfo{STOC'15,}{June 14--17, 2015, Portland, Oregon, USA.\\
{\mycrnotice{Copyright is held by the owner/author(s). Publication rights licensed to ACM.}}}
\copyrightetc{ACM \the\acmcopyr}
\crdata{978-1-4503-3536-2/15/06\ ...\$15.00.\\
\url{http://dx.doi.org/10.1145/2746539.2746579}}

\fi

\begin{document}

\title{Tight Bounds for Learning a Mixture of Two Gaussians}

\iflong
  \author{Moritz Hardt\\\texttt{m@mrtz.org}\\IBM Research Almaden \and Eric
Price\\\texttt{ecprice@cs.utexas.edu}\\The University of Texas at Austin}
\fi

  \maketitle

\begin{abstract}
  We consider the problem of identifying the parameters of an unknown
  mixture of two arbitrary $d$-dimensional gaussians from a sequence
  of independent random samples. Our main results are upper and lower
  bounds giving a computationally efficient moment-based estimator
  with an optimal convergence rate, thus resolving a problem
  introduced by Pearson (1894). Denoting by $\sigma^2$ the variance of
  the unknown mixture, we prove that $\Theta(\sigma^{12})$ samples are
  necessary and sufficient to estimate each parameter up to constant
  additive error when $d=1.$ Our upper bound extends to arbitrary
  dimension~$d>1$ up to a (provably necessary) logarithmic loss in~$d$
  using a novel---yet simple---dimensionality reduction technique. We
  further identify several interesting special cases where the sample
  complexity is notably smaller than our optimal worst-case bound. For
  instance, if the means of the two components are separated by
  $\Omega(\sigma)$ the sample complexity reduces to $O(\sigma^2)$ and
  this is again optimal.

  Our results also apply to learning each component of the mixture up
  to small error in total variation distance, where our algorithm
  gives strong improvements in sample complexity over previous work.
  We also extend our lower bound to mixtures of $k$ Gaussians, showing
  that $\Omega(\sigma^{6k-2})$ samples are necessary to estimate each
  parameter up to constant additive error.


%
%
%

\end{abstract}

\section{Introduction}

Gaussian mixture models are among the most well-studied models in
statistics, signal processing, and computer science with a venerable
history spanning more than a century. Gaussian mixtures arise
naturally as way of explaining data that arises from two or more
homogeneous populations mixed in varying proportions.  There have been
numerous applications of gaussian mixtures in disciplines including
astronomy, biology, economics, engineering and finance.

The most basic estimation problem when dealing with any mixture model
is to approximately identify the parameters that specify the model
given access to random samples. In the case of a gaussian mixture the
model is determined by a collection of means, covariance matrices and
mixture probabilities. A sample is drawn by first selecting a
component according to the mixture probabilities and then sampling
from the normal distribution specified by the corresponding mean and
covariance. Already in 1894, Pearson~\cite{Pearson} proposed the
problem of estimating the parameters of a mixture of two
one-dimensional gaussians in the context of evolutionary
biology. Pearson analyzed a population of crabs and found that a
mixture of two gaussians faithfully explained the size of the crab
``foreheads''. He concluded that what he observed was a mixture of two
species rather than a single species and further speculated that
``\emph{a family probably breaks up first into two species, rather
  than three or more, owing to the pressure at a given time of some
  particular form of natural selection}.''

Fitting a mixture of two gaussians to the observed crab data was a
formidable task at the time that required Pearson to come up with a
good approach. His approach is based on the \emph{method of moments}
which uses the empirical moments of a distribution to distinguish
between competing models. Given $n$ samples $x_1,\dots,x_n$ the $k$-th
empirical moment is defined as $\frac{1}{n}\sum_i x_i^k$, which for
sufficiently large $n$ will approximate the true moment~$\E x_i^k.$ A
mixture of two one-dimensional gaussians has $5$ parameters so one
might hope that $5$ moments are sufficient to identify the
parameters. Pearson derived a ninth degree polynomial $p_5$ in the
first $5$ moments and located the roots of this polynomial. Each root
gives a candidate mixture that matches the first $5$ moments; there
were two valid solutions, among which Pearson selected the one whose
$6$-th moment was closest to the observed empirical $6$-th moment.

In this work, we extend the method proposed by Pearson and prove that
the extended method reliably recovers the parameters of the unknown
mixture.  Moreover, we show that the sample complexity we achieve is
essentially optimal. To illustrate the quantitative bound that we get,
if the means and variances are separated by constants and the total
variance of the mixture is $\sigma^2,$ then we show that up to
constant factors it is necessary and sufficient to use $\sigma^{12}$
samples to recover the parameters up to small additive error. Our work
can be interpreted as providing an extension of Pearson's 120-year old
estimator that achieves an optimal convergence rate.
We extend our result to arbitrary dimension~$d$ using an apparently novel but
surprisingly simple dimensionality reduction technique. This allows us
to obtain the same sample complexity in any dimension up to a
logarithmic loss in~$d$, which we can also show is necessary.

Closely related to our results is an important recent work of Kalai,
Moitra and Valiant~\cite{KalaiMV10} who gave the first proof of a
computationally efficient estimator with an inverse-polynomial
convergence rate for the problem we consider.  In particular, they
show that six moments suffice to identify a mixture of two
one-dimensional gaussians. Moreover, the result is robust in the sense
that if the parameters of a mixture with variance~$\sigma^2$ are
separated by constants, then one of the first $6$ moments must differ
by~$1/\sigma^{66}.$ In particular, the first six \emph{empirical}
moments suffice provided that they're within $1/\sigma^{66}$ of the
true moments (which happens for $n\gg \sigma^{132}$). This then leads
to an estimator up to some polynomial loss. They also show that a
solution to the $1$-dimensional problem extends to any dimension~$d$
up to some loss that's polynomial in $d$ and $\sigma$ using a suitable
dimensionality reduction technique.  In contrast to their result which
is within a polynomial factor of optimal, our result is within a
constant factor of optimal in one dimension and within a $\log_d \log
\sigma$ factor of optimal in $d$ dimensions.

\subsection{Problem description}
A mixture~$F$ of two $d$-dimensional gaussians is specified by mixing
probabilities $p_1,p_2\ge 0$ such that $p_1+p_2=1,$ two means
$\mu_1,\mu_2\in\R^d$ and two covariance matrices
$\Sigma_1,\Sigma_2\in\R^{d\times d}.$ A sample from~$F$ is generated
by first picking an integer $i\in\{1,2\}$ from the distribution
$(p_1,p_2)$ and then sampling from the $d$-dimensional gaussian
measure $\Normal(\mu_i,\Sigma_i).$

The variance $\sigma^2$ of a $1$-dimensional mixture of two gaussians
is $p_1p_2(\mu_1 - \mu_2)^2 + p_1 \sigma_1^2 + p_2 \sigma_2^2$.  For a
$d$-dimensional mixture, it is useful to define its ``variance'' as
the maximum variance of any coordinate,
\begin{equation}\label{eq:d-var}
\Var( F ) \defeq p_1p_2\|\mu_1 -\mu_2\|_\infty^2
+p_1\|\Sigma_1\|_\infty
+p_2\|\Sigma_2\|_\infty\mper
\end{equation}

Given samples from $F$ our goal is to recover the parameters that
specify the mixture up to small additive error; this is known as
\emph{parameter distance}. It is easy to see that we can only hope to
recover the components of the mixture up to permutation.  For
simplicity it is convenient to combine the error in estimating the
parameters:

\begin{definition}\label{def:param}
  We say that mixture $\widehat F$ is $\eps$-close to mixture $F$
  if there is a permutation $\pi$ for which
  \[
  \max\left(\| \mu^{(i)}-\widehat\mu^{(\pi(i))}\|_\infty^2, \|\Sigma^{(i)}-\widehat\Sigma^{(\pi(i))}\|_\infty \right)
  \le\eps^2 \Var(F)\mper
  \]
  We say that an algorithm $(\eps,\delta)$-learns a mixture $F$ of
  two gaussians from $f(\eps,\delta)$ samples, if given
  $f(\eps,\delta)$ i.i.d.~samples from~$F$, it outputs a
  mixture~$\widehat F$ that is $\eps$-close to $F$ with probability
  $1-\delta.$
\end{definition}

Note that this definition does not require good recovery of the $p_i$.
If the two components of the mixture are indistinguishable, one cannot
hope to recover the $p_i$ to additive error.  On the other hand, if
the components are well-separated, one can use that the overall mean
is the $p$-weighted average of the component means---or an analogous
statement for the variances---to estimate the $p_i$ from estimates of
the parameters.  Our main theorem will give a more precise
characterization of how well we estimate $p$, but for simplicity we
ignore it in much of the paper.

We also consider learning a mixture of Gaussians component-wise in the
total variation norm.

\begin{definition}
  We say that mixture $\widehat F = \widehat p_1 \widehat G_1 +
  \widehat p_2 \widehat G_2$ is component-wise $\eps$-close to mixture
  $F=p_1G_1+p_2G_2$ in total variation if there is a permutation $\pi$
  for which
  \[
  \max_{i\in\{1,2\}} \mathrm{TV}(G_i,\widehat G_{(\pi(i))}) \le\eps\mper
  \]
  We say that an algorithm $(\eps,\delta)$-learns a mixture $F$ of two
  gaussians in total variation from $f(\eps,\delta)$ samples, if given
  $f(\eps,\delta)$ i.i.d.~samples from~$F$, it outputs a
  mixture~$\widehat F$ that is component-wise $\eps$-close to $F$ in
  total variation with probability $1-\delta.$
\end{definition}

\paragraph{Why parameter distance?}
We believe that proper learning of each component of a Gaussian
mixture in the parameter distance is the most natural
objective. Consider the simple example of estimating the height
distribution of adult men and women from unlabeled population data,
which is well approximated by a mixture of
Gaussians~\cite{humanheight}.  By using parameter distance, our
results give a tight characterization of how precisely you can
estimate the average male and female heights from a given number of
samples.  A guarantee in total variation norm is less easily
interpreted.

Focusing on parameter distance also has technical advantages in our
context.  First, it leads to a cleaner quantitative analysis. Second,
if the covariance matrices are (close to) \emph{sparse} we can recover
only the dominant entries of the covariance matrices and ignore the
rest, decreasing our sample complexity.  An affine invariant measure
such as total variation distance could not benefit from sparsity this
way. Nevertheless, to facilitate comparison with previous work, we
state our results for total variation norm as well.  


%
Finally, it is important that we learn each component rather than the
mixture distribution. Finding a distribution that closely approximates
the mixture distribution is easier but less useful than approximating
the individual components, and is the focus of a different line of
work that we discuss in the related work section. In fact, the
individal parameters are a strong reason for modeling with mixtures of
gaussians in the first place.

%

\subsection{Main results}

\paragraph{One-dimensional algorithm.}
Our main theorem is a general result that achieves tight bounds in
multiple parameter regimes.  As a consequence it's a little cumbersome
to state, so we start with two simpler corollaries.  The first
corollary is that the algorithm $(\eps, \delta)$-learns a mixture with
$O(\eps^{-12} \log(1/\delta))$ samples.
\begin{corollary}\label{cor:e12}
  Let $F$ be any mixture of $1$-dimensional gaussians where $p_1$ and
  $p_2$ are bounded away from zero.  Then
  Algorithm~\ref{alg:1dcombine} can $(\eps, \delta)$-learn $F$ with
  $O(\eps^{-12} \log (1/\delta))$ samples.
\end{corollary}
The $12$th power dependence on $\eps$ arises because our algorithm uses
the $6$th moment.  In fact, we will see that in general this result is
tight: there exist distributions for which one cannot reliably
estimate \emph{either} the $\mu_i$ to $\pm \eps \sigma$ or the $\sigma_i^2$
to $\pm \eps^2 \sigma^2$ with $o(1/\eps^{12})$ samples.

However, for many distributions one can estimate the parameters with fewer
samples.  One important special case is when the two gaussians have means that
are separated by $\Omega(1)$ standard deviations.  In this case, our algorithm
requires only $O(1/\eps^2)$ samples.

\begin{corollary}
  Let $F$ be a mixture of $1$-dimensional gaussians where $p_1$ and
  $p_2$ are bounded away from zero and $\abs{\mu_1 - \mu_2} =
  \Omega(\sigma)$.  Then Algorithm~\ref{alg:1dcombine} can $(\eps,
  \delta)$-learn $F$ with $O(\eps^{-2} \log (1/\delta))$ samples.
\end{corollary}

This result is also tight: even if the samples from the mixture were
labeled, it still would take $\Omega(1/\eps^2)$ samples to estimate the
mean and variance of each gaussian to the desired precision.  Our main
theorem gives a smooth tradeoff between these two corollaries.
\define{thm:1d}{Theorem}{%
  Let $F$ be any mixture of two gaussians with variance~$\sigma^2$ and
  $p_1,p_2$ bounded away from~$0.$ Then, given
  $O(\eps^{-2}n\log(1/\delta))$ samples Algorithm~\ref{alg:1dcombine}
  with probability $1-\delta$ outputs the parameters of a
  mixture~$\wh{F}$ so that for some permutation $\pi$ and all
  $i\in\{1,2\}$ we have the following guarantees:
\begin{itemize}
\item
  If $n \geq \big(\frac{\sigma^2}{|\mu_1-\mu_2|^2}\big)^{6}$, then
  $|\mu_i-\wh{\mu}_{\pi(i)}|\le\epsilon|\mu_1-\mu_2|$,
  $|\sigma_i^2-\wh{\sigma}_{\pi(i)}^2|\le\epsilon|\mu_1-\mu_2|^2$,
  and $\abs{p_i - \wh{p}_{\pi(i)}} \leq \eps$.
\item
  If $n \geq \big(\frac{\sigma^2}{|\sigma_1^2-\sigma_2^2|}\big)^{6}$, then
  $|\sigma_i^2-\wh{\sigma}_{\pi(i)}^2|\le\epsilon|\sigma_1^2-\sigma_2^2| + |\mu_1 - \mu_2|^2$
  and $\abs{p_i - \wh{p}_{\pi(i)}} \leq \eps + \frac{\abs{\mu_1 - \mu_2}^2}{\abs{\sigma_1^2-\sigma_2^2}}$.
\item For any $n \geq 1$, the algorithm performs as well as assuming the
  mixture is a single gaussian: $|\mu_i-\wh{\mu}_{\pi(i)}|\le
  |\mu_1-\mu_2|+\epsilon\sigma$ and
  $|\sigma_i^2-\wh{\sigma}_{\pi(i)}^2|\le|\sigma_1^2-\sigma_2^2|+|\mu_1-\mu_2|^2+\epsilon\sigma^2.$
\end{itemize}
}
\restate{thm:1d}
In essence, the theorem states that the algorithm can distinguish the
two gaussians in the mixture if it has at least
$(\frac{\sigma^2}{\max(\abs{\mu_1 - \mu_2}^2, \abs{\sigma_1^2 -
    \sigma_2^2})})^6$ samples.  Once this happens, the parameters can
be estimated to $\pm \eps$ relative accuracy with only a $1/\eps^2$
factor more samples.  If the means are reasonably separated, then the
first clause of the theorem provides the strongest bounds.  If there
is no separation in the means, we cannot hope to learn the means to
relative accuracy, but we can still learn the variances to relative
accuracy provided that they're separated. This is the content of the
second clause. If neither means nor variances are separated, our
algorithm is no better or worse than treating the mixture as a single
gaussian.

The only assumption present in our main theorem requires that
$\min(p_1,p_2)$ be bounded away from zero. Making this assumption
simplifies the proof on a syntactic level considerably.  A polynomial
dependence on the separation from~$0$ could be extracted from our
techniques, but we don't know if this dependence would be optimal.

\paragraph{Lower bound.}
Our second main result is that the bound in Theorem~\ref{thm:1d} is
essentially best possible among all estimators---even computationally
inefficient ones. More concretely, we exhibit a pair of mixtures
$F,\tilde F$ that satisfy the following strong bound on the squared
Hellinger distance\footnote{For probability measures $P$ and $Q$ with
  densities $p$ and $q,$ respectively, the squared Hellinger distance
  is defined as $\mathrm{\mathrm{H}^2}(P,Q)=\frac12\int
  (\sqrt{p(x)}-\sqrt{q(x)})^2\mathrm{d}x.$} between the two
distributions.

\begin{lemma}
\label{lem:hell}
There are two one-dimensional gaussian mixtures $F,\tilde F$ with
variances $\sigma^2$ and all of the $\mu_i, \sigma_i^2$, and $p_i$
separated by~$\Theta(1)$ from each other such that the squared
Hellinger distance satisfies
\[
\mathrm{H}^2(F,\tilde F) \le O\left(\sigma^{-12}\right)\mper
\]
\end{lemma}
Denoting by $F^{n}$ the distribution obtained by taking~$n$ independent samples
from~$F,$ the squared Hellinger distance satisfies the direct sum rule
$\mathrm{H}^2(F^n,\tilde F^n)\le n\cdot\mathrm{H}^2(F,\tilde F).$ Moreover, if
$\mathrm{H}^2(F^n,\tilde F^n)\le o(1)$ then the total variation distance also
satisfies $\mathrm{TV}(F^n,\tilde F^n)\le o(1).$ In particular, in this case no
statistical test can distinguish $F$ and $\tilde F$ from $n$ samples with high
confidence and parameter estimation is therefore impossible. The following
theorem follows, showing that Corollary~\ref{cor:e12} is optimal.

\define{thm:lower}{Theorem}{%
  Consider any algorithm that, given $n$ samples of any gaussian
  mixture with variance $\sigma^2$, with probability $1-\delta$ learns
  \emph{either} $\mu_i$ to $\pm \eps\sigma$ or $\sigma_i^2$ to $\pm
  \eps^2 \sigma^2$.  Then $n = \Omega(\eps^{-12}\log(1/\delta))$.
} \restate{thm:lower}

Since $(\eps, \delta)$-learning the mixture requires learning
\emph{both} the $\mu_i$ and the $\sigma_i^2$ to this precision, we get
that Corollary~\ref{cor:e12} is tight.  This also justifies our
definition of $\eps$-approximation in parameter distance meaning
approximating the means to $\pm \eps \sigma$ and the variances to $\pm
\eps^2 \sigma_i^2$.

\begin{corollary}
  Any algorithm that uses $f(\eps, \delta)$ samples to $(\eps,
  \delta)$-learn arbitrary mixtures of two $1$-dimensional gaussians
  with $p_1$ and $p_2$ bounded away from zero requires
  $f(\eps, \delta) = \Omega(\eps^{-12}\log(1/\delta))$.
\end{corollary}

We also note that our lower bound technique directly gives a lower
bound of $\Omega(\epsilon^{-6k+2})$ for the problem of learning a
mixture of $k$ Gaussians for constant $k\ge 2$
(Theorem~\ref{thm:lowerkd}). This is incomparable to the lower bound
of, roughly, $\exp(k/24)$ for $\eps < 1/k$ due to~\cite{MoitraV10}.
Our bound is useful when $k$ is a small constant and $\eps$ is going
to zero, while their bound is useful when both $k$ and $1/\eps$ are large.

\paragraph{Upper bound in arbitrary dimensions.}
Our main result holds for the $d$-dimensional problem up to replacing
$\log (1/\delta)$ by $\log (d \log (1/\eps)/\delta)$ in the sample
complexity.

\define{thm:d}{Theorem}{%
  Let $F$ be any mixture of $d$-dimensional gaussians where $p_1$ and
  $p_2$ are bounded away from zero.  Then we can $(\eps,
  \delta)$-learn $F$ with
  $O(\eps^{-12}\log(d\log(1/\epsilon)/\delta))$ samples.
}%
\restate{thm:d}

Notably, our bound is essentially dimension-free and incurs only a
logarithmic dependence on~$d.$ The best previous bound for the problem
is the bound due to~\cite{KalaiMV10} that gives a polynomial
dependence of $O((d/\eps)^{c})$ for some large constant $c$.
The
proof of our theorem is based on a new dimension-reduction technique
for the mixture problem that is quite different from the one
in~\cite{KalaiMV10}.  Apart from the quantitative improvement that it
yields, it is also notably simpler.

\paragraph{Lower bound in higher dimension.}
We can extend our lower bound (Theorem~\ref{thm:lower}) to show that
$\Omega(\eps^{-12} \log (d/\delta))$ samples are necessary to achieve
the guarantee of Theorem~\ref{thm:d}; one can embed a different
instance of the hard distribution in each of the $d$ dimensions, and
the guarantee requires that the algorithm solve all the copies.  That
this direct product is hard is shown in Theorem~\ref{thm:lowerd}.
Hence Theorem~\ref{thm:d} is optimal up to the $\log \log (1/\eps)$
term, and optimal up to constant factors when $d \geq \log (1/\eps)$.

\paragraph{Learning in total variation norm.}

In Section~\ref{sec:TV} we derive various results for learning mixtures of
gaussians in the total variation norm.

\define{thm:tv}{Theorem}{
  Let $F$ be any mixture of $d$-dimensional gaussians where $p_1$ and $p_2$
are bounded away from zero. For any dimension $d\ge1,$ Algorithm~\ref{alg:tvapprox}
  $(\eps, \delta)$-learns $F$ in total variation with
  $O(\eps^{-36}d^{30}\log^6(d/\eps)\log(1/\delta))$ samples.
}
\restate{thm:tv}

While the $d^{30}$ dependence here is probably not close to optimal, the
exponent is nonetheless several orders of magnitude smaller than the exponent of the
polynomial dependence that follows from previous work. Interestingly, this
large sample complexity of the general case can be improved if the covariances
of the two Gaussians have similar eigenvalues and eigenvectors (e.g., they are
isotropic):

\define{thm:tv-isotropic}{Theorem}{
  Let $F$ be any mixture of $d$-dimensional gaussians with covariance
  matrices $\Sigma_1$ and $\Sigma_2$ where the mixing probabilities
  $p_1$ and $p_2$ are bounded away from zero.  Further suppose that
  there exists a constant $C$ such that
  \[
  \Sigma_1 \preceq C\Sigma_2 \preceq C^2\Sigma_1.
  \]
  Then there is an algorithm that can $(\eps, \delta)$-learn $F$ in
  total variation with $O(\eps^{-12}d^{6}
  \log^6(d/\eps)\log(1/\delta))$ samples.
}
\restate{thm:tv-isotropic}

\subsection{Related Work}
The body of related work on gaussian mixture models is too broad to
survey here. We refer the reader to~\cite{KalaiMV10} for a helpful
discussion of work prior to 2010. Since then a number of works have
further contributed to the topic. Moitra and Valiant~\cite{MoitraV10}
gave polynomial bounds for estimating the parameters of a mixture of
$k$ gaussians based on the method of moments. Belkin and
Sinha~\cite{BelkinS10} achieved a similar result. It is an interesting
question if our techniques extend to the case of $k$ gaussians, but
as our lower bounds show the sample size must be at least
$\Omega(\epsilon^{-6k+2})$ which is prohibitive for small $\epsilon$ and even
moderate~$k$.

Work of Chan et al.~\cite{ChanDSS13,ChanDSS14} implies an
\emph{improper} learning algorithm for a mixture of two
single-dimensional gaussians that learns the overall mixture (not the
components) in total variation distance to error~$\epsilon$ using
$\tilde O(1/\epsilon^2)$ samples. An improper learning algorithm in
general does not return a mixture of gaussians nor does it return an
approximation to the individual components of the mixture.

Daskalakis and Kamath~\cite{DaskalakisK14} strengthen this result by giving a
\emph{proper} learning algorithm for learning a one-dimensional mixture with
the same sample complexity. However, unlike our algorithm, it does not learn
the individal components of the mixture. Indeed, this is impossible in general
given the stated sample complexity in light of our lower bound. Nonetheless,
our bounds do imply a proper learning algorithm for the mixture itself (which
is a strictly weaker task than learning both components). In the case where
$d>1,$ our algorithm for learning under total variation norm implies the best
known bounds also for this weaker task when no assumptions are placed on the
mixture.

A number of recent works have considered gaussian mixture models under
stronger assumptions on the components. See, for example,
\cite{HsuK13,AcharyaJOS14}. We are not aware of improvements
over~\cite{KalaiMV10} for the parameter estimation problem when no such
assumptions are made.

\iflong
\subsection{Proof overview}
\fi

We now give a high-level outline of our algorithmic approach (and the related
approach of Pearson). The starting point for the method of
moments is to set up a system of polynomial equations whose
coefficients are determined by the moments of the mixture and whose variables
are the unknown parameters. Solving the system of polynomial equations
recovers the unknown parameters. The main stumbling block is that the
roots of polynomials are notoriously unstable with respect to
small perturbations in the coefficients. A famous example is Wilkinson's
polynomial. Perturbations arise inevitably in our context because we
do not know the moments of the mixture model exactly but rather need to
estimate them empirically from samples. Our main contribution is to exhibit a
robust set of polynomial equations from which the parameters can be recovered.
We hope that similar techniques may be useful in extending our results to
other settings such as learning a mixture of more than two gaussians.

\paragraph{Reparametrization.}

We begin by reparametrizing the gaussian mixture
in such a way to get parameters that are independent of adding
gaussian noise to the mixture. Formally, adding or subtracting the
same term from each of the variances leaves these parameters
unchanged.  Assuming the overall mean of the mixture is~$0$, this
leaves us with $3$ free parameters that we call $\alpha,\beta,\gamma.$
Since these parameters are independent of adding gaussian noise it is
useful to also define the moments of the mixture in such a way that
they are independent under adding gaussian noise. This is accomplished
by considering what we call \emph{excess moments}. The name is
inspired by the term \emph{excess kurtosis}, a well-known measure of
``peakedness'' of a distribution introduced in~\cite{Pearson} that
corresponds to the fourth excess moment.  At this point, the third
through sixth excess moments give us four equations in the three
variables $\alpha, \beta, \gamma$.

\paragraph{Three different precision regimes.}
Our analysis distinguishes between three different parameter
regimes. In the first parameter regime we know each excess moment
$X_i$ for $i\le 6$ up to an additive error of
$\epsilon|\mu_1-\mu_2|^i.$ This analysis is applicable when the means
are separated and it leads to the first case in
Theorem~\ref{thm:1d}. The second regime is when the separation between
the means is small, but we nevertheless know each excess moment up to
error $\epsilon|\sigma_1^2-\sigma_2^2|^{i/2}.$ This analysis in this
case applies when the variances are separated and leads to the second
case in Theorem~\ref{thm:1d}. Finally, when neither of the cases
applies the two gaussians are indistinguishable and we simply fit a
single gaussian.  We show that we can figure out which parameter
regime we're in and run the appropriate algorithm.

We focus here on a discussion of the first parameter regime, since it is the
most interesting case. The full argument is in Section~\ref{sec:precision-mu}.

\paragraph{Robustifying Pearson's polynomial.} Expressing the excess moments in
terms of our new parameters $\alpha,\beta,\gamma,$ we can derive in a
fairly natural manner a ninth degree polynomial $p_5(y)$ whose
coefficients depend on $X_3, X_4,$ and $X_5$ so that $\alpha$ has to
satisfy $p_5(\alpha)=0.$ The polynomial $p_5$ was already used by
Pearson. Unfortunately, $p_5$ can have multiple roots and this is to
be expected since $5$ moments are not sufficient to identify a mixture
of two gaussians.  Pearson computed the mixtures associated with each
of the roots and threw out the invalid solutions (e.g. the ones that
give imaginary variances), getting two valid mixtures that matched on
the first five moments.  He then chose the one whose sixth moment was
closest to the observed sixth moment.

We proceed somewhat differently from Pearson after computing $p_5$.
First, we use the first 4 excess moments to compute an upper bound
$y_{max}$ on $\alpha$. We show that the set of valid mixtures that
match the first 5 moments correspond precisely to the roots $y$ of
$p_5(y)$ with $0 < y \leq y_{max}$. We then derive another ninth
degree polynomial using $X_3, X_4, X_5,$ and $X_6$ that we call
$p_6(y).$ We prove that $\alpha$ is the only solution to the system of
equations
\[
\big\{p_5(y)=0,\quad p_6(y)=0,\quad 0<y\leq y_{\mathrm{max}}\big\}\mper
\]
This approach isn't yet robust to small perturbations in the moments;
for example, if $p_5$ has a double root at $\alpha$, it may have no
nearby root after perturbation. We therefore consider the polynomial
$r(y) = p_5^2(y)+p_6^2(y)$ which we know is zero at $\alpha.$ We argue
that $r(y)$ is significantly nonzero for any $y$ significantly far
from $\alpha$.  This is the main technical claim we need.

For intuition of why this is the case, consider the normalization
$\abs{\mu_1 - \mu_2} = 1$ and the setting where $\abs{\sigma_1^2 -
  \sigma_2^2} = O(1)$.  Because the excess moments are polynomials in
$\alpha, \beta, \gamma$ we can think of $r(y)$ as a polynomial in $(y,
\alpha, \beta, \gamma)$.  We are interested in some region $R \subset
\R^4$ where every root of $r$ corresponds to a mixture matching the
first six moments.  Because six moments suffice to identify the
mixture by~\cite{KalaiMV10}, $r$ has no roots in $R$ outside $y = \alpha$.
This lets us show that $r / (y - \alpha)^2$ has no roots over $R$,
which for a compact $R$ implies that $r / (y - \alpha)^2 = \Omega(1)$
over $R$.  Thus $r = \Omega((y - \alpha)^2)$ over the region of
interest.

Now, with $O(\sigma^{12}/\eps^2)$ samples we can estimate all the
$X_i$ to $\pm \eps$, which lets us estimate both $p_5(y)$ and $p_6(y)$
to $\pm O(\eps)$.  This means $\sqrt{r(y)}$ is estimated to $\pm
O(\eps)$.  Since $\sqrt{r(y)} = \Omega(\abs{y - \alpha})$, this lets
us find $\alpha$ to $\pm O(\eps)$.  We then work back through our
equations to get $\beta$ and $\gamma$ to $\pm O(\eps)$, which give the
$\mu_i$ and $\sigma_i^2$ to $\pm O(\eps)$.

The analysis proceeds slightly differently in the setting where
$\abs{\sigma_1^2 - \sigma_2^2} \gg 1$.  In this setting the region $R$
of interest is not compact, because the parameter $\gamma$ (which here
equals $\sigma_1^2 - \sigma_2^2$) is unbounded.  However, we can show
directly that the highest (12th) degree coefficient of $\gamma$ in $r
/ (y - \alpha)^2$ is bounded away from zero, getting that $r =
\Omega(\gamma^{12}(y - \alpha)^2)$.  Since the $X_i$ are now not
constant, while we can estimate each $X_i$ to $\pm \eps$ with
$O(\sigma^{12}/\eps^2)$ samples, we only estimate $p_5(y)$ and
$p_6(y)$ to $\pm O(\eps \gamma^5)$.  Since $\sqrt{r(y)} =
\Omega(\gamma^6\abs{y-\alpha})$, this lets us estimate $\alpha$ to
$\pm O(\eps /\gamma)$.  This is sufficient to recover $\gamma$ to $\pm
O(\eps)$, which lets us recover $\beta$ to $\pm O(\eps)$ and then the
$\mu_i$ and $\sigma_i^2$ to $\pm O(\eps)$.

\paragraph{Dimension Reduction.} In Section~\ref{sec:dimension} we
extend our theorem to arbitrary dimensional mixtures using two simple
ideas. The first idea is used to reduce the $d$-dimensional case to
the $4$-dimensional case and is straightforward. The second argument
reduces the $4$-dimensional case to the $1$-dimensional and is only
slightly more involved. How can we use an algorithm for $d\le 4$ to
solve the problem in arbitrary dimension? Consider the case where
$\Sigma_1,\Sigma_2\in\R^{d\times d}$ differ in some entry $(i,j)$. We
can find $(i,j)$ by running our assumed algorithm for all pairs of
variables. Each pair of variables leads to a two-dimensional mixture
problem where the covariances are obtained by restricting
$\Sigma_1,\Sigma_2$ to the corresponding entries. Once we have found
an entry $(i,j)$ where $|(\Sigma_1-\Sigma_2)_{ij}|>\epsilon,$ we are
in good shape. We now iterate over all $k,l\in[d]$ and solve the
$4$-dimensional mixture problem on the variables $(i,j,k,l)$ to within
accuracy $\epsilon/10.$ This not only reveals an additional entry
$k,l$ of the covariance matrix but it also tells us which of the two
values for position $(k,l)$ is associated with which of the values for
position $(i,j).$ This is because we solved the $4$-dimensional
problem to accuracy $\epsilon/10$ and we know that
$|(\Sigma_1-\Sigma_2)_{ij}|>\epsilon.$ Hence, each newly recovered
value for position $(i,j)$ must be close to the value that we
previously recovered. This ensures that we do not mix up any entries
and so we recover the covariance matrices entry by entry. A similar but simpler argument works for the means.

Finally, the four-dimensional problem reduces to one dimension by
brute forcing over an $\eps$-net of all possible four-dimensional
solutions (which is now doable in polynomial time) and using the
algorithm for $d=1$ to verify whether we picked a valid solution. The
verification works by projecting the four-dimensional mixture in a
random direction. Using anti-concentration results for quadratic forms
in gaussian variables, we can show that any covariance matrix
$\eps$-far from the true covariance matrices will be ruled out with
constant probability by each projection.  Therefore $O(\log(1/\eps))$
projections will identify the covariance matrices among the
$\poly(1/\eps)$ possibilities.  A union bound requires $\delta \approx
1/ \log (1/\eps)$, giving $O(\log \log (1/\eps))$ overhead beyond the
$1$-dimensional algorithm.

\section{Lower bounds}\label{sec:lower}

\subsection{Mixtures with matching moments are very close under Gaussian noise}

\begin{figure*}[h]
\begin{center}
\includegraphics[width=0.95\textwidth]{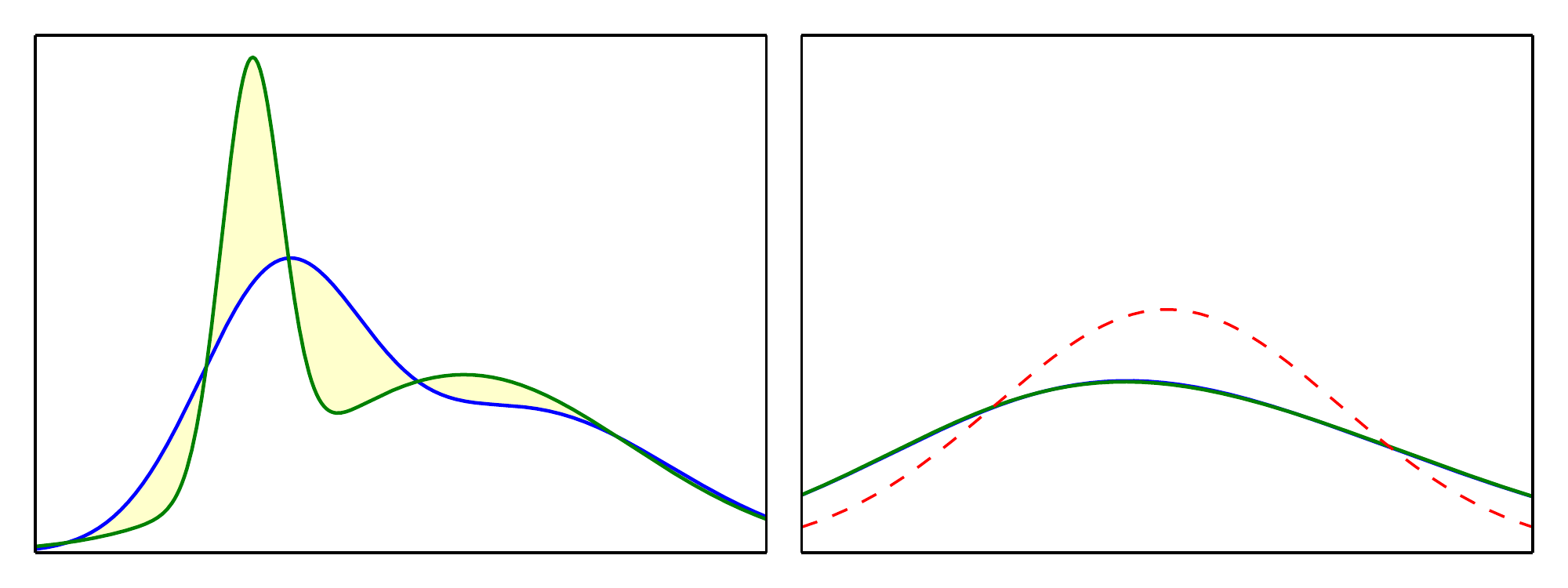}
\end{center}
\caption{Mixtures with matching first $5$ moments before and after
adding $N(0,2)$ (red).}
\label{fig:noise}
\end{figure*}

Our main lemma shows that if we have two gaussian mixtures whose first
$k$ moments are matching and we add a gaussian random variable
$N(0,\sigma^2)$ to each mixture, then the resulting distributions are
$O(1/\sigma^{2k+2})$-close in squared Hellinger distance. The idea is illustrated
in Figure~\ref{fig:noise}.
\begin{definition}
  Let $P,Q$ be probability distributions that are absolutely
  continuous with respect to the Lebesgue measure. Let $p$ and $q$
  denote density functions of $P$ and $Q,$ respectively. Then, the
  \emph{squared Hellinger distance} between $P$ and $Q$ is defined as
\[
\mathrm{\mathrm{H}^2}(P,Q)=\frac12\int_{-\infty}^{\infty} \left(\sqrt{p(x)}-\sqrt{q(x)}\right)^2\mathrm{d}x
\mper
\]
\end{definition}

\begin{lemma}\label{l:h2}
  Let $F$ and $G$ be distributions that are subgaussian with constant
  parameters and identical first $k$ moments for $k= O(1)$. Let $P = F
  + N(0, \sigma^2)$ and $Q = G + N(0, \sigma^2)$ for $\sigma \gtrsim
  1$.  Then
  \[
  \mathrm{H}^2(P,Q) \lesssim 1/\sigma^{2k+2}.
  \]
\end{lemma}
\iflong
\begin{proof}
  We have that $F$ and $G$ are subgaussian with constant parameters,
  i.e., for any $d \geq 0$ we have
  \[
  \tabs{\E_{t \sim F} t^d} \lesssim (O(\sqrt{d}))^{d}
  \]
  and similarly for $G$.  Denote by $p,q,f,g$ density functions of
  $P,Q,F,G,$ respectively. We would like to bound
  \begin{align}\label{eq:Hpq}
    \mathrm{H}^2(P,Q) = \frac12\int_{-\infty}^\infty \left(\sqrt{p(x)} -
\sqrt{q(x)}\right)^2\mathrm{d}x\mper
  \end{align}
  We split the integral~\eqref{eq:Hpq} into two regimes, $\abs{x} \geq T$
  and $\abs{x} \leq T$ for $T \eqsim \sigma \sqrt{\log \sigma}$.

For the $\abs{x} \geq T$ regime, we have
\begin{align*}
  \frac12\int_{\abs{x} \geq T} \left(\sqrt{p(x)} - \sqrt{q(x)}\right)^2\mathrm{d}x
& \leq \Pr_{x \sim P}\Set{\abs{x} \geq T} + Pr_{x \sim Q}\Set{\abs{x} \geq
T}\\
& \lesssim e^{-T^2/(2(\sigma^2 + O(1)))} \\
& \lesssim 1/\sigma^{2k+2}\mper
\end{align*}
The challenging part is the $\abs{x} \leq T$ regime.

\begin{claim}
  For $\abs{x} \leq T,$ we have
  \begin{align}\label{eq:plower}
    p(x) \gtrsim \frac{1}{\sqrt{2\pi}\sigma}e^{-x^2/(2\sigma^2)}.
  \end{align}
\end{claim}
\begin{minipage}{0.85\linewidth}
\begin{proof}
  Let $x$ be such that $\abs{x}\le T.$ Let $t$ be such that
  $F([-t,t])=1/2.$ Note that $t=O(1)$ since all the parameters of~$F$
  are constant.  In particular, denoting by $\nu(y)$ the density of
  $N(0,\sigma^2)$ we have for every $y\in[-t,t],$
  \[
  \nu(x-y)\ge
  \frac{e^{-(\abs{x}+t)^2/2\sigma^2}}{\sqrt{2\pi}\sigma}
  \ge
  \frac{e^{-x^2/2\sigma^2}e^{-O(\abs{x}/\sigma^2)}e^{-O(1/2\sigma^2)}} {\sqrt{2\pi}\sigma}
  \gtrsim \frac{1}{\sqrt{2\pi}\sigma}e^{-x^2/(2\sigma^2)}.
  \]
  Hence,
  \[
  p(x) \ge\int_{-t}^t f(x)\nu(x-y)\mathrm{d}y
  \ge \frac12\cdot\min_{y\in[-t,t]}\nu(x-y)
  \gtrsim \frac12\cdot \frac{1}{\sqrt{2\pi}\sigma}e^{-x^2/(2\sigma^2)}\mper
  \]
\end{proof}
\end{minipage}
\vspace{1em}

\noindent Now, we define
  \[
  \Delta(x) = \frac{q(x) - p(x)}{p(x)}.
  \]
  We have that
  \begin{align}
    \Delta(x) &= \frac{1}{p(x)}\int_{-\infty}^{\infty} \frac{1}{\sqrt{2\pi}
\sigma} e^{-\frac{(x-t)^2}{2\sigma^2}} (g(t)  - f(t))\mathrm{d}t\notag\\
    &= \frac{1}{p(x)}\frac{1}{\sqrt{2\pi} \sigma}
e^{-x^2/(2\sigma^2)}\int_{-\infty}^{\infty}
e^{tx/\sigma^2}e^{-\frac{t^2}{2\sigma^2}} (g(t)  - f(t))\mathrm{d}t\notag\\
    &\lesssim \int_{-\infty}^{\infty}
e^{tx/\sigma^2}e^{-\frac{t^2}{2\sigma^2}} (g(t)  - f(t))\mathrm{d}t\label{eq:deltaintegral}.
  \end{align}
  We take a power series expansion of the interior of the integral,
  \begin{align*}
    e^{tx/\sigma^2}e^{-\frac{t^2}{2\sigma^2}}
    &= \left(\sum_{d=0}^\infty
\frac{(tx/\sigma^2)^d}{d!}\right)\left(\sum_{d=0}^\infty
\frac{(-t^2/(2\sigma^2))^d}{d!}\right)\\
    &= \sum_{d=0}^\infty \sum_{\substack{j \in \Z\\0 \leq 2j \leq  d}}
\frac{(tx/\sigma^2)^{d-2j}}{(d-2j)!}\frac{(-t^2/(2\sigma^2))^{j}}{j!}\\
    &= \sum_{d=0}^\infty (t/\sigma)^d\sum_{\substack{j \in \Z\\0 \leq 2j \leq
d}} \frac{(x/\sigma)^{d-2j}(-1/2)^{j}}{(d-2j)!j!}.
  \end{align*}
  Now, for all $j \in [0, d/2]$,
  \[
  (d-2j)!j! \geq (d/2 - \sqrt{d})! =
  (\Omega(d))^{d/2 - \sqrt{d}} = (\Omega(d))^{d/2}\mper
  \]
Therefore each term in the inner sum has magnitude bounded by
$(x/\sigma)^{d-2j}(O(1/d))^{d/2}$, so the sum has magnitude bounded by
$(1 + (x/\sigma)^d)(O(1/d))^{d/2}$.  Hence there exists a constant $C$
and values $c_{x,d}$ with $\abs{c_{x, d}} \leq 1$ such that
  \[
  e^{tx/\sigma^2}e^{-\frac{t^2}{2\sigma^2}} = \sum_{d=0}^\infty
  c_{x,d}\left(\frac{C(1 + x/\sigma)}{\sigma\sqrt{d}}\right)^dt^d.
  \]
  Returning to~\eqref{eq:deltaintegral}, we have
  \begin{align*}
    \Delta(x) &\lesssim \int_{-\infty}^\infty \sum_{d=0}^\infty
c_{x,d}\left(\frac{C(1 + x/\sigma)}{\sigma\sqrt{d}}\right)^dt^d (g(t) -
f(t))\mathrm{d}t\\
    &= \sum_{d=0}^\infty c_{x,d}\left(\frac{C(1 +
x/\sigma)}{\sigma\sqrt{d}}\right)^d\int_{-\infty}^\infty t^d (g(t) -
f(t))\mathrm{d}t\\
    &\leq \sum_{d=0}^\infty \abs{c_{x,d}}\left(\frac{C(1 +
x/\sigma)}{\sigma\sqrt{d}}\right)^d\left\{
      \begin{array}{cl}
        0 & \text{if } d \leq k\\
        (O(\sqrt{d}))^{d} & \text{otherwise}.
      \end{array}\right.\\
    &\lesssim \sum_{d=k+1}^{\infty} \left(\frac{O(1) \cdot C(1 +
x/\sigma)}{\sigma}\right)^d\\
    &\eqsim \left(\frac{1 + x/\sigma}{\sigma}\right)^{k+1}.
  \end{align*}
  for all $\abs{x} \leq T \eqsim \sigma \log \sigma$.  Note that this
  implies $\abs{\Delta(x)} < 1$ which justifies the expansion
  \[
  \sqrt{1 + \Delta(x)} = 1 + \Delta(x)/2 \pm O(\Delta(x)^2).
  \]
  Therefore, following the approach outlined in~\cite{Pollard}, we can write
  \begin{align*}
    \mathrm{H}^2(P,Q) &= 1 - \int_{-\infty}^{\infty} \sqrt{p(x)q(x)}\mathrm{d}x\\
    &= 1 - O(1/\sigma^{2k+2}) - \int_{-T}^{T} p(x)\sqrt{1 +
\Delta(x)}\mathrm{d}x\\
    &= 1 - O(1/\sigma^{2k+2}) - \int_{-T}^{T} p(x)(1 + \Delta(x)/2 \pm
    O(\Delta(x)^2))\mathrm{d}x
  \end{align*}
  Now, we have that
  \begin{align*}
    \int_{-T}^{T} p(x)\left(1 + \frac{1}{2}\Delta(x)\right) \mathrm{d}x 
&= \Pr_{x \sim P}[\abs{x} \leq T] + \frac{1}{2}\left(\Pr_{x \sim Q}\Set{\abs{x} \leq T} 
- \Pr_{x \sim P}\Set{\abs{x} \leq T]}\right)\\
    &= 1 - \frac{1}{2}\left(\Pr_{x \sim Q}\Set{\abs{x} > T} + \Pr_{x \sim
P}\Set{\abs{x} > T}\right) \\
    &= 1 - O(1/\sigma^{2k+2})
  \end{align*}
  by our choice of $T.$ We conclude,
  \begin{align*}
    \mathrm{H}^2(P, Q) 
&\lesssim 1/\sigma^{2k+2} + \int_{-T}^T p(x) \left(\frac{1 +
x/\sigma}{\sigma}\right)^{2k+2} \mathrm{d}x\\
    &\lesssim 1/\sigma^{2k+2}\left(1 + \int_{-T}^T p(x)
(x/\sigma)^{2k+2}\mathrm{d}x\right)\\
    &\leq 1/\sigma^{2k+2}\left(1 + \E_{x \sim P} (x/\sigma)^{2k+2}\right)\\
    &\lesssim 1/\sigma^{2k+2}\mper
  \end{align*}
\end{proof}
\fi

\subsection{Lower bounds for mixtures of two Gaussians}

\begin{claim}\label{c:h2lower}
  Let $P$ and $Q$ be distributions with $\mathrm{H}^2(P, Q) \leq
  \eps$.  Then there exists a constant $c > 0$ such that
  $n=c\eps^{-1}\log(1/\delta)$ independent samples from $P$ and $Q$
  have total variation distance less than~$1-\delta.$ In particular,
  we cannot distinguish the distributions from $n$ samples with
  success probability greater than $1-\delta.$
\end{claim}
\begin{proof}
  Let $x_1, \dotsc, x_n \sim P$ and $y_1, \dotsc, y_n \sim Q$ for $n =
  c\eps^{-1}\log(1/\delta)$.  We will show that the 
total variation distance between $(x_1, \dotsc, x_n)$ and $(y_1, \dotsc, y_n)$ is
  less than $1-\delta$.

  We partition $[n]$ into $k$ groups of size $1/(10\eps)$, for $k =
  10c\log(1/\delta)$.  Within each group, by sub-additivity of squared
  Hellinger distance we have that
  \[
  \mathrm{H}^2\big( (x_1, \dotsc, x_{1/(10\eps)}), (y_1, \dotsc,
y_{1/(10\eps)})\big) \leq 1/10\mper
  \]
Appealing to the relation between total variation and Hellinger, this implies
\iflong
  \[
  \mathrm{TV}\big( (x_1, \dotsc, x_{1/(10\eps)}), (y_1, \dotsc,
y_{1/(10\eps)})\big) 
\le 2 \mathrm{H}\big( (x_1, \dotsc, x_{1/(10\eps)}), (y_1, \dotsc,
y_{1/(10\eps)})\big) \leq \frac23\mper
  \]
\fi
  Hence we may sample $(x_1, \dotsc, x_{1/(10\eps)})$ and $(y_1, \dotsc,
  y_{1/(10\eps)})$ in such a way that the two are identical with
  probability at least $1/3$.  If we do this for all $k$ groups, we
  have that $(x_1, \dotsc, x_n) = (y_1, \dotsc, y_n)$ with probability
  at least $1/3^k > 2\delta$ for sufficiently small constant $c$.
\end{proof}

\state{thm:lower}
\begin{proof}
  Take any two gaussian mixtures $F$ and $G$ with constant parameters
  such that the four means and variances are all $\Omega(1)$ different
  from each other, but $F$ and $G$ match in the first five moments.
  One can find such mixtures by taking almost any mixture $F$ with
  constant parameters and solving $p_5$ to find another root and the
  corresponding mixture (per Lemma~\ref{l:solutionworks}, this will
  cause the first five moments to match).  We can find such an $F$ and
  $G$ in~\cite{Pearson}, or alternatively take
  \begin{align}
    F &= \frac{1}{2}N(-1, 1) + \frac{1}{2}N(1, 2) \notag\\
    G &\approx 0.2968 N(-1.2257, 0.6100) + 0.7032 N(0.5173, 2.3960). \label{eq:FG}
  \end{align}
  While $G$ is expressed numerically, one can certainly prove that the
  $p_5$ derived from $F$ has a second root that yields something close
  to this mixture.  Plug the mixtures into Lemma~\ref{l:h2}.  We get
  that for any $\sigma > 0$, the mixtures
  \iflong
  \begin{align*}
    P &= \frac{1}{2}N(-1, 1 + \sigma^2) + \frac{1}{2}N(1, 2 + \sigma^2)\\
    Q &\approx 0.2968 N(-1.2257, 0.6100 + \sigma^2) + 0.7032 N(0.5173, 2.3960 + \sigma^2).
  \end{align*}
  \fi
  have
  \begin{align}
    \mathrm{H}^2(P, Q) \lesssim 1/\sigma^{12}.\label{eq:h2pq}
  \end{align}
  Since by Claim~\ref{c:h2lower} we cannot differentiate $P$ and $Q$
  with $o(\sigma^{12}\log (1/\delta))$ samples, it requires
  $\Omega(\sigma^{12}\log (1/\delta))$ samples to learn either the
  $\mu_i$ or the $\sigma_i^2$ to $\pm 1/10$ with $1-\delta$
  probability.  Set $\sigma = 1/(10\eps)$ to get the result.
\end{proof}

Our argument extends to $d$ dimensions. We gain a $\log(d)$ factor in our
lower bound by randomly planting a hard mixture learning problem in each of
the~$d$ coordinates.

\begin{claim}\label{c:h2lowerd}
  Let $P$ and $Q$ be distributions with $\mathrm{H}^2(P, Q) \leq \eps$.  Let
  $R_i \in \{P, Q\}$ uniformly at random for $i \in \{1,\dots,d\}$.
  Then there exists a constant $c > 0$ such that
  given $n=c\eps^{-1}\log(d/\delta)$ no algorithm can 
identify all $R_i$ with probability $1-\delta.$
\end{claim}
\begin{proof}
  As in Claim~\ref{c:h2lower}, we have that the total variation
  distance between $B = 1/(10\eps)$ samples from $P$ and $B$ samples
  from $Q$ is less than $2/3$.

  Partition our samples into $k = 10c \log d$ groups $x^1, \dotsc,
  x^k$, where for each group $j \in [k]$ and coordinate $i \in [d]$ we
  have $x^j_i \sim R_i^{\otimes B}$.  By the total variation
  bound between $P^{\otimes B}$ and $Q^{\otimes B}$, we could instead
  draw $x^j_i$ from a distribution independent of $R_i$ with
  probability $1/3$ and a distribution dependent on $R_i$ with
  probability $2/3$.  Suppose we do this.

  Then for any coordinate $i$, with probability $3^{-k} > \delta/d$
  all of $x^1_i, \dotsc, x^k_i$ are independent of $R_i$.  Since
  the coordinates are independent, this means that with probability at
  least
  \[
  1 - (1 - \delta/d)^d \geq \delta/4
  \]
  there will exist a coordinate $i$ such that all of $x^1_i, \dotsc,
  x^k_i$ are independent of $R_i$.  The algorithm must then 
guess~$R_i$ incorrectly with probability at least $1/2$, for 
a~$\delta/8$ probability of failure overall. 
Rescale $\delta$ to get the result.
\end{proof}

This immediately gives that Theorem~\ref{thm:lower} can be extended to
$d$ dimensions:
\begin{theorem}\label{thm:lowerd}
  Consider any algorithm that, given $n$ samples of any
  $d$-dimensional gaussian mixture $F$ with $\Var(F) = \sigma^2$, with
  probability $1-\delta$ for all $i \in [d]$ learns either $\mu_i$ to
  $\pm \eps\sigma$ or $\Sigma_{i,i}$ to $\pm \eps^2 \sigma^2$.  Then
  $n = \Omega(\eps^{-12}\log(d/\delta))$.
\end{theorem}
\begin{proof}
  Let $P, Q$ be as in Theorem~\ref{thm:lower}.  We choose a mixture
  $F$ to have independent coordinates, each of which is uniformly
  chosen from $\{P, Q\}$.  Then $\Var(F) \eqsim 1/\eps$, $H(P, Q)
  \leq \eps^{6}$, and learning the parameters of the mixture in the
  $i$th coordinate to the specified precision would identify whether
  it is $P$ or $Q$.  Claim~\ref{c:h2lowerd} gives the result.
\end{proof}

\iflong

\subsection{Lower bound for mixtures of $k$ gaussians}

We can extend these lower bounds to mixtures of $k$ gaussians.  The
main issue is that for $k=2$ we know that two explicit well-separated
mixtures exist that match on $5$ moments, by solving the method of
moments on a random input and getting two solutions.  For general $k$
we would like to show the existence of two well-separated mixtures that
match on nearly $3k$ moments.

We will formalize the following intuition.  With $k$ gaussians there
are $3k-1$ free parameters-- means, variances, and probabilities
subject to the probabilities summing to one.  Therefore if we only
take $3k-2$ moments, we are embedding a high dimensional space into a
lower dimensional space, and expect lots of collisions.  Therefore
some pair that collide should be well separated.

First we show a few lemmas that will be useful.  We use the following
fact, shown in the appendix:

\define{lem:formalanticoncentration}{Lemma}{%
  Let $p(x) = p(x_1, \dotsc, x_n)$ be a multivariate polynomial of degree $d$
  and smallest nonzero coefficient of magnitude $a$.  Then
  \[
  \Pr_{x \sim N(0, I_n)} [ \abs{p(x)} < a \eps] < d\eps^{1/d}.
  \]
}
\restate{lem:formalanticoncentration}

This lets us show the following:

\begin{lemma}\label{lem:Anonsingular}
  Let $k \geq 1$ be a constant.  There exists a set of $k$ gaussians
  $N(\mu_i, \sigma_i^2)$ such that all the $\mu_i$ and $\sigma_i$ are
  $O(1)$, all $\mu_i$ and $\sigma_i$ are $\Omega(1)$ far from the
  others and from zero, and the matrix $A \in \R^{m \times k}$ of
  moments, given by
  \[
  A_{j, i} = \E_{x \sim N(\mu_i, \sigma_i^2)} x^j,
  \]
  has minimum singular value $\Omega(1)$ for any $m \geq k$.
\end{lemma}
\begin{proof}
  We will show this to be true with good probability for a randomly
  drawn set of $k$ gaussians.  We consider drawing the gaussians
  randomly, so $\mu_i \sim N(0, 1)$ and $\sigma_i \sim N(0, 1)^2$.
  This immediately gives the first two properties with probability
  arbitrarily close to $1$, and we just need to bound the minimum
  singular value of $A$.  We can assume without loss of generality
  that $m = k$, since this minimizes the singular values.  Then, since
  $A$ has $O(1)$ dimension and coefficients, it suffices to show that
  the determinant of $A$---which is the product of the
  eigenvalues---is $\Omega(1)$ with positive probability.

  Now, consider the determinant of $A$ as a formal polynomial in the
  $\mu_i$ and $\sqrt{\sigma_i}$.  We know that this is nonzero,
  because (for example) the $\mu_1^1\mu_2^2\mu_3^3\dotsb\mu_k^k$
  monomial appears only from the diagonal term.  It is a fixed
  monomial, so its minimum coefficient is some constant.  Hence
  Lemma~\ref{lem:formalanticoncentration} shows for $\mu_i,
  \sqrt{\sigma_i} \sim N(0, 1)$ that the determinant will be
  $\Omega(1)$ with probability arbitrarily close to $1$.  In such
  cases, the minimum singular value is $\Omega(1)$ as well, giving the
  result.
\end{proof}

\begin{lemma}\label{lem:FGk}
  Let $k \geq 1$ be a constant.  There exist two mixtures $F$ and $G$ of
  $k$ gaussians each that match on the first $3k-2$ moments, for which
  all the parameters $(p_i, \mu_i, \sigma_i)$ are bounded by $O(1)$
  for each mixture, and for which one of the mixtures has either a
  $\mu_i$ or a $\sigma_i$ that is $\Omega(1)$ far from any $\mu_i$ or
  $\sigma_i$ in the other mixture.
\end{lemma}
\begin{proof}
  There are $t = 3k - 1$ ``free'' parameters in a mixture of $k$
  gaussians: the means, variances, and relative probabilities subject
  to the sum of probabilities equalling $1$.  With $t$ parameters, we
  expect there to be \emph{lots} of mixtures that match on $t-1$
  moments.

  Formally, consider the $(\mu_i, \sigma_i)$ given by
  Lemma~\ref{lem:Anonsingular}.  For the matrix of moments $A$ as
  given by the lemma, and for any vector $p \in \R^k$ of probabilities
  for each gaussian, the first $m$ moments of the mixture $(p_i,
  \mu_i, \sigma_i)$ are precisely $Ap$.  Let $p_i = 1/k$ for all $i$.

  Now, let $0 < \alpha < \beta \leq \frac{1}{k^2}$ be small constants
  to be determined later.  For any set of free parameters $z \in
  \R^t$, consider a mixture of gaussians $f(z) = \{(p'_i, \mu'_i,
  \sigma'_i)\}$ given by $\mu'_i = \mu_i + \alpha z_i$, $\sigma'_i =
  \sigma_i + \alpha z_{k+i}$, and $p'_i = p_i + \beta z_{2k+i}$ for $i
  \in [k-1]$ and $p'_k = p_k - \sum_{i \in [k-1]} \beta z_{2k+i}$.
  For every $z$ with $\norm{z}_2 = 1$, this is a valid mixture of $k$
  gaussians (because $\beta$ is small enough that $\sum_{i \in [k-1]}
  \beta z_{2k+i} < p_k$).  For such a $z$, define $A(z)\in \R^{m
    \times k}$ to be the matrix of component moments, $A(z)_{j, i} =
  \E_{x \sim N(\mu'_i, \sigma'_i)} x^j$, and $p(z) \in \R^k$ to be
  the vector of probabilities $p'$.  Then the first $m$ moments of the
  mixture $f(z)$ are $M(z) = A(z) p(z)$.

  Note that, for any $\alpha, \norm{z} \leq 1$, we have $\norm{A(z) -
    A(0)} \lesssim \alpha \norm{z}$: every monomial in the matrix
  $A(z) - A(0)$ includes $\alpha$ times a coordinate of $z$ in it, and
  there are only a constant number of monomials.

  Because $M(z)$ is a polynomial in $z$, it is continuous.  If $m = t
  - 1 = 3k-2$, then the Borsuk-Ulam theorem states that there exist
  two antipodal points $z, -z$ with $\norm{z}_2 = 1$ such that $M(z) =
  M(-z)$.  This immediately gives two different mixtures that match on
  the first $m$ moments.  This is almost what we want, but we also
  need the mixtures to be different over $\sigma$ or $\mu$, not just
  $p$.

  We have that $A(z)p(z) = A(-z)p(-z)$, or
  \[
  A(0)(p(z) - p(-z)) = (A(-z) - A(0)) p (-z) - (A(z) - A(0))p(z)
  \]
  For $\sigma_{min} = \Omega(1)$ being the minimal singular value of
  $A(0)$, we have
  \begin{align*}
    \sigma_{min}\norm{p(z) - p(-z)}_2 &\leq \norm{A(0)(p(z) - p(-z))} = \norm{(A(-z) - A(0)) p (-z) - (A(z) - A(0))p(z)}\\
    \norm{p(z) - p(-z)}_2 &\leq \frac{1}{\sigma_{min}} (\norm{A(-z) -
      A(0)} + \norm{A(z) - A(0)})(\norm{p(-z)} + \norm{p(z)})\\
    &\leq \frac{1}{\Omega(1)} \cdot O(\alpha) \cdot 2 \lesssim \alpha.
  \end{align*}
  On the other hand, $\norm{p(z) - p(-z)}_2 \geq \beta
  \sqrt{\sum_{i > 2k} z_i^2}$, so
  \[
  \sqrt{\sum_{i > 2k} z_i^2} \lesssim \frac{\alpha}{\beta}.
  \]
  If we set $\beta$ to some value (e.g. $1/k^2$), then if we choose
  $\alpha$ as a small enough constant we will have $\sqrt{\sum_{i >
      2k} z_i^2} \leq 1/2$.  Since $\norm{z}_2 = 1$, this means
  $\sqrt{\sum_{i \leq 2k} z_i^2} \geq \sqrt{3/4} \gtrsim 1$.
  Therefore the mixtures $f(z)$ and $f(-z)$ have at least one of their
  $\mu_i$ or $\sigma_i$ perturbed by $\Theta(\alpha)$.  For small
  enough constant $\alpha$, the perturbations will be much less than
  the $\Omega(1)$ gap between all the different $\mu_i$ and $\sigma_i$
  in the mixture $f(0)$.  Therefore the perturbed $\mu_i$ or
  $\sigma_i$ in $f(z)$ is $\Omega(1)$ far from any corresponding
  $\mu_i$ or $\sigma_i$ in $f(-z)$.  So $f(z)$ and $f(-z)$ give the
  desired mixtures.
\end{proof}

The mixtures given by Lemma~\ref{lem:FGk} let us extend the lower
bounds to mixtures of $k$ gaussians, but with a caveat: the mixtures
differ in \emph{at least one} of $\mu_i$ and $\sigma_i$, so the lower
bound now only applies to algorithms that recover~\emph{both} $\mu_i$
and $\sigma_i$ do the desired precision.

\begin{theorem}\label{thm:lowerk}
  Consider any algorithm that, given $n$ samples of any one
  dimensional gaussian mixture $F$ of $k = O(1)$ components, with probability
  $1-\delta$ learns both the $\mu_i$ to $\pm \eps\sigma$ and the
  $\sigma_i^2$ to $\pm \eps^2 \sigma^2$.  Then $n =
  \Omega_k(\eps^{2-6k}\log(1/\delta))$.
\end{theorem}
\begin{theorem}\label{thm:lowerkd}
  Consider any algorithm that, given $n$ samples of any
  $d$-dimensional gaussian mixture $F$ of $k = O(1)$ components with $\Var(F)
  = \sigma^2$, with probability $1-\delta$ for all $i \in [d]$ learns
  both $\mu_i$ to $\pm \eps\sigma$ and $\Sigma_{i,i}$ to $\pm \eps^2
  \sigma^2$.  Then $n = \Omega_k(\eps^{2-6k}\log(d/\delta))$.
\end{theorem}
\begin{proof}
  Identical to the proof of Theorems~\ref{thm:lower}
  and~\ref{thm:lowerd}, but applying Lemma~\ref{l:h2} to the $F$, $G$
  resulting from Lemma~\ref{lem:FGk} rather than the explicit ones in
  Equation~\eqref{eq:FG}.
\end{proof}
\fi

\iflong
\section{Algorithm for one-dimensional mixtures}

\subsection{Preliminaries and Notation}

\paragraph{Asymptotics.}  For any expressions $f$ and $g$, we use $f
\lesssim g$ to denote that there exists a constant $C > 0$ such that
$f \leq C g$.  Similarly, $f \gtrsim g$ denotes that $g \lesssim f$,
and $f \eqsim g$ denotes that $f \lesssim g \lesssim f$.

\paragraph{Parameters of the gaussian.}

The two gaussians have probabilities $p_i$, means $\mu_i$, and standard
deviations $\sigma_i$.  The overall mean and variance of the
mixture are $\mu = p_1 \mu_1 + p_2 \mu_2$ and 
\begin{equation}\label{eq:1d-var}
\sigma^2 = 
p_1((\mu_1-\mu)^2 + \sigma_1^2) + p_2((\mu_2-\mu)^2 + \sigma_2^2)
=
p_1p_2(\mu_1-\mu_2)^2 + p_1\sigma_1^2 + p_2\sigma_2^2\mper
\end{equation}
%

For almost all of the section, for simplicity of notation we will
assume that the overall mean $\mu = 0$.  We only need to consider $\mu
\neq 0$ when showing that we can estimate the moments precisely
enough.

We will also assume that $p_1, p_2 \in (0, 1)$ are both bounded away
from zero. We define
\begin{equation}
  \Delta_{\mu} = \abs{\mu_2 - \mu_1}
\qquad \Delta_{\sigma^2} = \abs{\sigma_2^2 - \sigma_1^2}\mper
\end{equation}

We also make use of a reparameterization of the gaussian distribution:
\begin{equation}\label{eq:reparam}
  \alpha = -\mu_1 \mu_2\qquad
  \beta = \mu_1 + \mu_2\qquad
  \gamma = \frac{\sigma_2^2 - \sigma_1^2}{\mu_2 - \mu_1}
\end{equation}
Note that these are independent of adding gaussian noise,
i.e. increasing both $\sigma_1^2$ and $\sigma_2^2$ by the same amount.
Also we have $\alpha \geq 0$, since the mean is zero.
With our assumption that $p_1, p_2$ are bounded away from zero we have
that
\[
\beta^2 \lesssim \alpha \eqsim \Delta_\mu^2.
\]
Finally, we will make use of the parameter
\[
\kappa = \max(1, \Delta_{\sigma^2}/\Delta_\mu^2)
\]
which will relate to how well-conditioned our equations are.  We have
that $\abs{\gamma} \lesssim \Delta_\mu \kappa$.

\paragraph{Excess moments.}

We define $M_i$ to be the $i$th moment of our distribution, $\E[x^i]$,
so $M_2 = \sigma^2$.

The \emph{excess kurtosis} of a distribution is a standard statistical
measure defined as $M_4/M_2^2 - 3$.  It is designed to be independent
of adding independent gaussian noise to the variable.  Inspired by
this, we define the \emph{excess moments} $X_i$ to be $M_i$ plus a
polynomial in $M_2, \dotsc, M_{i-1}$ such that the result is
independent of adding gaussian noise.  We have that:
\define{l:excess}{Lemma}{ For $\alpha = -\mu_1\mu_2, \beta = \mu_1 +
  \mu_2, \gamma = \frac{\sigma_2^2 - \sigma_1^2}{\mu_2 - \mu_1}$ we
  have that
  \begin{align}
    X_3 &:= M_3 &&= \alpha \beta + 3 \alpha \gamma\notag\\
    X_4 &:= M_4 - 3 M_2^2 &&= -2 \alpha^2 + \alpha \beta^2 + 6 \alpha \beta \gamma + 3\alpha \gamma^2
\ifdefined\excess\notag\else\label{eq:excess}\fi\\\
    X_5 &:= M_5 - 10 M_3 M_2 &&= \alpha(\beta^3 -8 \alpha\beta + 10 \beta^2 \gamma + 15\gamma^2\beta - 20 \alpha \gamma )\notag\\
  X_6 &:= M_6 - 15 M_4M_2 + 30 M_2^3 &&= \alpha (16 \alpha^2 - 12 \alpha
  \beta^2 - 60 \alpha \beta \gamma + \beta^4 + 15 \beta^3\gamma + 45
  \beta^2\gamma^2 + 15 \beta \gamma^3)\notag
  \end{align}
}%
\state{l:excess}%
\newcommand{\excess}{1}%
See Appendix~\ref{app:excess} for proof.  Since $\alpha, \beta,
\gamma$ are independent of adding gaussian noise, these definitions of
the $X_i$ are correct.

By inspection, we have for each $i \in \{3,4,5,6\}$ that
\begin{align}
  \abs{X_i} \lesssim \Delta_\mu^i \kappa^{i-2}.\label{eq:Xkappa}
\end{align}

For simplicity of notation, we also define $X_1 = \mu$ and $X_2 =
\sigma^2$, despite them not technically being ``excess,'' and refer to
$\{X_1, X_2, X_3, X_4, X_5, X_6\}$ as the excess moments.

\paragraph{Estimation of moments.}  All our algorithms in this section
proceed by first estimating the (excess) moments from the samples,
then estimating the mixture from these moments.  The relationship
between sample complexity and estimation error is as follows:
\define{l:estimatemoments}{Lemma}{%
  Suppose $p_1, p_2$ are bounded away from zero and our mixture has
  variance $\sigma^2$.  Given $O(\log(1/\delta)/\eps^2)$ samples, with
  probability $1-\delta$ we can compute estimates $\wh{X}_i$ of the
  first $O(1)$ excess moments $X_i$ satisfying $\abs{\wh{X}_i - X_i}
  \leq \eps \sigma^i$ .
}%
\state{l:estimatemoments}%
See Appendix~\ref{app:estimatemoments} for a proof.  We will state our
first two theorems in terms of the necessary error bound on
$\abs{\wh{X}_i - X_i}$ rather than sample complexity.  This is more
general, since it supports other forms of perturbation of the inputs.

For a statistic $f$ of the gaussian mixture, in general we use $f$ to
denote the true value of the statistic and $\wh{f}$ to denote the
estimate of $f$ from estimates of the moments.

\subsection{Algorithm overview}

\define[Precision better than $\Delta_\mu$]{thm:1dmean}{Theorem}{%
  Consider any mixture of gaussians where $p_1$ and $p_2$ are bounded
  away from zero, $c > 0$ a sufficiently small constant, and any $\eps
  < 1$.  If $\abs{\wh{X}_i - X_i} \leq c\eps \Delta_\mu^i$ for all $i
  \leq 6$, Algorithm~\ref{alg:1dgetmean} recovers the $p_i$ to $\pm
  \eps$, $\mu_i$ to $\pm \eps \Delta_\mu$, and $\sigma_i^2$ to $\pm
  \eps \Delta_\mu^2$.
}

\define[Precision between $\Delta_\mu$ and
$\sqrt{\Delta}_{\sigma^2}$]{thm:1dsamemean}{Theorem}{%
  Consider any mixture of gaussians where $p_1$ and $p_2$ are bounded
  away from zero, and any $\eps < 1$.  Suppose that $\eps
  \Delta_{\sigma^2} \gtrsim \Delta_\mu^2$.  If $\abs{\wh{X}_i - X_i}
  \leq \eps \Delta_{\sigma^2}^{i/2}$ for all $i \leq 6$,
  Algorithm~\ref{alg:1dsamemean} recovers the $p_i$ to $\pm O(\eps)$ and
  $\sigma_i^2$ to $\pm O(\eps) \Delta_{\sigma^2}$.
}

Our overall goal is to recover the $\mu_i$ to $\pm \eps \sigma$ and
the $\sigma_i^2$ to $\pm \eps^2 \sigma^2$ using roughly
$O(1/\eps^{12})$ samples.

We have two different algorithms for different parameter regimes.  The
first algorithm, Algorithm~\ref{alg:1dgetmean}, proceeds by first
learning the $\mu_i$, then using this to estimate the $\sigma_i$.
However, it only works well if we have the $X_i$ to within
$\Delta_\mu^i$; without this, we cannot get a nontrivial estimate of
the $\mu_i$, which causes the algorithm to also not get a decent
estimate of the $\sigma_i$.

\restate{thm:1dmean}

If $\Delta_{\sigma^2} \gg \Delta_\mu^2$, for example if $\Delta_\mu =
0$, one may hope to get a good estimate of the $\sigma_i^2$ despite
having more than $\Delta_\mu^i$ error in $X_i$.
Algorithm~\ref{alg:1dsamemean} does this by solving for the $\sigma_i$
under the assumption that $\mu_1 = \mu_2$.  We can show that the
solution is robust to $\Delta_\mu$ being small but nonzero, so the
algorithm does a good job when $\Delta_\mu^i \lesssim \abs{\wh{X}_i -
  X_i} \lesssim \Delta_{\sigma^2}^{i/2}$.  (When $\abs{\wh{X}_i -
  X_i}$ goes below this bound, the performance doesn't degrade but
does not improve as one would like.)

\restate{thm:1dsamemean}

In the remaining parameter regime, with $\abs{\wh{X}_i - X_i} >
\Delta_\mu^i + \Delta_{\sigma^2}^{i/2}$, the two gaussians are in
general indistinguishable and it suffices to just output the average
mean and variance.

To get a general result, we just need to figure out which of the three
parameter regimes we're in and apply the appropriate algorithm.
Algorithm~\ref{alg:1dcombine} does this by constructing sufficiently
good estimates of $\sigma, \Delta_{\mu},$ and $\Delta_{\sigma^2}$.  We
also invoke Lemma~\ref{l:estimatemoments} to get bounds on sample
complexity, showing:

\restate{thm:1d}

The regimes can be unified to get the following, simpler but weaker,
corollary:

\begin{corollary}
  Consider any mixture of gaussians where $p_1$ and $p_2$ are bounded
  away from zero, and any $\eps, \delta < 1$.  With $n = O(\log
  (1/\delta)/\eps^{12})$ samples, Algorithm~\ref{alg:1dcombine}
  returns $\{\mu_1, \mu_2\}$ to $\pm \eps \sigma$ additive error and
  the corresponding $\{\sigma_1^2, \sigma_2^2\}$ to $\pm \eps^2
  \sigma^2$ additive error with probability $1-\delta$.
\end{corollary}
\begin{proof}
  If $\eps\sigma = f \Delta_\mu$ for $f < 1$, then by the first
  clause of Theorem~\ref{thm:1d} with $\eps' = f^6$ we get the $\mu_i$
  to within $f^6 \Delta_\mu \leq f \Delta_\mu = \eps \sigma$ and the
  $\sigma_i^2$ to within $f^6 \Delta_\mu^2 \leq f^2 \Delta_\mu^2 =
  \eps^2 \sigma^2$.

  Otherwise, if $\eps \sigma > \Delta_\mu$ but $\eps^2\sigma^2 = f^2
  \Delta_{\sigma^2}$ for $f^2 < 1$, then by the second clause of
  Theorem~\ref{thm:1d} with $\eps' = f^6$ we get the $\sigma_i$ to
  within $f^6 \Delta_{\sigma^2} + \Delta_{\mu}^2 \lesssim \eps^2
  \sigma^2$.

  And by the last clause, if $\eps \sigma \gtrsim \Delta_\mu$ we get
  the $\mu_i$ to within $\Delta_\mu + \eps^6\sigma \lesssim \eps
  \sigma$, and if $\eps^2 \sigma^2 \gtrsim \max(\Delta_{\sigma^2},
  \Delta_\mu^2)$ we get the $\sigma_i^2$ to within $\Delta_\mu^2 +
  \Delta_{\sigma^2} + \eps^6 \sigma^2 \lesssim \eps^2 \sigma^2$.
\end{proof}

\subsection{Algorithm for better precision than $\Delta_\mu$}
\label{sec:precision-mu}

\begin{algorithm}
  \begin{algorithmic}
    \State\Comment{Recover gaussian mixture from (estimates of the) mean, variance, and excess moments}
    \State\Comment{Excess moments are function of moments, defined in~\eqref{eq:excess}}
    \Procedure{RecoverFromMoments}{$\eps, \mu, \sigma^2, X_3, X_4, X_5, X_6$}
    \State $\alpha \gets \Call{RecoverAlphaFromMoments}{X_3, X_4, X_5, X_6, \eps}$.
    \State $\gamma \gets \frac{1}{\alpha} \frac{\alpha^2 X_5 + 2X_3^3 + 2 \alpha^3 X_3 - 3
    X_3X_4\alpha}{4X_3^2 - 2 \alpha^3 - 3 X_4 \alpha}.$
    \State $\beta \gets \frac{1}{\alpha}(X_3 - 3\alpha \gamma)$.
    \State $\mu_1, \mu_2 \gets \frac{-\beta \mp \sqrt{\beta^2 + 4 \alpha}}{2}$.
    \State $p_1, p_2 \gets \frac{\mu_2}{\mu_2 - \mu_1}, \frac{-\mu_1}{\mu_2 - \mu_1}$.
    \State $\sigma_1^2 \gets \sigma^2 - \left(p_1\mu_1^2 + p_2\mu_2^2 - \mu_1\gamma\right)$.
    \State $\sigma_2^2 \gets \sigma_1^2 + (\mu_2 - \mu_1)\gamma$.
    \State \Return{$(p_1, \mu_1 + \mu, \sigma_1), (p_2, \mu_2 + \mu, \sigma_2)$}
    \EndProcedure
    \Procedure{RecoverAlphaFromMoments}{$X_3, X_4, X_5, X_6, \eps$}
    \State Let $y_{max}$ be the largest root of $2y^3 + X_4y - X_3^2 = 0$.
    \State $\kappa \gets 1 + \sqrt{\abs{X_4}}/y_{max}$ \Comment{$\kappa = \Theta(1 + \abs{\gamma}/\sqrt{\alpha})$}
    \State Define the $18$th degree polynomial $r(y) = p_5(y)^2 + p_6(y)^2$ for $p_5$ and $p_6$ given by~\eqref{eq:p5} and~\eqref{eq:p6}.
    \State \Comment{For the true moments, $\alpha$ is the only zero of $r$ on $(0, y_{max}]$.} \State Compute the set of roots $R$ of $r'(y)$.
    \State Let $\alpha$ be maximal element of $R \cup \{(1+\eps/\kappa)y_{max}\}$ satisfying
    \begin{align*}
      \alpha &\leq (1 + \eps/\kappa)y_{max}\\
      r(\alpha) &\leq \eps^2 \alpha^{18}\kappa^{10}.
    \end{align*}
    \State \Return $\alpha$.
    \EndProcedure
  \end{algorithmic}
  \caption{Algorithm for recovery of mixture of two gaussians in one
    dimension when the means are separated.}\label{alg:1dgetmean}
\end{algorithm}


In this section we derive Pearson's polynomial and extend it into a
robust algorithm.

\paragraph{Manipulation of $X_3, X_4, X_5$.}
Based on~\eqref{eq:excess} we can remove $\beta$ to get an equation
in $\{\alpha, \gamma, X_3, X_4\}$:
\begin{align}\label{eq:X4X3}
  6 (\alpha \gamma)^2 &=   X_3^2 - 2 \alpha^3 - X_4\alpha.
\end{align}
If we define $z := \alpha \gamma$, we can get another equation in
$\{\alpha, z, X_3, X_4, X_5\}$.
\begin{align*} 
  \alpha^2 X_5 - X_3^3 + 8 \alpha^3 X_3 &= \alpha^3 \beta^2 \gamma - 12 \alpha^3 \gamma^2 \beta + 4 \alpha^4\gamma - 27\alpha^3\gamma^3\\
  &= z (X_3 - 3z)^2 - 12 z^2(X_3 - 3z) + 4 \alpha^3z - 27 z^3\\
  &= 18z^3 - 18X_3z^2 + (4 \alpha^3 + X_3^2) z
\end{align*}
Substituting in~\eqref{eq:X4X3} we make the equation linear in $z$:
\begin{align*}
  \alpha^2 X_5 - X_3^3 + 8 \alpha^3 X_3 &= 3(z - X_3)(X_3^2 - 2
  \alpha^3 - X_4\alpha) + (4 \alpha^3 + X_3^2)z\\
  &= (4X_3^2 - 2 \alpha^3 - 3 X_4 \alpha)z + (3X_3X_4\alpha + 6 \alpha^3 X_3 - 3 X_3^3)
\end{align*}
or
\begin{align}\label{eq:zfrac5}
  z := \alpha \gamma &= \frac{\alpha^2 X_5 + 2X_3^3 + 2 \alpha^3 X_3 - 3
    X_3X_4\alpha}{4X_3^2 - 2 \alpha^3 - 3 X_4 \alpha}.
\end{align}
We can substitute $z$ back into~\eqref{eq:X4X3} and clear the
denominator to get a polynomial equation in the single variable
$\alpha$:\footnote{This polynomial~\eqref{eq:p5} is identical to (29)
  in Pearson's 1894 paper, up to rescaling variables by constant
  factors.  Our~\eqref{eq:zfrac5} is similarly identical to Pearson's
  (27).}
\begin{align}\label{eq:p5}
  p_5(\alpha) = 6 (2 X_3 \alpha^3 + X_5 \alpha^2 - 3 X_3X_4\alpha  + 2X_3^3)^2 + (2 \alpha^3 + 3 X_4 \alpha - 4X_3^2 )^2 (2 \alpha^3 + X_4\alpha - X_3^2) &= 0.
\end{align}
Therefore, given the excess moments $X_i$, we can find a set of
candidate $\alpha$ by solving for the positive roots of $p_5(y) = 0$.
Unfortunately, there are in general multiple such roots.  In fact, the
first five moments do not suffice to uniquely identify a gaussian
mixture, so we must incorporate the sixth moment.

\paragraph{Using the 6th moment.}
Analogously to the creation of~\eqref{eq:zfrac5}, we take the
expression for $X_6$ in~\eqref{eq:excess}, replace $\beta$ with
$X_4/\alpha - 3\gamma$, then remove $\gamma^2$ terms
using~\eqref{eq:X4X3} to get
\begin{align}\label{eq:zfrac6}
z := \alpha \gamma = \frac{4 X_3^4 - 4 X_3^2 X_4 \alpha - 8 X_3^2 \alpha^3 - X_4^2 \alpha^2 + 8 X_4 \alpha^4 + X_6 \alpha^3 + 4 \alpha^6}{10 X_3^3 - 7 X_3 X_4 \alpha - 2X_3\alpha^3}
\end{align}
(See Appendix~\ref{app:zfrac6} for a more detailed explanation.)
Combining with~\eqref{eq:zfrac5} and clearing the denominators gives
that
\begin{align}\label{eq:p6}
  p_6(\alpha) &= (4X_3^2 - 3X_4\alpha - 2\alpha^3)(4X_3^4 -
  4X_3^2X_4\alpha - 8X_3^2\alpha^3 - X_4^2\alpha^2 + 8X_4\alpha^4 +
  X_6\alpha^3 + 4\alpha^6) -\notag
  {}\\&\qquad
 (10X_3^3 - 7X_3X_4\alpha -
  2X_3\alpha^3)(2X_3^3 - 3X_3X_4\alpha + 2X_3\alpha^3 + X_5\alpha^2) = 0
\end{align}
which is another $9$th degree equation in $\alpha$ in terms of the
excess moments.

We would like to say that $\alpha$ is the only common positive root of
$p_5$ and $p_6$, but this is not always true.  Fortunately, we can
exclude the other common roots if we enforce an upper bound on
$\alpha$.

\paragraph{Restricting the domain.}
Let $y_{max}$ be the positive root of
\begin{align}
  2y^3 + X_4 y - X_3^2 = 0.\label{eq:ymax}
\end{align}
There is at most one such root by Descartes' rule of signs.  There
exists such a root because if $X_3 = \alpha(\beta + 3\gamma)$ is zero,
then $X_4 = \alpha(\beta + 3 \gamma)^2 - 6\alpha\gamma^2 - 2\alpha^2$
is negative.  And by~\eqref{eq:X4X3}, $\alpha \leq y_{max}$.

Moreover,
\begin{align}
  y_{max} \lesssim \alpha.\label{eq:ymaxbound}
\end{align}
(Since $p_1$ and $p_2$ are bounded away from zero, $\beta^2 \lesssim
\alpha$.  Then if $\gamma^2 \lesssim \alpha$, this follows from a
cubic polynomial with bounded coefficients having bounded roots.
Otherwise, $X_3^2 = \Theta(\alpha^2 \gamma^2)$ and $X_4 =
\Theta(\alpha \gamma^2)$ are positive so $y_{max} \leq X_3^2/X_4
\lesssim \alpha$.)

\paragraph{Combining the equations.}
We will show that $y = \alpha$ is the only solution to the set of
equations $0 < y \leq y_{max}, p_5(y) = 0, p_6(y) = 0$.  This
statement would suffice to recover the mixture given the exact excess
moments, but we also want the algorithm to have robustness to small
perturbations in the $X_i$.  We therefore define
\begin{align}
  r(y) := p_5(y)^2 + p_6(y)^2,\label{eq:r}
\end{align}
which we know is zero at $\alpha$.  We will show it is significantly
non-zero for any candidate $y$ that is far from $\alpha$ and still
within $[cy_{max}, y_{max}]$ for any constant $c > 0$.

The robustness will depend on the parameter
\begin{align}
  \label{eq:overlinegamma}
  \kappa := \max(1, \Delta_{\sigma^2}/\Delta_\mu^2) \eqsim 1 + \abs{\gamma}/\sqrt{\alpha}.
\end{align}
This is intuitive because the excess moments $X_i$ are bounded by
$O(\Delta_\mu^i \kappa^{i-2})$, which implies by inspection
of~\eqref{eq:p5} and~\eqref{eq:p6} that for all $\abs{y} \lesssim
y_{max}$, every monomial in $p_5$ and $p_6$ has magnitude bounded by
\begin{align}
  O(\Delta_\mu^{18}\kappa^{6}).\label{eq:monomialbound}
\end{align}

At this point, it is convenient to normalize so $\Delta_\mu \eqsim
\alpha \eqsim 1$.  While we state our lemmas in full generality, it is
better to think about this normalization and we will use it in the
proofs.

\define{lemma:quadratic}{Lemma}{%
  For any constant $c > 0$, and for all $\alpha \geq 0$ and $\beta,
  \gamma, y \in \R$ with $cy_{max} \leq y \leq y_{max}$ and $\beta^2
  \lesssim \alpha$ we have
  \[
  r(y) \gtrsim \kappa^{12}(y-\alpha)^2\alpha^{16}
  \]
} \state{lemma:quadratic} This is the key lemma of our proof, and
shown in Section~\ref{sec:quadratic}.
Note that the recovery algorithm will not know $y_{max}$ exactly, so
we need to extend the claim to slightly beyond $y_{max}$.

\define{lemma:quadratic+}{Lemma}{%
  For any constant $c > 0$, there exists a constant $c' > 0$ such that
  for all $\alpha \geq 0$ and $\beta, \gamma, y \in \R$ with $cy_{max}
  \leq y \leq y_{max} + c'(y_{max} - \alpha)$ and $\beta^2 \lesssim
  \alpha$ we have
  \[
  r(y) \gtrsim \kappa^{12}(y-\alpha)^2\alpha^{16}
  \]
}
\state{lemma:quadratic+}

The proof is in Section~\ref{sec:quadratic}.
This lemma lets us show that \textsc{RecoverAlphaFromMoments} returns
a good approximation to $\alpha$ if it is given good approximations to
the moments:

\define{l:alphafrommoments}{Lemma}{%
  Suppose $p_1, p_2$ are bounded away from zero, let $c > 0$ be a
  sufficiently small constant, and let $\eps < 1$.  Suppose further
  that $\abs{\wh{X}_i - X_i} \leq c\eps\Delta_\mu^i$ for all $i \in \{3,4,5,6\}$.  In this setting,
  the result $\wh{\alpha} = \Call{RecoverAlphaFromMoments}{\wh{X}_3,
    \wh{X}_4, \wh{X}_5, \wh{X}_6, \eps}$ satisfies
  \[
  \abs{\wh{\alpha} - \alpha} \lesssim \eps\Delta_\mu^2 / \kappa.
  \]
}
\state{l:alphafrommoments}

See Section~\ref{sec:getalpha} for a proof.
It is then easy to show that all the recovered parameters are good
approximations to the true parameters, getting the theorem:
\state{thm:1dmean}
\begin{proof}
  We normalize so $\Delta_\mu = 1$.  By
  Lemma~\ref{l:alphafrommoments},
  \[
  \abs{\wh{\alpha} - \alpha} \lesssim \eps/\kappa.
  \]
  Therefore $\alpha$ and the $\wh{X}_i$ for $i \in \{3,4,5,6\}$ all
  have error less than $\eps/\kappa$ times the corresponding upper
  bounds of $1$ and $\kappa^{i-2}$.  Then by Lemma~\ref{l:monomials},
  the error in any monomial in $\alpha$ and the $X_i$ is less than
  $\eps/\kappa$ times the upper bound on that monomial.

  Let us consider the error in $\wh{\gamma}$.  The equation is
  \[
  \wh{\gamma} := \frac{1}{\wh{\alpha}} \frac{\wh{\alpha}^2 \wh{X}_5 + 2\wh{X}_3^3 + 2
    \wh{\alpha}^3 \wh{X}_3 - 3 \wh{X}_3\wh{X}_4\wh{\alpha}}{4\wh{X}_3^2 - 2 \wh{\alpha}^3 - 3 \wh{X}_4 \wh{\alpha}}
  \]
  For the true $\alpha, X_3, X_4, X_5$, the numerator is
  $O(\kappa^3)$ and the denominator is
  $\Theta(\kappa^2)$, where to get the lower
  bound on the denominator we use from~\eqref{eq:X4X3} that
  \[
  4X_3^2 - 2 \alpha^3 - 3 X_4 \alpha =   X_3^2 + 4\alpha^3 + 3(X_3^2 - 2 \alpha^3 - X_4 \alpha) \geq 4\alpha^3 + 3(6\alpha \gamma)^2 \eqsim \kappa^2.
  \]
  Hence for the estimates, we have
  \[
  \wh{\gamma} = \frac{O(\kappa^3) \pm O(\eps\kappa^2)}{\Theta(\kappa^2) \pm O(\eps\kappa)} = \gamma \pm O(\eps)
  \]
  Then $\wh{\beta}$ and the $\wh{\mu}_i$ are trivial $\pm O(\eps)$
  approximations.  From this, the $\wh{p}_i$ are $\pm O(\eps)$
  approximations and the $\wh{\sigma}_i^2$ are $\pm
  O(\eps)$-approximations.  Rescaling $\eps$ gives the result.
\end{proof}

\subsection{Algorithm for precision between $\Delta_\mu$ and $\sqrt{\Delta}_{\sigma^2}$}

\begin{algorithm}
  \begin{algorithmic}
    \State\Comment{Recover gaussian mixture from (estimates of the) mean, variance, and excess moments}
    \State\Comment{Excess moments are function of moments, defined in~\eqref{eq:excess}}
    \Procedure{SameMeanRecoverFromMoments}{$\mu, \sigma^2, X_4, X_6$}
    \State $\Delta_{\sigma^2} \gets \sqrt{\frac{4}{3}X_4 + \frac{X_6^2}{25X_4^2}}$ \Comment{$\Delta_{\sigma^2} := \sigma_2^2 - \sigma_1^2$}
    \State $p_i \gets \frac{1}{2}(1 \mp \frac{X_6}{5X_4\Delta_{\sigma^2}})$\Comment{$p_1$ takes $-$ branch}
    \State $\sigma_1^2 \gets \sigma^2 - p_2 \Delta_{\sigma^2}$.
    \State $\sigma_2^2 \gets \sigma^2 + p_1 \Delta_{\sigma^2}$.
    \State \Return{$(p_1, \mu, \sigma_1), (p_2, \mu, \sigma_2)$}
    \EndProcedure
  \end{algorithmic}
  \caption{Algorithm for recovery of mixture of two gaussians in one
    dimension, when $\mu_1 \approx \mu_2$.}\label{alg:1dsamemean}
\end{algorithm}

Algorithm~\ref{alg:1dsamemean} solves for the gaussian mixture under
the assumption that $\mu_1 = \mu_2$.  First, we show that it is
correct and robust to perturbations in the moments; we will then show
that the moments are robust to perturbation of the means.

\begin{lemma}\label{l:1dsamemean}
  Suppose $\mu_1 = \mu_2$ and $p_1, p_2$ are bounded away from zero.
  Let $\Delta_{\sigma^2} := \sigma_1^2 - \sigma_2^2$.  For any $\eps$
  less than a sufficiently small constant, if $\abs{\wh{X}_i - X_i}
  \lesssim \eps \Delta_{\sigma^2}^{i/2}$ for all $i \in \{2,4,6\}$,
  then Algorithm~\ref{alg:1dsamemean} recovers $\sigma_i^2$ to $\pm
  O(\eps \Delta_{\sigma^2})$ additive error and $p_i$ to $\pm O(\eps)$
  additive error.
\end{lemma}
\begin{proof}
  First, we show that Algorithm~\ref{alg:1dsamemean} gives exact
  recovery when the moments are exact; then we show robustness.

  We choose to disambiguate the mixtures by $\sigma_2^2 \geq
  \sigma_1^2$ so $\gamma = \Delta_{\sigma^2} / (\mu_2 - \mu_1)$.  By
  examining Lemma~\ref{l:excess} as $\mu_i \to 0$ and $\gamma \to
  \Delta_{\sigma^2}/\mu_i$, we observe that when $\Delta_\mu = 0$ we
  have
  \begin{align}
    X_4 \to 3\alpha \gamma^2 \label{eq:xnomean}\\
    X_6 \to 15\alpha\beta\gamma^3\notag
  \end{align}
  which, in terms of $p_1, p_2, \Delta_{\sigma^2}$, for $\Delta_\mu =
  0$ implies that
  \begin{align*}
    X_4 &= 3p_1p_2\Delta_{\sigma^2}^2\\
    X_6 &= 15p_1p_2(p_2 - p_1)\Delta_{\sigma^2}^3
  \end{align*}
  Therefore
  \begin{align*}
    \frac{4}{3}X_4 + \frac{X_6^2}{25 X_4^2} &= 4p_1p_2\Delta_{\sigma^2}^2 + (p_2 -
    p_1)^2\Delta_{\sigma^2}^2\\
    &= (p_1 + p_2)^2 \Delta_{\sigma^2}^2 = \Delta_{\sigma^2}^2
  \end{align*}
  and
  \[
  p_2 - p_1 = \frac{X_6}{5X_4\Delta_{\sigma^2}}.
  \]
  The algorithm is thus correct given the exact moments.

  How robust is the algorithm?  We have that $X_4 \eqsim \Delta_{\sigma^2}^2$ and
  $\abs{X_6} \lesssim \Delta_{\sigma^2}^3$.  Hence $\wh{X_4} = (1 \pm O(\eps))
  X_4 \gtrsim \Delta_{\sigma^2}^2$ and $\abs{\wh{X}_6^2 - X_6^2} \lesssim \eps
  \Delta_{\sigma^2}^6$, and
  \[
  \abs{\frac{\wh{X}_6^2}{\wh{X}_4^2} - \frac{X_6^2}{X_4^2}} \leq \abs{\frac{\wh{X}_6^2}{\wh{X}_4^2} - \frac{\wh{X}_6^2}{X_4^2}} + \abs{\frac{\wh{X}_6^2}{X_4^2}
    - \frac{X_6^2}{X_4^2}}
  \lesssim \eps \Delta_{\sigma^2}^2.
  \]
  Therefore $\wh{\Delta}_{\sigma^2} = (1 \pm O(\eps))\Delta_{\sigma^2}$.  And since
  $\abs{\frac{\wh{X}_6}{\wh{X}_4} - \frac{X_6}{X_4}} \lesssim \eps
  \Delta_{\sigma^2}$, this means
  \[
  \abs{\wh{p}_i - p_i} \lesssim \abs{\frac{\wh{X}_6}{\wh{X}_4\wh{\Delta}_{\sigma^2}} - \frac{X_6}{X_4\Delta_{\sigma^2}}} \lesssim \eps
  \]
  as desired.
\end{proof}

\state{thm:1dsamemean}
\begin{proof}
  Let $G$ be the given gaussian mixture and $G'$ be the mixture with
  probabilities $p_i$ and variances $\sigma_i^2$ but both means moved
  to $\mu$, which we may assume without loss of generality is $0$.
  Then we can express $x \sim G$ as $y + z$ for $y \sim G'$ and
  $\abs{z} \leq \Delta_\mu$.  Define $X'_i$ to be the $i$th excess
  moment of $G'$.

  Since the sign of $y$ is independent of the mixture chosen, $\E[yz]
  = 0$.  Therefore $\abs{X'_2 - X_2} = \E[z^2] = p_1 \mu_1^2 + p_2
  \mu_2^2 \lesssim \Delta_\mu^2$.
  From~\eqref{eq:excess} and~\eqref{eq:xnomean}, we have that
  \begin{align*}
    \abs{X_4 - X'_4} &= \abs{-2\alpha^2 + \alpha \beta^2 + 6 \alpha \beta \gamma} \lesssim \Delta_\mu^2 \Delta_{\sigma^2}\\
    \abs{X_6 - X'_6} &= \abs{\alpha (16 \alpha^2 - 12 \alpha \beta^2 - 60 \alpha
    \beta \gamma + \beta^4 + 15 \beta^3\gamma + 45
    \beta^2\gamma^2)} \lesssim \Delta_\mu^2 \Delta_{\sigma^2}^2
  \end{align*}
  Therefore $\abs{X'_i - X_i} \lesssim \Delta_\mu^2
  \Delta_{\sigma^2}^{i/2 - 1}$ for all $i \in \{2, 4, 6\}$, so
  $\abs{X'_i - \wh{X}_i} \lesssim \eps \Delta_{\sigma^2}^{i/2}$.
  Lemma~\ref{l:1dsamemean} immediately implies the result.
\end{proof}

\subsection{Combining the algorithms to get general precision}

\begin{algorithm}
  \begin{algorithmic}
    \Procedure{Recover1DMixture}{$x_1, \dotsc, x_n, \delta$}
    \State Compute (excess) moments $\mu, \sigma^2, X_3, \dotsc, X_6$
    \State $f \gets (\frac{\log(1/\delta)}{n})^{1/12}$
    \Comment{Error $f^6 \sigma^i$ in each $X_i$.}
    \State $
    \overline{\Delta}_{\mu} \gets \left\{
      \begin{array}{cl}
        \min(\abs{X_3}^{1/3} + \abs{X_4}^{1/4},~~X_3/\sqrt{X_4}) & \text{if } X_4 > 0 \\

        \phantom{\min(}\abs{X_3}^{1/3} + \abs{X_4}^{1/4}\phantom{,~~X_3/\sqrt{X_4})} & \text{otherwise}
      \end{array}
    \right.
    $
    \Comment{$\Theta(\Delta_{\mu}) + O(f^{3/2} \sigma)$}
    \State $\overline{\Delta}_{\sigma^2} \gets \sqrt{\abs{X_4}}$ \Comment{$\Theta(\Delta_{\sigma^2}) \pm O(f^3 \sigma^2) \pm O(\Delta_{\mu}^2)$}
    \If{$f^2 \lesssim \overline{\Delta}_{\mu}^2/\sigma^2$}\Comment{Can get within $\pm \Delta_\mu^2$}
    \State \Return result of Algorithm~\ref{alg:1dgetmean} with $\eps \eqsim \sqrt{\frac{(\sigma/\overline{\Delta}_\mu)^{12}\log(1/\delta)}{n}}$.
    \ElsIf{$f^2 \lesssim \overline{\Delta}_{\sigma^2}/\sigma^2$}\Comment{Between $\pm \Delta_\mu^2$ and $\pm \Delta_{\sigma^2}$}
    \State \Return result of Algorithm~\ref{alg:1dsamemean}
    \Else\Comment{Can't distinguish either $\mu_i$ or $\sigma_i$, so output a single gaussian.}
    \State \Return $(1/2, \mu, \sigma^2), (1/2, \mu, \sigma^2)$
    \EndIf
    \EndProcedure
  \end{algorithmic}
  \caption{Combined algorithm for recovery of mixture of two gaussians
    in one dimension.}\label{alg:1dcombine}
\end{algorithm}

\xxx{Make sure no $p_i$ dependence in the constants in the algorithm.}

\state{thm:1d}

We compare the algorithm to the ``ideal'' algorithm which uses
$\Delta_\mu$ and $\Delta_{\sigma^2}$ instead of their estimates
$\overline{\Delta}_\mu$ and $\overline{\Delta}_{\sigma^2}$ to decide
which algorithm to use.  We show that:
\begin{itemize}
\item If the first branch is taken in either the ideal or the actual
  setting, then $\overline{\Delta}_\mu \eqsim \Delta_\mu$.
\item If the second branch is taken in either the ideal or the actual
  setting, then $\overline{\Delta}_{\sigma^2} \eqsim \Delta_{\sigma^2}$.
\end{itemize}
Therefore, up to constant factors in the sample complexity, the
Algorithm~\ref{alg:1dcombine} performs as well as the ideal algorithm,
which performs as well as the best of Algorithm~\ref{alg:1dgetmean},
Algorithm~\ref{alg:1dsamemean}, and outputting a single gaussian.  The
proof is given in Appendix~\ref{app:1dcombine}.

\section{Dimension Reduction}
\label{sec:dimension}

We first give a simple argument showing that the $d$-dimensional problem
reduces to the $4$-dimensional problem. We then give a separate result showing
that the $4$-dimensional problem reduces to the $1$-dimensional problem.
Since we previously saw a solution to the $1$-dimensional problem, our
reductions show how to solve the general $d$-dimensional problem.

To describe our reduction we need to know $\Var(F)$ up to a constant factor.
This can be accomplished with few samples as shown next.

\begin{lemma}\label{lem:estimate-var}
Given $n=O(\log(1/\delta))$ samples from a mixture~$F$ we can output a
parameter $\sigma^2$ such that $\Pr\Set{\sigma^2\in[\Var(F),2\Var(F)]}\ge 1-\delta.$
\end{lemma}
\begin{proof}
This follows from estimating the second moment of the distribution up to
constant multiplicative error and is shown in the proof of
Lemma~\ref{l:estimatemoments}.
\end{proof}

\begin{theorem}[($d$ to $4$)-reduction]
\label{thm:dto4}
Assume there is a polynomial time algorithm 
that $(\epsilon,\delta)$-learns mixtures of two gaussians in $\R^4$ from
$f(\epsilon,\delta)$ samples.
Then, for every $d\ge 4,$ there is a polynomial time algorithm 
that $(\epsilon,\delta)$-learns mixtures of two gaussians
using $f(\epsilon/20,\delta/10d^2)+O(\log(1/\delta))$ samples.
\end{theorem}

\begin{proof}
Let $\cA$ denote the assumed algorithm for~$\R^4.$ 
We give an algorithm $\cB$ for the $d$-dimensional problem. The algorithm is
given sample access to a mixture~$F$ of variance $\sigma^2=\Var(F).$ 
The algorithm $\cB$ always invokes $\cA$ with error parameter
$\epsilon/20$ and failure probability $\delta/10d^2.$ 

\paragraph{Algorithm $\cB$:}
\begin{enumerate}
\item Use Lemma~\ref{lem:estimate-var} to obtain a parameter $\widehat\sigma^2$
such that $\widehat\sigma^2\in[\sigma^2,2\sigma^2]$ with probability
$1-\delta/2.$ 

\textbf{\hspace{-2em}Determine $\widehat\mu^{(i)}$:}
\item For every $i\in[d]$ use $\cA$ to obtain numbers $\xi_i^p$ for
every $p\in\{1,2\}$ there is $q\in\{1,2\}$ such that 
$|\xi_i^p - \mu_{i}^{(q)}|\le\epsilon\widehat\sigma/20.$ For
each $i$ this can be done by invoking $\cA$ to solve the $1$-dimensional mixture problem
obtained by restricting the samples to coordinate~$i.$ 
\item\label{mu-easy} If for all $i$ we have $|\xi_{i}^1 -
\xi_{i}^2|\le\epsilon\widehat\sigma/4,$ then
put $\widehat\mu^{(1)}=\widehat\mu^{(2)}= (\xi_{i}^1)_{i\in[d]}$
\item\label{mu-anchor} Otherwise, let $i$ be the first index 
such that $|\xi_{i}^1-\xi_{i}^2|>\epsilon\widehat\sigma/4$ and do
for each $j\in[d]:$
\begin{enumerate}
\item Use $\cA$ to solve the $2$-dimensional mixture problem obtained by restricting to the
coordinates $i,j$ to accuracy $\epsilon\widehat\sigma/20$ in order to 
obtain numbers $(\nu_{i}^p,\nu_{j}^p)$ for $p=1,2$ as the estimate for the
two-dimensional means.  
\item Determine $p\in\{1,2\}$ such that $|\xi_{i}^1 - \nu_{i}^p|\le
\epsilon\widehat\sigma/10.$ 
If no such $p$ exists, terminate and output ``failure''.
\item Put $\widehat\mu^{(1)}_j = \nu_{j}^p$ and put $\widehat\mu^{(2)}_j 
= \nu_{j}^{3-p}.$
\end{enumerate}

\textbf{\hspace{-2em}Determine $\widehat\Sigma^{(i)}$:}
\item For every $i,j\in[d]$ use $\cA$ to obtain numbers $\xi_{ij}^p$ for
every $p\in\{1,2\}$ there is $q\in\{1,2\}$ such that 
$|\xi_{ij}^p - \Sigma_{ij}^{(q)}|\le\epsilon^2\widehat\sigma^2/20.$ For
each $i,j$ this can be done by using $\cA$ to solve the $2$-dimensional mixture problem
obtained by restricting the samples to coordinates $i,j.$
\item\label{Sigma-easy} If for all $i,j$ we have $|\xi_{ij}^1 -
\xi_{ij}^2|\le\epsilon^2\widehat\sigma^2/4,$ then
put $\widehat\Sigma^{(1)}=\widehat\Sigma^{(2)}= [\xi_{ij}^1]_{i,j\in[d]}$
\item\label{Sigma-anchor} Otherwise, let $i,j$ be the first indices in lexicographic order such
that $|\xi_{ij}^1-\xi_{ij}^2|>\epsilon^2\widehat\sigma^2/4$ and do
for each $k,l\in[d]:$
\begin{enumerate}
\item Invoke $\cA$ to solve the $4$-dimensional mixture problem obtained by restricting to the
coordinates $i,j,k,l$ to accuracy $\epsilon^2\widehat\sigma^2/20$ in order to 
obtain numbers $\sigma_{ij}^p,\sigma_{kl}^p$ for $p=1,2.$ 
\item Determine $p\in\{1,2\}$ such that $|\xi_{ij}^1 - \sigma_{ij}^p|\le
\epsilon^2\widehat\sigma^2/10.$
If no such $p$ exists, terminate and output ``failure''.
\item Put $\widehat\Sigma^{(1)}_{kl} = \sigma_{kl}^p$ and put
$\widehat\Sigma^{(2)}_{kl} = \sigma_{kl}^{3-p}.$
\end{enumerate}

\textbf{\hspace{-2em}Matching up $\wh{\Sigma}$ and $\wh{\mu}$.}
\item If there exist an $i, j, k$ with $\abs{\xi_{ij}^1 - \xi_{ij}^2}
  \geq \eps\sigma/2$ and $\abs{\xi_{k}^1 - \xi_{k}^2} \geq
  \eps\sigma/2$, then run $\cA$ on $\{i, j, k\}$ to get estimates
  $(\sigma_{ij}^p, \nu_k^p)$ of $(\Sigma_{ij}^p, \mu_k^p)$ for $p = 1,
  2$.  If there exists a permutation $\pi: \{1, 2\} \to \{1, 2\}$ with
  $\abs{\wh{\Sigma}_{ij}^{(1)} - \sigma_{ij}^{(\pi(1))}} <
  \abs{\wh{\Sigma}_{ij}^{(1)} - \sigma_{ij}^{(\pi(2))}}$ and
  $\abs{\wh{\mu}_{k}^{(1)} - \nu_{k}^{(\pi(1))}} > \abs{\wh{\mu}_{k}^{(1)}
    - \nu_{k}^{(\pi(2))}}$, then switch $\wh{\Sigma}^{(1)}$ and
  $\wh{\Sigma}^{(2)}$.
\end{enumerate}
\paragraph{Correctness of $\widehat\sigma^2$ and invocations of $\cA$.}
Appealing Lemma~\ref{lem:estimate-var} 
we have with probability $1-\delta/2,$
\[
\sigma^2\le \widehat\sigma^2\le 2\sigma^2\mper
\]
Moreover, we know that each invocation of~$\cA$ is on a mixture problem of
variance at most~$\sigma^2$ and we run the algorithm with accuracy parameter
$\epsilon/20$ and error probability $\delta/10d^2.$ The total number of
invocations is at most $5d^2$ and therefore every invocation is successful
with probability $1-\delta/2.$ In this case, we have that all the ``mean
parameters'' returned by $\cA$ are $\epsilon\sigma/20$-accurate and all the
``variance parameters'' are $\epsilon^2\sigma^2/400$-accurate. Both events
described above occur with probability $1-\delta$ and we will 
show that~$\cB$ succeeds in outputting a mixture that's $\epsilon$-close
to~$F$ assuming that these events occur. 
\paragraph{Correctness of means.}
On the one hand,
suppose that the case described in Step~\ref{mu-easy} occurs. In this
case, each pair of parameters is within distance 
$\epsilon\widehat\sigma/4\le\epsilon\sigma/2$
and the estimates are $\epsilon\sigma/20$ accurate. 
Hence, the output is $\epsilon\sigma$-close for both means.

On the other hand, consider the case described in Step~\ref{mu-anchor} and let $i$
denote the coordinate found by the algorithm. Since
$|\xi^1_i-\xi^2_i|>\epsilon\widehat\sigma/4\ge\epsilon\sigma/4$ 
it must be the case that
\[
\left|\mu^{(1)}_i-\mu^{(2)}_i\right| 
\ge \frac{\epsilon\sigma}{4}-\frac{\epsilon\sigma}{10}
= \frac{\epsilon\sigma}{8}\mper
\]
Further since all estimates are
$(\epsilon\sigma/20)$-accurate, there always must exist a
$p\in\{1,2\}$ such that $|\xi^1_i -
\nu^p_i|\le\epsilon\sigma/10\le\epsilon\widehat\sigma/10.$ There is at most
one such~$p$ since $|\xi^1_i-\xi^2_i|>\epsilon\widehat\sigma/4.$
For this $p$ we have $\nu^p_i$ and $\xi^1_i$ are either both
$(\epsilon\sigma/20)$-close to $\mu^{(1)}_i$ or they are both
$(\epsilon\sigma/20)$-close to $\mu^{(2)}_i$ but not both. It follows that for
every $j\in[d]$ our estimates $\nu^p_j$ all belong to the same $d$-dimensional
mean. This shows that we correctly identify $\mu^{(1)},\mu^{(2)}\in\R^d$ up to
additive error $\epsilon\sigma/20$ in each coordinate.

\paragraph{Correctness of covariances.}
The argument for $\widehat\Sigma^{(1)},\widehat\Sigma^{(2)}$ is
analogous.  Suppose that the case described in Step~\ref{Sigma-easy}
occurs. In this case, each pair of parameters is within distance
$\epsilon^2\sigma^2/2$ and the estimates are $\epsilon^2\sigma^2/20$
accurate. Hence, the output is $\epsilon^2\sigma^2$ accurate for both
covariance matrices.

Now consider the case described in Step~\ref{Sigma-anchor}
and let $(i,j)$ denote the pair of coordinates found by the algorithm. 
Since $|\xi^1_{ij}-\xi^2_{ij}|>\epsilon^2\widehat\sigma^2/4\ge\epsilon^2\sigma^2/4$ 
it must be the case
that 
\[
\left|\Sigma^{(1)}_{ij}-\Sigma^{(2)}_{ij}\right|
> \epsilon^2\sigma^2/4 -\epsilon^2\sigma^2/10 = \epsilon^2\sigma^2/8.
\] 
Further since all
estimates are $(\epsilon^2\sigma^2/20)$-accurate, there always must
exist a $p\in\{1,2\}$ such that $|\xi^1_{ij} -
\sigma^p_{ij}|\le\epsilon^2\sigma^2/10.$
For this $p$ we know that $\sigma^p_{ij}$ and $\xi^1_{ij}$ are either both
$(\epsilon^2\sigma^2/10)$-close to $\Sigma^{(1)}_{ij}$ or they are
both $(\epsilon^2\sigma^2/10)$-close to $\Sigma^{(2)}_{ij}.$ In
particular, for every $k,l\in[d]$ our estimates $\sigma^p_{kl}$ all
belong to the same $n$-dimensional covariance matrix. This shows that
we correctly identify $\Sigma^{(1)},\Sigma^{(2)}\in\R^{d\times d}$ up
to additive error $\epsilon^2\sigma^2/20$ in each coordinate.

\paragraph{Correctness of matching $\Sigma$ to $\mu$.}
If there does not exist an $i, j, k$ with $\abs{\xi_{ij}^1 -
  \xi_{ij}^2} \geq \eps\sigma/2$ and $\abs{\xi_{k}^1 - \xi_{k}^2} \geq
\eps\sigma/2$, then either the means or the variances are
indistinguishable and the order of matching doesn't matter.
Otherwise, since $\cA$ gives accuracy $(\eps\sigma/20,
\eps^2\sigma^2/20)$ and the true parameters are separated by at least
$(\eps \sigma/8, \eps^2\sigma^2/8)$, the correct pairing will only
have $\abs{\wh{\mu}_{k}^{(1)} - \nu_{k}^{(\pi(1))}} <
\abs{\wh{\mu}_{k}^{(1)} - \nu_{k}^{(\pi(2))}}$ or
$\abs{\wh{\Sigma}_{ij}^{(1)} - \sigma_{ij}^{(\pi(1))}} <
\abs{\wh{\Sigma}_{ij}^{(1)} - \sigma_{ij}^{(\pi(2))}}$ when $\pi$ is
the correct permutation for $\wh{\mu}$ and $\wh{\Sigma}$, respectively.

\end{proof}

\subsection{From $4$ to $1$ dimension}

For our reduction from $\R^4$ to $\R$ we invoke a powerful
anti-concentration result for polynomials in gaussian variables due to
Carbery and Wright.

\begin{theorem}[\cite{CarberyW01}]
\label{thm:CW}
Let $p(x_1,\dots,x_d)$ be a degree $r$ polynomial, normalized such
that $\Varr(p) = 1$ under the normal distribution.  Then, for any
$t\in\R$ and $\delta>0,$ we have
\[
\Pr_{x\sim N(0,1)^d}
\Set{ |p(x) - t | \le \tau } \le O(r) \cdot \tau^{1/r}\mper
\]
\end{theorem}

\begin{lemma}
\label{lem:anti-concentration}
Let $\mathrm{N}_9^d$ be the $d$-dimensional normal distribution
$N(0,1)^d$ conditioned on vectors of norm at most $9.$ There is a
constant $c>0$ such that for every $B\in\R^{d\times d},$
\[
\Pr_{a\sim \mathrm{N}_9^d}
\Set{ \left| a^\trans B a\right| \le c\cdot\|B\|_\infty } \le
\frac13 \mper
\]
\end{lemma}
\begin{proof}
  Observe that $p(x)= x^\trans Bx$ is a degree~$2$ polynomial in $n$
  gaussian variables. It is easy to see that the variance of $p$ under
  the normal distribution is at least the square of the largest entry
  of $B.$ That is, $\|B\|_\infty^2.$ Hence, we can apply
  Theorem~\ref{thm:CW} to $p(x)/V$ for some number $V\ge\|B\|_\infty$
  to conclude that
\[
\Pr_{x\sim N(0,1)^d}
\Set{ \left| x^\trans B x\right| \le c\cdot\|B\|_\infty } \le
\frac16 \mper
\]
On the other hand, $\|x\|^2 > 9$ with probability less than $1/6.$ Hence, the
claim follows.
\end{proof}

The next lemma is a direct consequence.

\begin{lemma}\label{lem:Sigma-reject}
  There is a constant $c>0$ such that for every
  $\epsilon>0,m\in\mathbb{N}$ and
  $\widehat\Sigma,\Sigma^{(1)},\Sigma^{(2)}\in\R^{d\times d}$ such
  that $\Norm{\widehat\Sigma-\Sigma^{(1)}}_\infty>\epsilon$ and
  $\Norm{\widehat\Sigma-\Sigma^{(2)}}_\infty>\epsilon$ we have
  \[
  \Pr_{a_1,\dots,a_m\sim \mathrm{N}_9^d}\Set{\exists i\in[m]\colon
    \left|a_i^\trans(\widehat\Sigma-\Sigma^{(1)})a_i\right|>c\epsilon
    \quad\text{and}\quad
    \left|a_i^\trans(\widehat\Sigma-\Sigma^{(2)})a_i\right|>c\epsilon
  }\ge 1-\left(\frac13\right)^{m}\mper
  \]
\end{lemma}
\begin{proof}
  For every fixed $i\in[m],$ by Lemma~\ref{lem:anti-concentration} and
  the union bound we have,
\[
\Pr_{a_i\sim \mathrm{N}_9^d}\Set{
  \left|a_i^\trans(\widehat\Sigma-\Sigma^{(1)})a_i\right|\le c\epsilon
  \quad\text{or}\quad
  \left|a_i^\trans(\widehat\Sigma-\Sigma^{(2)})a_i\right|\le c\epsilon
}\le 2/3\mper
\]
The claim therefore follows since the samples are independent.
\end{proof}

We have the analogous statement for vectors instead of matrices.

\begin{lemma}\label{lem:mu-reject}
  There is a constant $c>0$ such that for every
  $\epsilon>0,m\in\mathbb{N}$ and $\widehat\mu,\mu^{(1)},\mu^{(2)}\in\R^d$
  such that $\Norm{\widehat\mu-\mu^{(1)}}_\infty>\epsilon$ and
  $\Norm{\widehat\mu-\mu^{(2)}}_\infty>\epsilon$ we have
  \[
  \Pr_{a_1,\dots,a_m\sim \mathrm{N}_9^d}\Set{\exists i\in[m]\colon
    \left|\langle \widehat\mu-\mu^{(1)},a_i\rangle \right|>c\epsilon
    \quad\text{and}\quad \left|\langle
      \widehat\mu-\mu^{(2)},a_i\rangle \right|>c\epsilon }\ge
  1-\left(\frac13\right)^{m}\mper
  \]
\end{lemma}
\begin{proof}
  The proof is analogous to that of Lemma~\ref{lem:Sigma-reject}, but
  instead of Lemma~\ref{lem:anti-concentration} we directly appeal to
  the anti-concentration properties of the one-dimensional Normal
  distribution.
\end{proof}

We also note two obvious bounds.

\begin{lemma}\label{lem:Sigma-accept}
\label{lem:mu-accept}
Let $B\in\R^{4\times 4}$ and $\mu\in\R^{4}.$ Then,
\begin{enumerate}
\item $\Pr_{a\sim \mathrm{N}_9^4}\Set{|a^\trans Ba|\le
    O(\|B\|_\infty)}=1,$ and
\item $\Pr_{a\sim \mathrm{N}_9^4}\Set{|\langle a,\mu\rangle|\le
    O(\|\mu\|_\infty)}=1.$
\end{enumerate}
\end{lemma}
\begin{proof}
  This is immediate because the dimension is constant and the norm of
  $a$ is at most~$9$ with probability~$1.$
\end{proof}

We now have all the ingredients for our reduction from four to one dimension.

\begin{theorem}[($4$ to $1$)-reduction]
  \label{thm:4to1}
  Assume there is a polynomial time algorithm that
  $(\epsilon,\delta)$-learns a mixture of two gaussians in~$\R$
  from~$f(\epsilon,\delta)$ samples.  Then for some constant $c > 0$
  there is a polynomial time algorithm that $(\epsilon,\delta)$-learns
  mixtures of two gaussians in~$\R^4$ from
  $f(c\epsilon,c\delta/\log(\epsilon/\delta))+O(\log(1/\delta))$ samples.
\end{theorem}
\begin{proof}
  Let $\cA$ denote the assumed algorithm for one-dimensional mixtures.  We
give an algorithm $\cC$ for the $4$-dimensional problem.  We prove that
the algorithm $(O(\epsilon),O(\delta))$-learns mixtures of two gaussians
in~$\R^4$ given the stated sample bounds. We get the statement of the theorem by
rescaling~$\epsilon,\delta.$

We use Lemma~\ref{lem:estimate-var} to obtain a parameter $\widehat\sigma^2$
such that $\widehat\sigma^2\in[\sigma^2,2\sigma^2]$ with probability
$1-\delta.$ 

We will locate the unknown mixture parameters by doing a grid search
and checking each solution using the previous lemmas that we saw.  To
find a suitable grid for the means, we first find an estimate
$\widehat\mu\in\R^4$ so that $\mu^{(1)}$ and $\mu^{(2)}$ are both
within $2\widehat\sigma$ of $\widehat\mu$ in each coordinate.  This
can be done by invoking $\cA$ on each of the $4$ coordinates with
error parameter $1/2$ and success probability $1-\delta.$ For
$i=1,\dots,4$ we take $\widehat\mu_i$ to be either of the two
estimates for the means in the $i$-th invocation of~$\cA.$ Since
$\|\mu^{(1)}-\mu^{(2)}\|_\infty\le\sigma$ by assumption we know that
$\mu^{(1)},\mu^{(2)}$ are both within distance $2\sigma$ of
$\widehat\mu_i.$

Let $N_\mu$ be a $(c\epsilon\widehat\sigma)$-net in
$\ell_\infty$-distance around the point $\widehat\mu$ of width
$2\widehat\sigma$ in every coordinate.  For small enough $c>0,$ the
true parameters must be $(\epsilon\widehat\sigma/20)$ close to a point
contained in $N_\mu.$ This is because
$\sigma^2\ge\|\mu^{(1)}-\mu^{(2)}\|_\infty^2$ by definition.
Similarly, we let $N_\sigma =
([-\widehat\sigma^2,\widehat\sigma^2]\cap
(c\epsilon\widehat\sigma^2)\mathbb{Z})^{4\times 4}.$ Since
$\|\Sigma^{(i)}\|_\infty\le\sigma^2\le\widehat\sigma^2$ this net must
contain an $(\epsilon\sigma/20)$-close point to each true covariance
matrix.  Note that $|N_\mu|\times |N_\sigma|=\poly(1/\epsilon).$

\paragraph{Algorithm $\cC$:}
\begin{enumerate}
\item\label{step:estimates} Let
  $m=10\log((|N_\mu|\times |N_\sigma|)/\delta).$ Sample $a_1,\dots,a_m\sim
  \mathrm{N}_9^4$ and sample $x_1,\dots,x_{m'}\sim F$ where $m' =
  f(\epsilon',\delta')$ with $\epsilon' = c\epsilon$ and $\delta' =
  \delta/m$ for sufficiently small constant $c>0.$ For each $a_i$ run
  $\cA$ on $\{\langle a_i,x_j\rangle\colon j\in[m']\}$ with error
  parameter $\epsilon'$ and confidence $\delta'.$

  Denote the outputs of $\cA$ by $\Set{ (\widehat
    \mu^{1,i},\widehat\Sigma^{1,i}, \widehat \mu^{2,i},\widehat
    \Sigma^{2,i}) \colon i\in[m]}.$
\item For every vector $\widehat\mu\in N_\mu,$ do the following:
\begin{enumerate}
\item If there exists an $i\in[m]$ such that $|\langle
  a_i,\widehat\mu\rangle-\widehat\mu^{1,i}|>\epsilon\widehat\sigma/2$ and
  $|\langle
  a_i,\widehat\mu\rangle-\widehat\mu^{2,i}|>\epsilon\widehat\sigma/2,$ then
  label $\widehat\mu$ as ``rejected''. Otherwise if there is no such
  $i\in[m],$ label $\widehat\mu$ as ``accepted''.
\end{enumerate}
\item Let $M$ be the set of accepted vectors. If $M=\emptyset$ output
  ``failure'' and terminate.  Otherwise, choose $\wh{\mu}^{(1)},
  \wh{\mu}^{(2)} \in M$ to maximize $\norm{\wh{\mu}_1 -
    \wh{\mu}_2}_\infty$.
\item For every symmetric $\widehat\Sigma\in N_\sigma,$ do the
  following:
\begin{enumerate}
\item If there exists an $i\in[m]$ such that $|\langle
  a_i,\widehat\Sigma
  a_i\rangle-\widehat\Sigma^{1,i}|>\epsilon^2\widehat\sigma^2/2$ and $|\langle
  a_i,\widehat\Sigma
  a_i\rangle-\widehat\Sigma^{2,i}|>\epsilon^2\widehat\sigma^2/2,$ then label
  $\widehat\Sigma$ as ``rejected''.  Otherwise if there is no such
  $i\in[m],$ label $\widehat\Sigma$ as ``accepted''.
\end{enumerate}
\item Let $S$ be the set of accepted matrices.  If $S=\emptyset$
  output ``failure'' and terminate.  Otherwise, choose $\wh{\Sigma}^{(1)},
  \wh{\Sigma}^{(2)} \in S$ to maximize $\norm{\wh{\Sigma}^{(1)} -
    \wh{\Sigma}^{(2)}}_{\infty}$.

\item If there exists an $i \in [m]$ where there does not exist a
  permutation $\pi: \{1, 2\} \to \{1, 2\}$ such that for all $p \in
  \{1, 2\}$ we have
  $\abs{\inner{a_i, \wh{\Sigma}^{(p)} a_i} - \wh{\Sigma}^{\pi(p), i}} \leq
\eps^2 \widehat\sigma^2/2$
  and
  $\abs{\inner{a_i, \wh{\mu}^{(p)}} - \wh{\mu}^{\pi(p), i}} \leq \eps
\widehat\sigma/2,$
  then switch $\wh{\Sigma}^{(1)}$ and $\wh{\Sigma}^{(2)}$.
\end{enumerate}
\begin{claim}\label{cl:max-var}
  Let $\sigma_i^2$ denote the variance of the mixture problem induced
  by $a_i.$ Then we have that $\max_{i\in[m]} \sigma_i^2\le
  O(\sigma^2)$ with probability $1.$
\end{claim}
\begin{proof}
This follows directly from the concentration bounds in
Lemma~\ref{lem:Sigma-accept}.
\end{proof}
We need the following claim which shows that with high probability the
estimates obtained in Step~\ref{step:estimates} are
$(\epsilon/10)$-accurate.
\begin{claim}\label{cl:accurate}
  With probability $1-\delta,$ for all $i\in[m],$ there is a
  permutation $\pi_i$ so that:
\begin{enumerate}
\item $|\widehat \mu^{1,i}-\langle a_i,\mu^{(\pi_i(1))}\rangle|\le
  \epsilon\sigma/10$ and $|\widehat \mu^{2,i}-\langle
  a_i,\mu^{(\pi_i(2))}\rangle|\le \epsilon\sigma/10$
\item $|\widehat \Sigma^{1,i}-\langle a_i,\Sigma^{(\pi_i(1))}
  a_i\rangle|\le \epsilon^2\sigma^2/100$ and $|\widehat
  \Sigma^{2,i}-\langle a_i,\Sigma^{(\pi_i(2))}a_i\rangle|\le
  \epsilon^2\sigma^2/100$
\end{enumerate}
\end{claim}
\begin{proof}
  If $x\sim F$ then $\langle a,x\rangle$ is sampled from a
  $1$-dimensional mixture model with means $\langle
  a,\mu^{(1)}\rangle,\langle a,\mu^{(2)}\rangle$ and variances
  $\langle a,\Sigma^{(1)} a\rangle,\langle a,\Sigma^{(2)}a\rangle.$
  Note that we chose the error probability of $\cA$ small enough so
  that we can take a union bound over all $m$ invocations of the
  algorithm. Moreover, by Claim~\ref{cl:max-var}, all of these
  mixtures have variance at most $O(\sigma^2).$
\end{proof}

We suppose in what follows that the result of Claim~\ref{cl:accurate}
holds.

\paragraph{Correctness of means.}
Let
\[
A = \Set{\mu \colon \Norm{\mu^{(1)} - \mu}_\infty\le C\epsilon\sigma}
\cup
   \Set{\mu \colon \Norm{\mu^{(2)} - \mu}_\infty\le C\epsilon\sigma}.
\]
for a sufficiently large constant $C$.  We first claim that with
probability $1-\delta,$ every element that gets accepted is in~$A.$ To
establish the claim we need to show that with probability $1-\delta$
every element in $A^c\cap N_\mu$ gets rejected.  For any $\mu \in
A^c$, we have from Lemma~\ref{lem:mu-reject} that for some constant
$c' > 0$, with probability $1-1/3^m$ we have for some $i \in [m]$ that
\[
\min(\abs{\inner{\mu - \mu^{(1)}, a_i}}, \abs{\inner{\mu - \mu^{(2)},
    a_i}}) > c' C \eps \sigma \geq \eps \widehat\sigma
\]
if $C$ is sufficiently large, and hence $\mu$ is rejected.  By our
choice of $m$ and a union bound, with $1-\delta$ probability all $\mu
\in A^c \cap N_\mu$ are rejected.

We also need to show there exists $\bar\mu^{(1)},\bar\mu^{(2)}$ that
get accepted such that
$\Norm{\bar\mu^{(1)}-\mu^{(1)}}\le\epsilon\sigma$ and
$\Norm{\bar\mu^{(2)}-\mu^{(2)}}\le\epsilon\sigma.$ To see this take
$\bar\mu^{(1)}$ to be the nearest neighbor of $\mu^{(1)}$, which has
$\abs{\bar\mu^{(1)} - \mu^{(1)}} \leq c \eps\sigma$.  By
Lemma~\ref{lem:mu-accept}, it follows that
\[
\max_{i \in [m]} \abs{\inner{\bar\mu^{(1)} - \mu^{(1)}, a_i}} \leq O(c \eps \sigma)
\]
and hence $\bar \mu^{(1)}$ is accepted if $c$ is sufficiently small.
The symmetric argument holds for the nearest neighbor of $\mu^{(2)}$.

Now we can finish the argument by distinguishing two cases.  Consider
the case where $\Norm{\mu^{(1)} -\mu^{(2)}}_\infty\le 5C\epsilon\sigma.$ In
this case any accepted element must be $6C\eps \sigma$-close to both
means.  The other case is when $\Norm{\mu^{(1)} -\mu^{(2)}}_\infty >
5C\eps\sigma.$ In this case, $A$ contains two distinct clusters of
elements centered around each mean.  Each pair within a single cluster has
distance at most $2C\eps \sigma$, while any pair spanning the two
clusters has distance at least $3C\eps \sigma$.  Hence the pair of
largest distance are in different clusters and within $C\eps \sigma$
of the corresponding means.

\paragraph{Correctness of covariances.}
The argument is very similar to the previous one.  Let
\[
A = \Set{\Sigma \colon \Norm{\Sigma^{(1)} -
    \Sigma}_\infty\le C\epsilon^2\sigma^2} \cup \Set{\Sigma \colon
  \Norm{\Sigma^{(2)} - \Sigma}_\infty\le C\epsilon^2\sigma^2}.
\]
for a sufficiently large constant $C$.  We first claim that with
probability $1-\delta,$ every element that gets accepted is in~$A.$ To
establish the claim we need to show that with probability $1-\delta$
every element in $A^c\cap N_\sigma$ gets rejected.  For any $\Sigma \in
A^c$, we have from Lemma~\ref{lem:Sigma-reject} that for some constant
$c' > 0$, with probability $1-1/3^m$ we have for some $i \in [m]$ that
\[
\min(\abs{a_i^\trans(\Sigma - \Sigma^{(1)})a_i},
\abs{a_i^\trans(\Sigma - \Sigma^{(2)})a_i}) > c' C \eps^2 \sigma^2
\geq \eps^2 \widehat \sigma^2
\]
if $C$ is sufficiently large, and hence $\Sigma$ is rejected.  By our
choice of $m$ and a union bound, with $1-\delta$ probability all $\Sigma
\in A^c \cap N_\sigma$ are rejected.

We also need to show there exists $\bar\Sigma^{(1)},\bar\Sigma^{(2)}$ that
get accepted such that
$\Norm{\bar\Sigma^{(1)}-\Sigma^{(1)}}\le\epsilon^2\sigma^2$ and
$\Norm{\bar\Sigma^{(2)}-\Sigma^{(2)}}\le\epsilon^2\sigma^2.$ To see this take
$\bar\Sigma^{(1)}$ to be the nearest neighbor of $\Sigma^{(1)}$, which has
$\abs{\bar\Sigma^{(1)} - \Sigma^{(1)}} \leq c \eps^2\sigma^2$.  By
Lemma~\ref{lem:Sigma-accept}, it follows that
\[
\max_{i \in [m]} \abs{\inner{\bar\Sigma^{(1)} - \Sigma^{(1)}, a_i}} \leq O(c \eps^2 \sigma^2)
\]
and hence $\bar \Sigma^{(1)}$ is accepted if $c$ is sufficiently small.
The symmetric argument holds for the nearest neighbor of $\Sigma^{(2)}$.
\sloppy
Now we can finish the argument by distinguishing two cases.  Consider
the case where $\Norm{\Sigma^{(1)} -\Sigma^{(2)}}_\infty\le 5C\eps^2\sigma^2.$
In this case any accepted element must be $6C\eps^2 \sigma^2$-close to
both $\Sigma^{(i)}$.  The other case is when $\Norm{\Sigma^{(1)}
  -\Sigma^{(2)}}_\infty > 5C\eps^2\sigma^2.$ In this case, $A$
contains two distinct clusters of elements centered around each
$\Sigma^{(i)}$.  Each pair within a single cluster has distance at most
$2C\eps^2 \sigma^2$, while any pair spanning the two clusters has
distance at least $3C\eps^2 \sigma^2$.  Hence the pair of largest
distance are in different clusters and within $C\eps^2 \sigma^2$ of
the corresponding $\Sigma^{(i)}$.

\paragraph{Correctness of matching $\Sigma$ to $\mu$.}
If either $\Norm{\Sigma^{(1)} -\Sigma^{(2)}}_\infty\le
5C\eps^2\sigma^2$ or $\Norm{\mu^{(1)} -\mu^{(2)}}_\infty\le
5C\epsilon\sigma$, then matching the $\wh{\Sigma}^{(p)}$ to the
$\wh{\mu}^{(p)}$ is unnecessary.  Otherwise, we have for each $i \in
[m]$ that the probability that either $\wh{\Sigma}^{(1)}$ or
$\mu^{(1)}$ matches the wrong mean under $a_i$ is at most $2/3$.
Hence with $1-\delta$ probability, one of the $a_i$ will disambiguate
the two.  (One must be careful because $\wh{\mu}$ and $\wh{\Sigma}$
depend on the randomness in $a_i$, but $m$ is large enough that we can
union bound over all $\abs{N_\mu} \times \abs{N_\sigma}$ possibilities.)
\end{proof}

Combining our two reductions we immediately have the following result.

\state{thm:d}

\begin{proof}
  By Corollary~\ref{cor:e12}, there is an algorithm that
  $(\epsilon,\delta)$-learns mixtures of two gaussians in~$R$ from
  $O(\eps^{-12} \log(1/\delta))$ samples.
  Hence, by Theorem~\ref{thm:4to1}, there is an algorithm that
  $(\epsilon,\delta)$-learns mixtures of two gaussians in~$\R^4$ using sample
size
\[
  O(\eps^{-12}\log(\log(1/\epsilon\delta)/\delta) +\log(1/\delta))
=  O(\eps^{-12}\log(\log(1/\epsilon\delta)/\delta))\mper
\]

Finally, by Theorem~\ref{thm:dto4}, there is an algorithm that
$(\epsilon,\delta)$-learns mixtures of two gaussians in $\R^d$ from
using a number of samples that is bounded by
\[
  O(\eps^{-12}\log(2d\log(d/\epsilon\delta)/\delta)+\log(1/\delta))
  = O(\eps^{-12}\log(2d\log(d/\epsilon\delta)/\delta))\mper
\]
\end{proof}

\section{Algorithm in the TV norm}
\label{sec:TV}

\begin{algorithm}
  \begin{algorithmic}
    \Procedure{RecoverTVApprox}{$\eps, \delta$}
    \State Take $O(d^2 \log(1/\delta))$ samples and compute the empirical
    overall covariance $\wh{\Sigma}$.
    \State Factor $\wh{\Sigma}$ as $AA^T$.
    \State Take $O((\eps')^{-12}\log(1/\delta))$ samples $x_1, \dotsc, x_n$ for $\eps' = \Theta(\frac{\eps^3}{d^{2.5}\sqrt{\log(d/\eps)}})$

    \State $(\wh{\mu}_1, \wh{\Sigma}_1), (\wh{\mu}_2, \wh{\Sigma}_2) \gets \text{RecoverParamApprox}(A^{-1}x_1, \dotsc, A^{-1}x_n)$
    \State $\wh{\lambda}^2, v \gets $ minimum eigenvalue/eigenvector in either $\wh{\Sigma}_1$ or $\wh{\Sigma}_2$ (WLOG of $\wh{\Sigma}_1$)
    \If{$\wh{\lambda}^2 > 2(\eps')^2d/\eps^2$}
    \State \Return $(\wh{\mu}_1, A\wh{\Sigma}_1A^T), (\wh{\mu}_2, A\wh{\Sigma}_2A^T)$.
    \Else
    \For{$t \in [T] = [O(\log(1/\delta))]$}
    \State Take $O(d^2/\eps^2)$ samples $x_1, \dotsc, x_m$.
    \State Partition into $S = \{i : \abs{v^T(A^{-1}x_i - \wh{\mu}_1)} \lesssim \eps'\sqrt{d \log \frac{d}{\eps}}/\eps\}$, $\overline{S} = [n]\setminus S$.
    \State $G_1^{(t)}, G_2^{(t)} \gets$ empirical means/covariances of $x_S$ and $x_{\overline{S}}$.
    \EndFor
    \State For $i \in \{1, 2\}$,
    \[
    \wh{G}_i \gets \argmin_{G_i^{(t)}: t \in [T]} \median_{j \in [T]} \TV(G_i^{(t)}, G_i^{(j)})
    \]
    \State \Return $\wh{G}_1, \wh{G}_2$
    \EndIf
    \EndProcedure
  \end{algorithmic}
  \caption{Algorithm for TV approximation}
  \label{alg:tvapprox}
\end{algorithm}

In order to get a good approximation in total variation distance, we
use the following lemma (proven in the appendix):

\define{l:GTVparam}{Lemma}{
  Let $G, G'$ be $d$-dimensional gaussians with means and covariance
  matrices $(\mu, \Sigma)$ and $(\mu', \Sigma')$, respectively.
  Suppose that the minimum eigenvalue of $\Sigma$ is $\lambda^2$.
  Then
  \[
  \TV(G, G') \lesssim \frac{1}{\lambda}(\norm{\mu-\mu'} + \norm{\Sigma-\Sigma'}_F)
  \]
}
\restate{l:GTVparam}

The intuition for the algorithm is that either the minimum eigenvalue
is large (in which case parameter estimation works well) or it is
small (in which case we can cluster based on that direction).

%
%
\state{thm:tv}
\begin{proof}
  Because the overall mixture is subgaussian, our estimate
  $\wh{\Sigma}$ of the overall covariance $\Sigma$ will be good: for
  all $v$ we have $v^T\wh{\Sigma}v = (1 \pm 0.25) v^T \Sigma v$.
  Hence the normalized samples $y_1, \dotsc, y_n = A^{-1}x_1, \dotsc,
  A^{-1}x_n$ have covariance $\Sigma'$ with all eigenvalues between
  $3/4$ and $5/4$.  Because our algorithm is linear in normalization
  factor $A$, we may assume that $A$ is the identity and $\Sigma$ has
  all eigenvalues between $3/4$ and $5/4$.

  By Corollary~\ref{cor:e12}, therefore, the parameter estimation
  algorithm will learn (up to permutation) the $\mu_i$ to $\pm \eps'$
  in each coordinate and $\Sigma_i$ to $\pm (\eps')^2$ in each
  coordinate.

  Now, let $\lambda^2$ be the minimum eigenvalue of either $\Sigma_1$ or
  $\Sigma_2$.  We have by Lemma~\ref{l:GTVparam} that there is a
  permutation with
  \begin{align*}
    \TV(\wh{G_1}, G_{\pi_1}) + \TV(\wh{G_2}, G_{\pi_2}) &\lesssim
    \frac{1}{\lambda}(\eps'\sqrt{d} + (\eps')^2d) \\
    &\lesssim \eps'\sqrt{d}/\lambda
  \end{align*}
  where the second step follows from $\eps'\sqrt{d} < 1$ in order for
  the result to be nontrivial (since $\TV \leq 1$ always).  Hence the
  results from parameter approximation are sufficient if $\lambda
  \gtrsim \eps'\sqrt{d}/\eps$.  Since our approximations
  $\wh{\Sigma_i}$ of $\Sigma_i$ are within $d(\eps')^2$ in Frobenius
  norm, $\abs{\wh{\lambda}^2 - \lambda^2} \leq d(\eps')^2$.  Hence the
  first branch of the conditional will be taken if and only if
  $\lambda \gtrsim \eps'\sqrt{d}/\eps$, in which case it gives an
  $\eps$ approximation in the $\TV$ norm.

  If the other branch is taken, it finds a $v$ for which
  $v^T(\Sigma_1) v - \mu_1\mu_1^T \leq 3 (\eps'\sqrt{d}/\eps)^2$
  (where we choose $\Sigma_1$ without loss of generality).  Hence
  \[
  \E_{x \sim G_1} (v^T(x-\mu_1))^2 \leq  3 (\eps'\sqrt{d}/\eps)^2.
  \]
  Since $v^TG_1$ is Gaussian, this means for $x \sim G_1$ that
  $\abs{v^T(x-\mu_1)} \lesssim \eps'\sqrt{d\log
    (\frac{d}{\eps})}/\eps$ with probability $1 -
  O(\frac{\eps^2}{d^2})$.  By a union bound, in any inner loop, with
  $9/10$ probability all the samples in $x_1, \dotsc, x_m$ that are
  drawn from $G_1$ will then lie in $S$.

  On the other hand, $\E_{x \sim F} (v^T(x - \mu_1))^2 \geq 3/4$ by
  our assumption on the overall covariance, so
  \[
  \E_{x \sim G_2} (v^T(x-\mu_1))^2 \geq 3/4.
  \]
  Therefore for $x \sim G_2$, $\Pr[\abs{v^T(x-\mu_1)} \lesssim
  \eps'\sqrt{d\log (\frac{d}{\eps})}/\eps] \lesssim \eps'\sqrt{d\log
    (\frac{d}{\eps})}/\eps$.  For our choice of $\eps'$, this is less
  than $\eps^2/d^2$, so by a union bound, in any inner loop, with
  $9/10$ probability none of the samples drawn from $G_2$ will lie in
  $S$.

  Overall, this means that each inner loop will correctly classify $S$
  as the samples from $G_1$ and $\overline{S}$ as the samples from
  $G_2$ with $8/10$ probability.  Furthermore, given the correct
  classifications, Lemma~\ref{l:empiricalgaussian} shows that
  $G_1^{(t)}$ and $G_2^{(t)}$ will be $\eps/3$ approximations to $G_1$
  and $G_2$ in the $\TV$ norm with at least $19/20$ probability.  By
  Lemma~\ref{l:median}, this means that the result $\wh{G}_1$ and
  $\wh{G}_2$ of the procedure will be a $\eps$ approximation to $G_1$
  and $G_2$ with $1 - \delta/2$ probability.

  Thus regardless of which branch is taken, the result of
  \texttt{RecoverTVApprox} will be an $\eps$ approximation in the
  $\TV$ norm with $1-\delta$ probability.  The measurement complexity
  is dominated by the parameter estimation call.
\end{proof}

This sample complexity can be improved if the covariances of the two
Gaussians have similar eigenvalues and eigenvectors (e.g., they are
isotropic):

\state{thm:tv-isotropic}
\begin{proof}
  The algorithm is just Algorithm~\ref{alg:tvapprox}, changed to have
  $\eps' = \Theta(\frac{\eps}{\sqrt{d \log(d/\eps)}})$.  The only
  difference from the proof of Theorem~\ref{thm:tv} is that now, when
  we consider
  \[
  \E_{x \sim G_2} (v^T(x-\mu_1))^2 \geq 3/4
  \]
  we also have
  \[
  \E_{x \sim G_2} (v^T(x-\mu_2))^2 \leq C \E_{x \sim G_1}
  (v^T(x-\mu_1))^2 \leq 3(\eps'\sqrt{d}/\eps)^2.
  \]
  Thus $v^T\mu_1$ is $\Theta(\frac{\eps}{\eps'\sqrt{d}})$
  standard deviations away from the mean of $v^TG_2$.  So
  \[
  \Pr_{x \sim G_2}[\abs{v^T(x - \mu_1)} \lesssim \eps'\sqrt{d\log(d/\eps)}] \lesssim \text{exp}(-\Omega(\frac{\eps}{\eps'\sqrt{d}})^2)
  \]
  which is $O(\eps^2/d^2)$ for the given $\eps'$.
\end{proof}

By Theorem~\ref{thm:lowertv}, $\Omega(d^6/\eps^{12})$ samples are
\emph{necessary} for a $\TV$ approximation even in the approximately
isotropic case.  Hence Theorem~\ref{thm:tv-isotropic} is tight up to
logarithmic factors.
\fi

\bibliography{gaussian}
\bibliographystyle{alpha}
\iflong
\appendix

\section{Utility Lemmas}

\subsection{Approximating the coefficients of a polynomial will approximate its value}

The following lemma shows that a constant size monomial is robust to
perturbations of its inputs.

\begin{lemma}\label{l:monomials}
  Let $a, b, c \in \R^k$ for constant $k$ and $0 \leq \eps \leq 1$.  If
  $\abs{a_i} \leq b_i$ and $\abs{c_i} \leq \eps b_i$, we have that
  \[
  \left|\prod_{i=1}^k (a_i + c_i) - \prod_{i=1}^k a_i\right| \lesssim \eps \prod_{i=1}^k b_i.
  \]
\end{lemma}
\begin{proof}
  All $2^k-1 \lesssim 1$ terms on the left are bounded by the value on
  the right.
\end{proof}

\begin{lemma}\label{l:polynomials}
  Let $p$ be any constant-degree polynomial in a constant number of
  variables $a_1, \dotsc, a_t$ with constant coefficients.  Let $\overline{q}$
  equal $p$ except with all the coefficients having their absolute
  value taken. Suppose $\abs{a_i} \leq b_i$ and $\abs{c_i} \leq \eps
  b_i$ for some $b, c \in \R^t$ and $0 \leq \eps \leq 1$.  Then
  \[
  \abs{p(a_1, \dotsc, a_t) - p(a_1 + c_1, \dotsc, a_t + c_t)} \lesssim
  \eps \overline{p}(b_1, \dotsc, b_t).
  \]
\end{lemma}
\begin{proof}
  Apply Lemma~\ref{l:monomials} to each monomial.
\end{proof}

\subsection{Estimating moments of a distribution from samples}

The following lemma shows that we can estimate moments well.

\label{app:estimatemoments}
\restate{l:estimatemoments}
\begin{proof}
  We partition the samples $x_i$ into $O(\log(1/\delta))$ groups of
  size $k = O(1/\eps^2)$, then compute the median (over groups) of the
  empirical excess moment of the group.  We will show that this gives
  the desired result.

  Suppose we want to compute $t = O(1)$ moments.  Because our samples
  $x_i$ are the sum of a gaussian and a bounded variable and hence
  subgaussian, $\E[x_i^p] \lesssim \sigma^p$ for any $p \leq t$.
  Therefore $\Varr(x_i^p) \leq \E[x_i^{2p}] \lesssim \sigma^{2p}$.

  For a group of $k$ samples $x_i$, consider how well the empirical
  $p$th moment $\wh{M}_p = \frac{1}{k} \sum x_i^p$ approximates the
  true moment $M_p$.  We have that $\Varr(\wh{M}_p) \lesssim
  \sigma^{2p}/k$.  By Chebyshev's inequality, then, for any $c > 0$ we
  have
  \[
  \Pr[ \abs{\wh{M}_p - M_p} > O( \sigma^p / \sqrt{ck})] \leq c.
  \]
  Setting the constant $c = 1/(4t)$ and then choosing $k =
  O(1/(c\eps^2))$, we have with $3/4$ probability that
  \[
  \abs{\wh{M}_p - M_p} \leq \eps \sigma^p
  \]
  for all $p \leq t$.  Then since the $X_p$ are polynomials in the
  $M_p$ with total degree $p$, by Lemma~\ref{l:polynomials} we have for
  all $p \leq t$ that
  \begin{align}\label{eq:xp1}
    \abs{\wh{X}_p - X_p} \lesssim \eps \sigma^p.
  \end{align}
  Call a block where this happens ``good.''  Since each block is good
  with $3/4$ probability and there are $O(\log (1/\delta))$ blocks,
  with $1 - \delta$ probability more than half the blocks are
  ``good.''  If this is the case, then for each $p$ the median
  $\wh{X}_p$ will also satisfy~\eqref{eq:xp1}.  Rescaling $\eps$ gives
  the result.
\end{proof}

\subsection{Relationship between parameter distance and TV distance}

The next few lemmas are elementary results relating parameter distance
and total variation distance of Gaussians.

\begin{lemma}\label{l:gaussdist}
  Let $G, G'$ be $d$-dimensional gaussians with means and covariance
  matrices $(\mu, \Sigma)$ and $(\mu', \Sigma')$, respectively.
  Suppose that the maximum eigenvalue of $\Sigma$ is $\lambda^2$.
  Then
  \[
  \TV(G, G') \gtrsim \min(1, \norm{\mu-\mu'}/\lambda).
  \]
\end{lemma}
\begin{proof}
  Project onto $(\mu - \mu')$ and solve the $d=1$ case.
\end{proof}

\state{l:GTVparam}
\begin{proof}
  First, note that by symmetry $\TV(N(\mu, \Sigma), N(\mu', \Sigma'))
  \geq \TV(N(\mu, \Sigma), N(\mu, \Sigma')) = \TV(N(0, \Sigma), N(0,
  \Sigma'))$.  Combined with the triangle inequality we get
  \[
  \TV(N(\mu, \Sigma), N(\mu', \Sigma')) \leq \TV(N(0, \Sigma), N(0, \Sigma')) + \TV(N(\mu, \Sigma), N(\mu', \Sigma))
  \]
  so it suffices to consider the variation in $\mu$ and $\Sigma$
  separately.  From Pollard~\cite{Pollard} we have that $\TV(N(\mu,
  I_d), N(\mu', I_d)) \lesssim \norm{\mu-\mu'}$, so
  \[
  \TV(N(\mu, \Sigma), N(\mu', \Sigma)) \lesssim \norm{\mu-\mu'} / \lambda.
  \]
  For the covariance term, in one dimension it is an easy calculation
  that $\TV(N(0, \lambda^2), N(0, \lambda^2 + \eps^2)) \lesssim
  \eps/\lambda$.  In general, by rotation invariance and scaling we
  may assume that $\Sigma = I$, and by decomposing $\Sigma'$ we may
  assume that $\Sigma' - \Sigma$ is very sparse: either a single
  diagonal term or a single pair of symmetric off-diagonal terms.  If
  $\Sigma' - \Sigma$ is a single diagonal term it reduces directly to
  the one dimensional case, and the remaining option reduces to the
  two dimensional case
  \[
  \Sigma = \left(
    \begin{array}{cc}
      1&0\\0&1
    \end{array}\right) \hspace{1in}
    \Sigma' = \left(
    \begin{array}{cc}
      1&\eps\\\eps&1
    \end{array}
\right)
  \]
  which is just two one-dimensional cases on the eigenvectors $(1, 1)$
  and $(1, -1)$, giving the result.
\end{proof}

\begin{lemma}\label{l:empiricalgaussian}
  Given $n$ samples from a single Gaussian $G = (\mu, \Sigma)$ in $d$
  dimensions, the empirical $\wh{G} = (\wh{\mu}, \wh{\Sigma})$
  satisfies $\TV(G, \wh{G}) \lesssim d\sqrt{\frac{\log
  (1/\delta)}{n}}$ with probability $1-\delta$.
\end{lemma}
\begin{proof}
  The $\TV$ distance is independent of linear transformations, so
  assume $\mu = 0$ and $\Sigma = I_d$ without loss of generality.
  Then $\wh{\mu} \sim N(0, I_d/n)$, so $\norm{\wh{\mu}} \leq
  \sqrt{\frac{d + \log (1/\delta)}{n}}$ with probability $1-\delta$.
  For the covariance, for $i \neq j$ we have that $\wh{\Sigma}_{i,j}
  \sim N(0, 1)^2/n$ which is subexponential with constant parameters, so
  and $\norm{\Sigma - \wh{\Sigma}}_F \leq d\sqrt{\frac{\log
      (1/\delta)}{n}}$ with probability $1-\delta$.
  Apply Lemma~\ref{l:GTVparam} to get the result.
\end{proof}

\subsection{Getting a TV approximation with arbitrarily high probability}

The next lemma shows that we can amplify our probability of getting a
$\TV$ approximation using $O(\log (1/\delta))$ samples.

\begin{lemma}\label{l:median}
  Let $X$ be an arbitrary space with norm $\norm{\cdot}$.  Let $x_1,
  \dotsc, x_m$ be drawn independently from $X$ so that $\Pr[\norm{x_i
    - x^*} \leq \eps] \geq 3/4$ for some $x^*$.  Then
  \[
  \wh{x} := \argmin_{x_i : i \in [m]} \median_{j \in [m]} \norm{x_i - x_j}
  \]
  satisfies $\norm{x^*-\wh{x}} \leq 3\eps$ with $1 - 2^{-\Omega(m)}$
  probability.
\end{lemma}
\begin{proof}
  With the given probability, more than $m/2$ of the $x_i$ will have
  $\norm{x_j - x^*} \leq \eps$.  Each of these will have $\median_{j
    \in [m]} \norm{x_i - x_j} \leq 2 \eps$.  Therefore the resulting
  $\wh{x}$ is within $2\eps$ of at least one of these $m/2$ elements,
  or within $3\eps$ of $x^*$.
\end{proof}

\subsection{Gaussian anticoncentration}

This section gives some lemmas showing for a polynomial $p$ that
$p(x)$ is not too concentrated around zero if $x$ is drawn from a
univariate or multivariate Gaussian.

\begin{lemma}\label{lem:unip}
  Let $p(x)$ be a univariate polynomial of degree $d$ and with leading
  coefficient $a x^d$.  Then for any $\eps > 0$,
  \[
  \Pr_{x \sim N(0, 1)} [ \abs{p(x)} < \abs{a} \eps] < d\eps^{1/d}.
  \]
\end{lemma}
\begin{proof}
  By scaling we can assume $a = 1$.  Then we can factor $p(x)$ over
  $\C$, getting $p(x) = \prod_{i=1}^d (x - z_i)$ for some complex
  numbers $z_i$.  For any $t > 0$, we have
  \[
  \Pr[\abs{x - z_i} < t] \leq \frac{2t}{\sqrt{2\pi}} < t
  \]
  Hence
  \[
  \Pr[\abs{p(x)} < \eps] \leq \Pr[\exists i \text{ such that } \abs{x - z_i} < \eps^{1/d}] < d \eps^{1/d}.
  \]
\end{proof}

\state{lem:formalanticoncentration}

\begin{proof}
  We proceed by induction on $n$.  The base case is
  Lemma~\ref{lem:unip}.  If the degree of $x_n$ in $p$ is $d'$, then
  we write
  \[
  p(x_1, \dotsc, x_n) = \sum_{i=0}^{d'} q_i(x_1, \dotsc, x_{n-1}) x_n^i
  \]
  where $q_{d'}$ is a nonzero formal polynomial of degree $d - d'$
  with smallest nonzero coefficient $a$.  By induction, we
  have for any $t$ that
  \[
  \abs{q_i(x_1, \dotsc, x_{n-1})} \geq a t
  \]
  with probability at least $1 - (d - d') t^{1/(d - d')}$.  Consider any setting of $x_1,
  \dotsc, x_{n-1}$ for which this happens.  Conditioned on this,
  $p(x)$ is a univariate polynomial in $x_n$, so Lemma~\ref{lem:unip}
  shows for any $\eps$ that
  \[
  \Pr[ \abs{p(x)} < a \eps] < d'(\eps/t)^{1/d'}.
  \]
  Hence, unconditionally and using a union bound, we have that
  \[
  \Pr[ \abs{p(x)} < a \eps] < d' (\eps/t)^{1/d'} + (d - d') t^{1/(d - d')}
  \]
  for an arbitrary $t$.  Optimizing with $t = \eps^{(d-d')/d}$, we get
  \[
  \Pr[ \abs{p(x)} < a \eps] < d \eps^{1/d}.
  \]
  This shows the inductive step, getting the result.
\end{proof}

\section{Algorithm for $d = 1$}

\subsection{Sympy}\label{sec:sympy}

The proofs in this section involve a fair amount of algebraic
manipulation.  To make these computations more reliable and easier to
verify, in some cases we provide code for a computer to do them.  We
use Sympy~\cite{sympy}, a standard Python package for symbolic
manipulation.  We only use Sympy for simple tasks -- multiplying and
adding polynomials, substituting expressions for variables -- that can
be verified by hand.

\subsection{Excess Moments of a gaussian Mixture}\label{app:excess}

\restate{l:excess}
\begin{proof}
For a standard $N(\mu, \sigma^2)$ gaussian we have moments
\begin{align*}
  M_2 &= \mu^2 + \sigma^2\\
  M_3 &= \mu^3 + 3\mu \sigma^2\\
  M_4 &= \mu^4 + 6 \mu^2 \sigma^2 + 3 \sigma^4\\
  M_5 &= \mu^5 + 10 \mu^3 \sigma^2 + 15 \mu \sigma^4\\
  M_6 &= \mu^6 + 15 \mu^4 \sigma^2 + 45 \mu^2 \sigma^4 + 15 \sigma^6.
\end{align*}
and the mixture has probability $p_1 = \mu_2/(\mu_2-\mu_1)$ and $p_2 = -\mu_1/(\mu_2-\mu_1)$.

Therefore the following Sympy code (see Section~\ref{sec:sympy} for an
explanation of Sympy) can be used to formally verify the result:
\begin{verbatim}
#! /usr/bin/python
from sympy import *

# Define variables
mu1 = Symbol(r'\mu_1')
sigma1 = Symbol(r'\sigma_1')
mu2 = Symbol(r'\mu_2')
sigma2 = Symbol(r'\sigma_2')
alpha = Symbol(r'\alpha')
beta = Symbol(r'\beta')
gamma = Symbol(r'\gamma')

p1 = mu2/(mu2 - mu1)

# Moments of single (mu1, sigma1) gaussian
M2 = mu1**2 +               sigma1**2
M3 = mu1**3 +  3 * mu1    * sigma1**2
M4 = mu1**4 +  6 * mu1**2 * sigma1**2 +  3 *          sigma1**4
M5 = mu1**5 + 10 * mu1**3 * sigma1**2 + 15 * mu1    * sigma1**4
M6 = mu1**6 + 15 * mu1**4 * sigma1**2 + 45 * mu1**2 * sigma1**4 + 15*sigma1**6

# Convert to moments of mixture
M2 = p1 * M2 + (1-p1)*M2.subs({mu1:mu2, sigma1:sigma2})
M3 = p1 * M3 + (1-p1)*M3.subs({mu1:mu2, sigma1:sigma2})
M4 = p1 * M4 + (1-p1)*M4.subs({mu1:mu2, sigma1:sigma2})
M5 = p1 * M5 + (1-p1)*M5.subs({mu1:mu2, sigma1:sigma2})
M6 = p1 * M6 + (1-p1)*M6.subs({mu1:mu2, sigma1:sigma2})

# Claimed excess moments
x3 = alpha*beta + 3*alpha*gamma
x4 = -2*alpha**2 + alpha*beta**2 + 6*alpha*beta*gamma  + 3*alpha*gamma**2
x5 = alpha * (beta**3 - 8*alpha*beta + 10*beta**2*gamma + 15*gamma**2*beta
              - 20*alpha*gamma)
x6 = alpha*(16*alpha**2 - 12*alpha*beta**2 - 60*alpha*beta*gamma + beta**4 +
            15*beta**3*gamma + 45*beta**2*gamma**2 + 15*beta*gamma**3)

# Check that they match

alphadefs = {alpha: -mu1*mu2, beta: mu1+mu2,
             gamma: (sigma2**2-sigma1**2)/(mu2-mu1)}

print (M3 - x3.subs(alphadefs)).factor()
print (M4-3*M2**2 - x4.subs(alphadefs)).factor()
print (M5-10*M3*M2 - x5.subs(alphadefs)).factor()
print (M6-15*M4*M2 + 30*M2**3 - x6.subs(alphadefs)).factor()
\end{verbatim}
All the results are zero, so the claimed $X_i$ are correct.
\end{proof}

\subsection{Expressing $\alpha \gamma$ using $X_3, X_4, X_6, \alpha$}
\label{app:zfrac6}

We prove~\eqref{eq:zfrac6}, which is analogous to~\eqref{eq:zfrac5}.
We demonstrate using the following Sympy code (see
Section~\ref{sec:sympy} for an explanation of Sympy):
\begin{verbatim}
#! /usr/bin/python
from sympy import *

# define variables
alpha = Symbol(r'\alpha')
beta = Symbol(r'\beta')
gamma = Symbol(r'\gamma')
X3 = Symbol('X3')
X4 = Symbol('X4')
X5 = Symbol('X5')
X6 = Symbol('X6')
z = Symbol('z')

# define expressions for X_i in terms of alpha, beta, gamma
x3 = alpha*beta + 3*alpha*gamma
x4 = -2*alpha**2 + alpha*beta**2 + 6*alpha*beta*gamma  + 3*alpha*gamma**2
x5 = alpha * (beta**3 - 8*alpha*beta + 10*beta**2*gamma + 15*gamma**2*beta
              - 20*alpha*gamma)
x6 = alpha*(16*alpha**2 - 12*alpha*beta**2 - 60*alpha*beta*gamma + beta**4 +
            15*beta**3*gamma + 45*beta**2*gamma**2 + 15*beta*gamma**3)

# we know that this should be zero.
eqn = alpha**3 * (x6 - X6)
print eqn, '= 0'
\end{verbatim}
\[
-\alpha^3 (X_6 - 16 \alpha^3 + 12 \alpha^2 \beta^2 + 60 \alpha^2 \beta \gamma - \alpha \beta^4 - 15 \alpha \beta^3 \gamma - 45 \alpha \beta^2 \gamma^2 - 15 \alpha \beta \gamma^3) = 0
\]
\begin{verbatim}
eqn = eqn.expand().subs(alpha*beta, X_3-3*alpha*gamma) # remove beta
print eqn, '= 0'
\end{verbatim}
\[
X_3^4 + 3 X_3^3 \alpha \gamma - 12 X_3^2 \alpha^3 - 36 X_3^2 \alpha^2 \gamma^2 + 12 X_3 \alpha^4 \gamma + 42 X_3 \alpha^3 \gamma^3 - X_6 \alpha^3 + 16 \alpha^6 + 72 \alpha^5 \gamma^2 + 36 \alpha^4 \gamma^4 = 0
\]
\begin{verbatim}
# Use (7) to remove gamma**2 terms
eqn = eqn.expand().subs(alpha**2*gamma**2, (X3**2 - 2*alpha**3 - X4*alpha)/6)
print eqn, '= 0'
\end{verbatim}
\[
-4 X_3^4 + 10 X_3^3 \alpha \gamma + 4 X_3^2 X_4 \alpha + 8 X_3^2 \alpha^3 - 7 X_3 X_4 \alpha^2 \gamma - 2 X_3 \alpha^4 \gamma + X_4^2 \alpha^2 - 8 X_4 \alpha^4 - X_6 \alpha^3 - 4 \alpha^6 = 0
\]
\begin{verbatim}
eqn = eqn.subs(alpha*gamma, z).expand().collect(z)  # this of the form f*z + g
print eqn, '= 0'
\end{verbatim}
\[
z (10 X_3^3 - 7 X_3 X_4 \alpha - 2 X_3 \alpha^3) -4 X_3^4 + 4 X_3^2 X_4 \alpha + 8 X_3^2 \alpha^3 + X_4^2 \alpha^2 - 8 X_4 \alpha^4 - X_6 \alpha^3 - 4 \alpha^6 = 0
\]
\begin{verbatim}
answer = -eqn.subs(z, 0) / eqn.coeff(z)
print 'z = ', answer
\end{verbatim}
\[
z = \frac{4 X_3^4 - 4 X_3^2 X_4 \alpha - 8 X_3^2 \alpha^3 - X_4^2 \alpha^2 + 8 X_4 \alpha^4 + X_6 \alpha^3 + 4 \alpha^6}{X_3 (10 X_3^2 - 7 X_4 \alpha - 2 \alpha^3)}
\]
which is \eqref{eq:zfrac6}.

\subsection{Bounding $r$ away from zero}\label{sec:quadratic}

This section proves the following lemma:
\restate{lemma:quadratic}

We start by showing that $r(y) = 0$ has a unique solution on $(0,
y_{max}]$.  The following lemma shows that for any such solution $y$
there exists a gaussian mixture with $\alpha = y$ and matching excess
moments; since the first six moments uniquely identify a gaussian
mixture, this gives uniqueness.

\begin{lemma}\label{l:solutionworks}
  For any solution $y$ to the system of equations
  \begin{align}
    p_5(y) &= 0\notag\\
    p_6(y) &= 0\label{eq:set}\\
    y &> 0\notag\\
    X_3^2 - 2y^3 + X_4 y&\geq 0\notag
  \end{align}
  there exists a mixture of gaussians with $\alpha = y$ and excess
  moments $X_3, \dotsc, X_6$.
\end{lemma}
\begin{proof}
  We set the recovered $\wh{\alpha} = y$, recover $\gamma$
  via~\eqref{eq:zfrac5}:
  \[
  \wh{\gamma} = \frac{1}{\wh{\alpha}}\frac{\wh{\alpha}^2 X_5 + 2X_3^3 + 2 \wh{\alpha}^3 X_3 - 3 X_3X_4\wh{\alpha}}{4X_3^2 - 3 X_4 \wh{\alpha}- 2
    \wh{\alpha}^3}.
  \]
  which is well defined, using~\eqref{eq:X4X3} and that the
  denominator is
  \[
  4X_3^2 - 3X_4\wh{\alpha} - 2\wh{\alpha}^3 = 4 \wh{\alpha}^3 + X_3^2 + 3 (X_3^2 - 2\wh{\alpha}^3 - X_4\wh{\alpha}) > 0.
  \]
  We then recover
  \[
  \wh{\beta} = \frac{1}{\wh{\alpha}} (X_3 - 3 \wh{\alpha} \wh{\gamma}).
  \]

  Now, our $\wh{\alpha}$ and $\wh{\gamma}$ satisfy~\eqref{eq:zfrac5}
  and~\eqref{eq:p5}, which implies that~\eqref{eq:X4X3} is satisfied
  as well.

  Now, consider the excess moments $X'_i$ of the gaussian mixture with
  parameters $(\wh{\alpha}, \wh{\beta}, \wh{\gamma})$.  By choice of
  $\wh{\beta}$, $X'_3 = X_3$.  Then since~\eqref{eq:X4X3} is satisfied
  by both $(X_3, X_4)$ and $(X_3, X_4')$ and the coefficient of $X_4'$
  is nonzero, $X'_4 = X_4$.  Similarly with~\eqref{eq:zfrac5}, $X'_5 = X_5$.

  What remains is to show $p_6(\wh{\alpha}) = 0$ implies $X'_6 = X_6$.
  The coefficient of $X_6$ in $p_6$ is
  \[
  -2\alpha^2(4X_3^2 - 3X_4\wh{\alpha} - 2\wh{\alpha}^3) < 0.
  \]
  Thus since $p_6 = 0$ is satisfied by both $(\wh{\alpha}, X_3, X_4,
  X_5, X_6)$ and $(\wh{\alpha}, X_3, X_4, X_5, X_6')$, $X_6' = X_6$.
\end{proof}

\begin{corollary}\label{cor:unique}
  The set of equations~\eqref{eq:set} has exactly one solution, $y =
  \alpha$.
\end{corollary}
\begin{proof}
  By construction, $y = \alpha$ is a solution to~\eqref{eq:set}.
  Suppose there existed another solution $y' \neq y$.  Then by
  Lemma~\ref{l:solutionworks}, there exist two mixtures of gaussians
  with different $\alpha$ and matching $X_3, \dotsc, X_6$.

  The excess moments are constructed to be indifferent to adding
  gaussian noise, i.e. increasing $\sigma_1^2$ and $\sigma_2^2$ by the
  same amount.  Hence, by ``topping off'' the second moment, we can
  construct two different mixtures of gaussians with identical second
  moment as well as identical $X_3, \dotsc, X_6$; these mixtures also
  both have mean zero.  Such mixtures would have identical first six
  moments.  But by~\cite{KalaiMV10}, any two different mixtures of
  gaussians differ in their first six moments.  So this is a
  contradiction.
\end{proof}

\begin{lemma}\label{l:double}
  For no $\alpha, \beta, \gamma$ with $\alpha > 0$ is it the case that
  both $p_5$ and $p_6$ have a double root at $y = \alpha$.
\end{lemma}
\begin{proof}
  Define $q_5(y) = p_5(y) / (y - \alpha)$, $q_6(y) = p_6(y) / (y -
  \alpha)$.  We need to show that it is not true that both
  $q_5(\alpha) = 0$ and $q_6(\alpha) = 0$.  We show this by repeatedly
  adding multiples of one to the other, essentially taking the GCD.
  The below is a transcript of a Sympy session (see
  Section~\ref{sec:sympy} for an explanation of Sympy) proving this
  claim.  \begin{verbatim}
#! /usr/bin/python
# Setup variables and expressions

from sympy import *

alpha = Symbol(r'\alpha')
beta = Symbol(r'\beta')
gamma = Symbol(r'\gamma')
y = Symbol('y')

X3 = alpha * beta + 3 * alpha * gamma
X4 = -2*alpha**2 + alpha*beta**2 + 6*alpha*beta*gamma  + 3*alpha*gamma**2
X5 = alpha * (beta**3 - 8*alpha*beta + 10*beta**2*gamma + 15*gamma**2*beta
              - 20*alpha*gamma)
X6 = alpha*(16*alpha**2 - 12*alpha*beta**2 + beta**4 + 45*beta**2*gamma**2 +
            15*beta*gamma**3 + 15*gamma*(-4*alpha*beta + beta**3))
p5 = (6*(2*X3*y**3 + X5*y**2 - 3*X3*X4*y + 2*X3**3)**2 +
      (2*y**3+3*X4*y - 4*X3**2)**2*(2*y**3 + X4*y - X3**2))
p6 = ((4*X3**2 - 3*X4*y - 2*y**3) *
      (4*X3**4 - 4*X3**2*X4*y - 8*X3**2*y**3 - X4**2*y**2 +
       8*X4*y**4 + X6*y**3 + 4*y**6) -
      (10*X3**3 - 7*X3*X4*y - 2*X3*y**3) *
      (2*X3**3 - 3*X3*X4*y + 2*X3*y**3 + X5*y**2))

# Start actual code

q5 = (p5 / (y - alpha)).factor().subs(y, alpha).factor()
q6 = (p6 / (y - alpha)).factor().subs(y, alpha).factor()

# Our goal is to show that (q5, q6) != (0, 0) for any alpha, beta, gamma.

print q5
\end{verbatim}
\[
\alpha^5(4\alpha + \beta^2 + 6\beta\gamma + 27\gamma^2)(16\alpha^2 + 8\alpha\beta^2 + 72\alpha\gamma^2 + \beta^4 + 18\beta^2\gamma^2 - 135\gamma^4)
\]
\texttt{\# and since $\alpha^5(4\alpha + \beta^2 + 6\beta\gamma + 27\gamma^2) > 0$, we can reduce by}
\begin{verbatim}
q5a = q5 / (alpha**5*(4*alpha + beta**2 + 6*beta*gamma + 27*gamma**2))
print q5a.subs(gamma, 0).factor()
\end{verbatim}
\[
(4\alpha + \beta^2)^2
\]
\texttt{\# nonzero, so must have $\gamma \neq 0$.}
\begin{verbatim}
print q6
\end{verbatim}
\[
\alpha^5\gamma(16\alpha^2\beta - 48\alpha^2\gamma + 8\alpha\beta^3 - 24\alpha\beta^2\gamma - 72\alpha\beta\gamma^2 - 72\alpha\gamma^3 + \beta^5 - 3\beta^4\gamma - 18\beta^3\gamma^2 - 18\beta^2\gamma^3 + 405\beta\gamma^4 + 2025\gamma^5)
\]
\begin{verbatim}
q6a = (q6 / (alpha**5*gamma) - beta * q5a).factor()
print q6a
\end{verbatim}
\[
-3\gamma(16\alpha^2 + 8\alpha\beta^2 + 48\alpha\beta\gamma + 24\alpha\gamma^2 + \beta^4 + 12\beta^3\gamma + 6\beta^2\gamma^2 - 180\beta\gamma^3 - 675\gamma^4)
\]
\begin{verbatim}
q6b = (q6a / (-3*gamma) - q5a).factor()
print q6b
\end{verbatim}
\[
12\gamma(4\alpha\beta - 4\alpha\gamma + \beta^3 - \beta^2\gamma - 15\beta\gamma^2 - 45\gamma^3)
\]
\begin{verbatim}
q6c = q6b / (12*gamma)

z = 4*alpha + beta**2
q5b = (z + 9*gamma**2)**2 - 216*gamma**4
q6d = (z - 15*gamma**2)*(beta - gamma) - 60*gamma**3
print (q5a - q5b).factor(), (q6d - q6c).factor()
\end{verbatim}
\[
(0, 0)
\]

So the solution must have $(z + 9\gamma^2)^2 - 216\gamma^4 = 0$ and
$(z - 15\gamma^2)(\beta - \gamma) - 60\gamma^3 = 0$ for $z = 4 \alpha
+ \beta^2$.  From the first equation,
\[
z = -9\gamma^2 \pm 6\sqrt{6} \gamma^2
\]
and since $z > 0$, this means
\[
z = (6\sqrt{6} - 9) \gamma^2.
\]
Plugging into the second equation and dividing by $\gamma^2$,
\[
(6\sqrt{6} - 24)\beta - (36 + 6 \sqrt{6})\gamma = 0
\]
and so
\begin{align*}
  \gamma &= -\frac{4 - \sqrt{6}}{6 + \sqrt{6}} \beta\\
  z &= \frac{(6\sqrt{6}-9)(4 - \sqrt{6})^2}{(6 + \sqrt{6})^2}\beta^2 \approx 0.19\beta^2\\
  & \leq \beta^2 < 4 \alpha + \beta^2 = z
\end{align*}
a contradiction.
\end{proof}

\begin{lemma}[$r$ is large when $\gamma$ is unbounded]\label{l:qlargegamma}
  There exists a constant $C$ such that for all $\beta, \gamma \in \R$
  and $\alpha, y > 0$ with $\gamma^2 \geq C\alpha$, $y \lesssim
  y_{max}$, and $\beta^2 \lesssim \alpha$ we have
  \[
  r(y) \gtrsim (y-\alpha)^2\gamma^{12}\alpha^{10}
  \]
\end{lemma}
\begin{proof}
  As in Lemma~\ref{l:double}, define $q_5(y) = p_5(y) / (y -
  \alpha)$ and $q_6(y) = p_6(y) / (y - \alpha)$ Per~\eqref{eq:ymax},
  we have that $y_{max} \lesssim \alpha + \beta^2 \lesssim \alpha$.

  Then $q_5$ is a homogeneous polynomial in $\sqrt{y}, \sqrt{\alpha},
  \beta, \gamma$, all of which are $O(\sqrt{\alpha})$ except $\gamma$.
  The leading $\gamma^6$ term of $q_5$ is

  \texttt{>>> q5 = (p5 / (y - alpha)).factor()}

  \texttt{>>> q5.expand().coeff(gamma**6).factor()}
  \[
  243\alpha^3((y + 4 \alpha)^2 - 40 \alpha^2)\gamma^6
  \]
  Hence for $y \notin (2\alpha, 3\alpha)$, $\abs{q_5(y)} \gtrsim
  \gamma^6 \alpha^5$.

  Doing the same for $p_6$, for $p_6 / (y - \alpha)$ the leading
  $\gamma^6$ term is

  \texttt{>>> q6 = (p6 / (y - alpha)).factor()}

  \texttt{>>> q6.expand().coeff(gamma**6).factor()}
  \[
  162\alpha^3(y - 6 \alpha)^2\gamma^6
  \]
  so for $y \notin (5\alpha, 7\alpha)$, $\abs{q_5(y)} \gtrsim \gamma^6
  \alpha^5$.

  Thus for all $y$ with $\gamma^2$ sufficiently much greater than $\alpha$,
  \[
  r(y) = (y-\alpha)^2(q_5(y)^2 + q_6(y)^2) \gtrsim (y - \alpha)^2 \gamma^{12} \alpha^{10}.
  \]
\end{proof}

\begin{lemma}[$r$ is large when $\gamma$ is bounded]\label{l:qsmallgamma}
  For any constant $c > 0$, and for all $\alpha \geq 0$ and $\beta,
  \gamma, y \in \R$ with $cy_{max} \leq y \leq y_{max}$ and $\beta^2,
  \gamma^2 \lesssim \alpha$ we have
  \[
  r(y) \gtrsim (y-\alpha)^2\alpha^{16}
  \]
\end{lemma}
\begin{proof}
  Define the polynomial $q(y, \alpha, \beta, \gamma) = r(y) / (y -
  \alpha)^2 = q_5^2 + q_6^2$.  By homogeneity of the constraints and
  result, we may normalize so that $\alpha = 1$ and consider $q(y,
  \beta, \gamma) := q(y, 1, \beta, \gamma)$.  Our goal is to show that
  $q(y, \beta, \gamma) \gtrsim 1$.

  Define $R$ to be the region of $(y, \beta, \gamma)$ allowed by the
  lemma constraints.  By our normalization, $y \eqsim y_{max} \eqsim
  \alpha = 1$ and $\gamma^2, \beta^2 \lesssim 1$ over $R$.  This means
  $R$ is closed and bounded and hence compact.

  By Corollary~\ref{cor:unique}, $q$ can only be zero over $R$ when $y
  = 1$.  By Lemma~\ref{l:double}, $q(1, \beta, \gamma) \neq 0$ for all
  $\beta, \gamma$.  Hence $q$ has no roots over $R$.  Because $R$ is
  compact, this means $q \gtrsim 1$ over $R$, giving the result.
\end{proof}
\restate{lemma:quadratic}
\begin{proof}
  This is simply the union of Lemma~\ref{l:qlargegamma} and
  Lemma~\ref{l:qsmallgamma}.
\end{proof}

\restate{lemma:quadratic+}
\begin{proof}
  By homogeneity of the equations, we may assume $y_{max} = 1$ and $\alpha
  \eqsim 1$.

  By Lemma~\ref{lemma:quadratic}, $r(1) \gtrsim (1 -
  \alpha)^2\kappa^{12}$.  That is, one of $\abs{p_5(1)}$ and
  $\abs{p_5(1)}$ is $\Omega((1-\alpha)\kappa^6)$.  Since $p_5(y)$ and
  $p_6(y)$ are constant degree polynomials with coefficients of magnitude
  $O(\kappa^6)$, their derivatives over $[1, 2]$ are bounded in
  magnitude by $O(\kappa^6)$.  Hence for all $y \in [1, 1 + c'(1 -
  \alpha)]$,
  \[
  \abs{p_5(y)} \geq \abs{p_5(1)} - O(\kappa^6) \cdot \abs{y - 1} \geq
  \abs{p_5(1)} - O(c'(1-\alpha)\kappa^6)
  \]
  and similarly for $p_6$.  For sufficiently small $c$, if
  $\abs{p_5(1)}$ is the $\Omega((1-\alpha)^2\kappa^6)$ term, this is
  $\Omega((1-\alpha)\kappa^6)$.  If instead $\abs{p_6(1)}$ is the
  $\Omega((1-\alpha)\kappa^6)$ term, then $\abs{p_6(y)}$ is
  $\Omega((1-\alpha)\kappa^6)$; regardless, the conclusion holds.
\end{proof}

\subsection{Accuracy of estimating $r$}\label{sec:accuracy}

\begin{lemma}\label{l:papprox}
  Suppose that $\abs{\wh{X}_i - X_i} \lesssim
  \eps\kappa^{i-2}\Delta_\mu^i$ for all $i \in \{3,4,5,6\}$ and some
  $\eps < 1$.  Then for any $y \lesssim \Delta_\mu^2$,
  \begin{align*}
    \abs{\wh{p_5}(y) - p_5(y)} &\lesssim \eps\kappa^6 \Delta_\mu^{18}\\
    \abs{\wh{p_6}(y) - p_6(y)} &\lesssim \eps\kappa^6 \Delta_\mu^{18}
  \end{align*}
  Hence
  \[
  \sqrt{\wh{r}(y)} - \sqrt{r(y)} \lesssim \eps\kappa^6 \Delta_\mu^{18}.
  \]
\end{lemma}
\begin{proof}
  Recall from~\eqref{eq:Xkappa} that $\abs{X_i} \lesssim
  \kappa^{i-2}\Delta_\mu^i$ and from~\eqref{eq:monomialbound} that
  this means each monomial of $p_5(y)$ and $p_6(y)$ is bounded by
  $O(\kappa^6\Delta_\mu^{18})$.  Since $p_5$ and $p_6$ are constant
  size polynomials, the first result follows from by
  Lemma~\ref{l:polynomials}.

  For the second claim, we use that $\sqrt{r(y)} = \sqrt{p_5(y)^2 +
    p_6(y)^2}$ is Lipschitz in $p_5(y)$ and $p_6(y)$.
\end{proof}

\subsection{Recovering $\alpha$}\label{sec:getalpha}

\begin{lemma}\label{l:ymax}
  In \textsc{RecoverAlphaFromMoments}, for any $\eps$ if
  $\abs{\wh{X}_i - X_i} \leq \eps \Delta_\mu^i$ for $i \in \{3, 4\}$,
  then the estimation $\wh{y}_{max}$ of $y_{max}$ satisfies
  \[
  \abs{\wh{y}_{max} - y_{max}} \lesssim \eps \Delta_\mu^2 / \kappa^2.
  \]
\end{lemma}
\begin{proof}
  We would like to know the stability of the largest root $y_{max}$ of
  the polynomial $s(y) := 2y^3 + X_4y - X_3^2$ to perturbations in
  $X_4$ and $X_3$.  Without loss of generality we normalize so that
  $y_{max} \eqsim \Delta_\mu^2 \eqsim 1$.

  Since $y_{max}$ is a root, $2y_{max}^3 + X_4y_{max} = X_3^2 \geq 0$ so
  $2y_{max}^2 + X_4 \geq 0$.  Hence for all $y \geq (2/3)y_{max}$,
  \[
  s'(y) = 6y^2 + X_4 \geq (8/3)y_{max}^2 + X_4 \geq (2/3) y_{max}^2
  \gtrsim 1.
  \]
  But we also have that
  \[
  s'(y) = 6y^2 + X_4 \geq X_4 \gtrsim \gamma^2 -
  O(1) \eqsim (\kappa^2 - O(1)).
  \]
  Combining gives that for all $y \geq (2/3)y_{max}$,
  \[
  s'(y) \gtrsim \kappa^2.
  \]
  This implies for any parameter $t > 0$ that
  \[
  s(y_{max} - t \eps ) \lesssim -t\eps \kappa^2
  \]
  as long as $t \eps \leq y_{max}/3$, and
  \[
  s(y_{max} + t \eps ) \gtrsim t\eps \kappa^2
  \]
  for all $t > 0$.  On the other hand, for all $y \eqsim 1$
  we have
  \[
  \abs{\wh{s}(y) - s(y)} \leq \abs{\wh{X}_4 - X_4}y + \abs{\wh{X}_3^2 - X_3^2} \lesssim \eps.
  \]
  Let $t = C/\kappa^2$ for sufficiently large constant $C$.  

  If $t \eps \leq y_{max}/3$, then combining gives that $\wh{s}$ must
  have a root within $y_{max} \pm t\eps$ and no root above this range.
  This is the desired result.

  On the other hand, if $t \eps \geq y_{max}/3$, then it
  is still true that $\wh{s}$ has no root above $y_{max} + t \eps$.
  Since $\wh{s}(0) = -\wh{X}_3^2 \leq 0$, we also have that
  $\wh{y}_{max} \geq 0$.  Hence the result lies in $[0, y_{max} + t
  \eps]$, which is still $y_{max} \pm O(\eps /\kappa^2)$ in this
  parameter regime.
\end{proof}

\restate{l:alphafrommoments}

\begin{proof}
  We will suppose $\abs{\wh{X}_i - X_i} \lesssim \eps \Delta_\mu^i$,
  and show for a sufficiently large constant $C$ that $\wh{\alpha} =
  \Call{RecoverAlphaFromMoments}{\wh{X}_3, \wh{X}_4, \wh{X}_5,
    \wh{X}_6, C\eps}$ satisfies $\abs{\wh{\alpha} - \alpha} \lesssim
  \eps \alpha / \kappa$.  Rescaling $\eps$ gives the result.

  We normalize so $\alpha \eqsim \Delta_\mu^2 \eqsim 1$.

  By Lemma~\ref{l:ymax}, $\abs{\wh{y}_{max} - y_{max}} \lesssim
  \eps/\kappa^2 \leq \eps/\kappa$.  Since $\wh{y}_{max} \eqsim y_{max}
  \eqsim 1$, $\wh{X}_4 \lesssim \kappa^2$, and $\wh{X}_4 \eqsim
  \kappa^2$ if $\kappa \gg 1$, the estimation $\wh{\kappa}$ of
  $\kappa$ is always
  \[
  \wh{\kappa} := 1 + \sqrt{\abs{\wh{X}_4}}/\wh{y}_{max} \eqsim \kappa.
  \]
  Hence $\alpha \leq y_{max} < (1 + O(\eps/\kappa))\wh{y}_{max}$.

  We have by Lemma~\ref{l:papprox} with $\eps' = \eps / \kappa$ that
  $\sqrt{\wh{r}(y)} = \sqrt{r(y)} \pm O(\eps\kappa^5)$ for all $y
  \lesssim 1$.  In particular, this means that
  \begin{align}
    \sqrt{\wh{r}(\alpha)} \lesssim \eps\kappa^{5}.
  \end{align}
  Moreover, by Lemma~\ref{lemma:quadratic} for all $y \in [\alpha/2,
  y_{max}]$ we have $r(y) \gtrsim (y - \alpha)^2\kappa^{12}$.
  Therefore for some sufficiently large constant $c$, for all $y \in
  [\alpha/2, y_{max}]$ with $\abs{y - \alpha} > c\eps/\kappa$ we have
  \begin{align}
    \sqrt{\wh{r}(y)} \gtrsim \abs{y - \alpha}\kappa^6 - O(\eps\kappa^5) \geq
    \frac{1}{2}\abs{y-\alpha}\kappa^{6} \gtrsim c\eps\kappa^{5}
    > \sqrt{\wh{r}(\alpha)}.
  \end{align}

  And by Lemma~\ref{lemma:quadratic+}, if $\alpha <
  (1-O(c\eps/\kappa))y_{max}$ then $r(y') \gtrsim
  c^2\eps^2\kappa^{10}$ for all $y' \in [y_{max}, (1 +
  O(c\eps/\kappa))y_{max}$.

  This implies (A) that a local minimum of $\wh{r}$ over $[0, (1 +
  O(c\eps/\kappa))\wh{y}_{max}]$ is $\alpha \pm O(c\eps/\kappa)$,
  and (B) that any larger local minimum $y'$ has $\wh{r}(y') \gtrsim
  c^2\eps^2\kappa^{10}$.

  By definition, $\Call{RecoverAlphaFromMoments}{\wh{X}_3, \wh{X}_4,
    \wh{X}_5, \wh{X}_6, C\eps}$ finds the largest local minimum
  $\wh{\alpha}$ of $\wh{r}$ with $\wh{\alpha} \leq (1 +
  C\eps/\kappa)\wh{y}_{max} \leq (1 + (C + O(1))\eps/\kappa)y_{max}$
  and $\wh{r}(\wh{\alpha}) \leq C^2\eps^2\kappa^{10}$.  For
  sufficiently large $C$ and $c$, (A) and (B) imply that $\wh{\alpha}
  = \alpha \pm O(\eps/\kappa)$.
\end{proof}

\subsection{Proof of Theorem~\ref{thm:1d}}\label{app:1dcombine}

\restate{thm:1d}

\begin{proof}
  Suppose the number of samples is $f^{-12}\log (1/\delta)$ so
  $f^{-12} \eqsim \eps^{-2}n$.  By Lemma~\ref{l:estimatemoments} we
  have with probability $1-\delta$ that all the $\wh{X}_i$ are within
  $\pm O(f^6 \sigma^i)$ of the true moments $X_i$.  Suppose this
  happens.

  First, we show that $\overline{\Delta}_{\sigma^2}$ and
  $\overline{\Delta}_{\mu}$ are good approximations to
  $\Delta_{\sigma^2}$ and $\Delta_\mu$, and therefore the conditionals
  are followed roughly in the same cases as they would if they were
  not approximations.

  \paragraph{The first conditional.}
  We have that, if $\Delta_{\sigma^2} \gg \Delta_\mu^2$, then $X_4
  \eqsim \Delta_{\sigma^2}^2$ and $X_3 \eqsim \Delta_\mu
  \Delta_{\sigma^2}$.  Then $X_4 \eqsim \Delta_{\sigma^2}^2 > 0$ and
  \[
  \min(\abs{X_3}^{1/3} + \abs{X_4}^{1/4},~~X_3/\sqrt{X_4}) \eqsim \min(\Delta_{\sigma^2}^{1/2}, \Delta_\mu) = \Delta_\mu
  \]
  Therefore, as long as $f^6 \sigma^4 \ll \Delta_{\sigma^2}^2$, in
  the $\Delta_{\sigma^2} \gg \Delta_{\mu}^2$ setting we have
  $\overline{\Delta}_{\mu} \eqsim \Delta_{\mu}$.

  Otherwise, i.e. when $\Delta_{\sigma^2} \lesssim \Delta_\mu^2$, we
  have that $\abs{X_4} \lesssim \Delta_\mu^4$ and $\abs{X_3} \lesssim
  \Delta_\mu^3$.  Moreover, in this case, since $X_4 = X_3^2/\alpha -
  6 \alpha \gamma^2 - 2 \alpha^2 = \Theta(X_3^2 / \Delta_\mu^2) -
  \Theta(\Delta_\mu^4)$, either $\abs{X_3} \eqsim \Delta_\mu^3$ or
  $\abs{X_4} \eqsim \Delta_\mu^4$.  If $X_4 > 0$, then $X_3^2/X_4
  \gtrsim \Delta_\mu^2$.  This is because the single positive root
  $y_{max}$ of~\eqref{eq:ymax} is $\Theta(\Delta_\mu^2)$ and the
  polynomial is positive at $X_3^2/X_4$.

  Thus, when $\Delta_{\sigma^2} \lesssim \Delta_\mu^2$,
  \begin{align*}
    \abs{X_3}^{1/3} + \abs{X_4}^{1/4} &\eqsim \Delta_\mu\\
    X_3/\sqrt{X_4} &\gtrsim  \Delta_{\mu} \text{ if } X_4 > 0
  \end{align*}
  As long as $f^6 \sigma^4 \lesssim \Delta_\mu^4$, we will have
  $\abs{\wh{X}_3}^{1/3} + \abs{\wh{X}_4}^{1/4} \eqsim \abs{X_3}^{1/3}
  + \abs{X_4}^{1/4} \eqsim \Delta_\mu$.  For
  $\wh{X}_3/\sqrt{\wh{X}_4}$, if $X_3 \ll \Delta_\mu^3$ then $X_4
  \eqsim -\Delta_\mu^4$ so $\wh{X}_4 < 0$.  Otherwise,
  $\wh{X}_3/\sqrt{\wh{X}_4} \gtrsim (\Delta_\mu^3 - f^6
  \sigma^3)/\sqrt{\Delta_\mu^4 + f^6 \sigma^4} \gtrsim \Delta_\mu$.
  Therefore, as long as $f^6 \sigma^4 \ll \Delta_\mu^4$, in the
  $\Delta_{\sigma^2} \lesssim \Delta_{\mu}^2$ setting we have
  $\overline{\Delta}_{\mu} \eqsim \Delta_{\mu}$.

  Thus, regardless of the relationship between $\Delta_\mu^2$ and
  $\Delta_{\sigma^2}$, we have that $\overline{\Delta}_{\mu} \eqsim
  \Delta_{\mu}$ as long as $f^6 \sigma^4 \ll \Delta_\mu^4 +
  \Delta_{\sigma^2}^2$.  In particular, $\overline{\Delta}_{\mu}
  \eqsim \Delta_{\mu}$ whenever $f^2 \lesssim \Delta_\mu^2 / \sigma^2$.

  Therefore the first conditional in Algorithm~\ref{alg:1dcombine} is
  followed if and only if $f^2 \lesssim \Delta_\mu^2/\sigma^2$; that
  is to say, if the first conditional is that $f^2 \leq C_1
  \overline{\Delta}_\mu^2/\sigma^2$, then it is followed whenever
  $f^2 \leq (C_1 - O(1)) \Delta_\mu^2 / \sigma^2$ and not followed
  whenever $f^2 \geq (C_1 + O(1)) \Delta_\mu^2 / \sigma^2$.

  \paragraph{The second conditional.}
  We have that $X_4 = \Theta(\Delta_{\sigma^2}^2) \pm
  O(\Delta_{\mu}^4)$.  Therefore $\overline{\Delta}_{\sigma^2}
  \lesssim \Delta_{\sigma^2} + \Delta_{\mu}^2 + f^3\sigma^2$ in
  general, and if $f^3 \sigma^2 \leq c\Delta_{\sigma^2}$ and
  $\Delta_{\mu}^2 \leq c\Delta_{\sigma^2}$ for some sufficiently small
  constant $c > 0$, then $\overline{\Delta}_{\sigma^2} \eqsim
  \Delta_{\sigma^2}$.

  Therefore the second conditional in Algorithm~\ref{alg:1dcombine} is
  taken if and only if $\Delta_\mu^2/\sigma^2 \lesssim f^2 \lesssim
  \Delta_{\sigma^2} / \sigma^2$; that is to say, if the second
  conditional is that $f^2 \leq C_2
  \overline{\Delta}_\mu^2/\sigma^2$, then that branch is taken
  whenever 
  \[
  (C_1 + O(1))\Delta_\mu^2 / \sigma^2 \leq f^2 \leq (C_2 -
  O(1)) \Delta_{\sigma^2}/\sigma^2
  \]
  and not taken whenever
  \[
  (C_1 - O(1))\Delta_\mu^2 / \sigma^2 \geq f^2 \text{ or } f^2 \geq (C_2 +
  O(1)) \Delta_{\sigma^2}/\sigma^2
  \]

  \paragraph{The third branch.}
  The remainder is the third branch, which is taken when $f^2 \geq
  (C_2 + O(1)) \Delta_{\sigma^2}/\sigma^2$ and $f^2 \geq (C_1 + O(1))
  \Delta_{\mu}^2/\sigma^2$, and not taken if either condition is false
  after replacing $+O(1)$ by $-O(1)$.

  \paragraph{Completing the theorem.}
  We have shown that the three branches are taken in the same settings
  as they would be taken with the true $\Delta_\mu$ and
  $\Delta_{\sigma^2}$.  We now show that the clauses of the theorem
  correspond to the branches.

  In the first clause of the theorem, we have that $f^{-12} =
  C\frac{1}{\eps^2}n$ for sufficiently large $C$ and $n >
  (\sigma^2/\Delta_\mu^2)^6$.  Then $f^2 < C^{-1/6}
  \Delta_\mu^2/\sigma^2$, so the first branch will be taken and
  Algorithm~\ref{alg:1dgetmean} is run.  By Theorem~\ref{thm:1dmean},
  running Algorithm~\ref{alg:1dgetmean} with
  $O(\frac{1}{\eps^2}(\frac{\sigma}{\Delta_{\mu}})^{12}\log(1/\delta))$
  samples will recover the $p_i$ to additive $\eps$ error, the $\mu_i$
  to additive $\eps \Delta_\mu$ error, and the $\sigma_i^2$ to
  additive $\eps \Delta_\mu^2$ error.

  In the second clause of the theorem, we have that $f^2 < C^{-1/6}
  \Delta_{\sigma^2}/\sigma^2$ for sufficiently large $C$.  Hence, if
  the first branch is not taken, then the second branch is taken.  The
  first branch is only taken if it can recover the $\sigma_i^2$ to
  better than $\pm \abs{\mu_1 - \mu_2}^2$, which satisfies the second
  clause of the theorem.  The second branch invokes
  Algorithm~\ref{alg:1dsamemean}, which by
  Theorem~\ref{thm:1dsamemean} uses
  $O(\frac{1}{\eps^2}(\frac{\sigma}{\sqrt{\Delta_{\sigma^2}}})^{12}\log(1/\delta))$
  samples to estimate the $p_i$ to additive $\eps$ error and the
  $\sigma_i^2$ to additive $\eps \Delta_{\sigma^2}$ error, again
  satisfying the theorem.

  Finally, the last clause of the theorem is satisfied by both
  algorithms and by outputting the single gaussian $N(\mu, \sigma^2)$.
  (If the second branch is taken, the mean outputted is the sample
  mean, which suffices for this purpose.)
\end{proof}

\fi

\end{document}